\documentclass[jair,twoside,11pt,theapa]{article}
\usepackage{jair, theapa, rawfonts}

\jairheading{74}{2022}{1823-1863}{11/2021}{08/2022}
\firstpageno{1823}

\usepackage{booktabs}
\usepackage[toc,page]{appendix}
\usepackage{multirow}
\usepackage[hidelinks,implicit=false]{hyperref}
\usepackage{amsmath}
\usepackage{empheq}
\usepackage{subfig}
\usepackage{amsthm}
\usepackage{algorithm}
\usepackage[noend]{algpseudocode}
\usepackage{dsfont}
\usepackage{comment}
\newtheorem{theorem}{Theorem}

\newtheorem{example}{Example}[section]
\usepackage{color}
\usepackage{amssymb}
\usepackage{blkarray}
\usepackage{graphicx}
\usepackage{dsfont}

\newcommand{\MaxSAT}{\ensuremath{\mathsf{MaxSAT}}}
\newcommand{\weight}[1]{\ensuremath{\mathsf{wt} \! \left( #1 \right)}}

\newcommand{\Rule}{\ensuremath{{\mathcal{R}}}}

\newtheorem{construction}[theorem]{Construction}
\newcommand{\clause}{\ensuremath{\mathsf{clause}}}
\newcommand{\framework}{\ensuremath{\mathsf{IMLI}}}

\usepackage{algorithm}
\usepackage[noend]{algpseudocode}

\newcommand{\blue}[1]{\textcolor{blue}{#1}}

\newtheorem{prop}{Proposition}

\begin{document}

\title{Efficient Learning of  Interpretable Classification Rules}
\ShortHeadings{Efficient Learning of  Interpretable Classification Rules}{Ghosh, Malioutov, \& Meel}

\author{\name Bishwamittra Ghosh  \email bghosh@u.nus.edu\\
	\addr National University of Singapore
	\AND
	Dmitry Malioutov \email dmal@alum.mit.edu
	\AND
	\name Kuldeep S. Meel \email meel@comp.nus.edu.sg\\
	\addr National University of Singapore
}


\maketitle

\begin{abstract}
Machine learning has become omnipresent with applications in various safety-critical domains such as medical, law, and transportation. In these domains, high-stake decisions provided by machine learning necessitate researchers to design \emph{interpretable} models, where the prediction is understandable to a human. In interpretable machine learning, \emph{rule-based classifiers} are particularly effective in representing the decision boundary through a set of rules comprising input features. Examples of such classifiers include decision trees, decision lists, and decision sets. The interpretability of rule-based classifiers is in general related to the size of the rules, where smaller rules are considered more interpretable. To learn such a classifier, the brute-force direct approach is to consider an optimization problem 
that tries to learn the smallest classification rule that has close to maximum accuracy. This optimization problem is computationally intractable due to its combinatorial nature and thus, the problem is not scalable in large datasets. To this end, in this paper we study the triangular relationship among the accuracy, interpretability, and scalability of learning rule-based classifiers.

The contribution of this paper is an interpretable learning framework {\framework}, that is based on maximum satisfiability (MaxSAT) for synthesizing classification rules expressible in proposition logic. {\framework} considers a joint objective function to optimize the accuracy and the interpretability of classification rules and learns an optimal rule by solving an appropriately designed MaxSAT query. Despite the progress of MaxSAT solving in the last decade, the straightforward MaxSAT-based solution cannot scale to practical classification datasets containing thousands to millions of samples. Therefore, we incorporate an efficient incremental learning technique inside the MaxSAT formulation by integrating mini-batch learning and iterative rule-learning. The resulting framework learns a classifier by iteratively covering the training data, wherein in each iteration, it solves a sequence of smaller MaxSAT queries corresponding to each mini-batch. In our experiments, {\framework} achieves the best balance among prediction accuracy, interpretability, and scalability. For instance, {\framework} attains a competitive prediction accuracy and interpretability w.r.t.\ existing interpretable classifiers and  demonstrates impressive scalability on large datasets where both interpretable and non-interpretable classifiers fail. As an application, we deploy {\framework} in learning popular interpretable classifiers such as decision lists and decision sets. The source code is available at \url{https://github.com/meelgroup/mlic}.

\end{abstract}

\section{Introduction}

Machine learning models are presently deployed in safety-critical and high-stake decision-making arising in medical~\shortcite{erickson2017machine,kaissis2020secure,kononenko2001machine}, law~\shortcite{kumar2018law,surden2014machine}, and transportation~\shortcite{peled2019model,zantalis2019review}. In safety-critical domains, a surge of interest has been observed in designing interpretable, robust, and fair models instead of merely focusing on an accuracy-centric learning objective~\shortcite{bhagoji2018enhancing,doshi2017towards,du2019techniques,hancox2020robustness,holstein2019improving,mehrabi2021survey,molnar2020interpretable,murdoch2019interpretable}. In this study, our focus is on learning an interpretable machine learning classifier~\shortcite{rudin2019stop}, where the rules governing the prediction are explicit (by design) in contrast to black-box models.


In order to achieve the interpretability of predictions, decision functions in the form of classification rules such as decision trees,  decision lists, decision sets, etc.\ are particularly effective~\shortcite{bessiere2009minimising,dash2021lprules,ignatiev2021reasoning,izza2020explaining,lakkaraju2017interpretable,lakkaraju2016interpretable,letham2015interpretable,narodytska2018learning,rivest1987learning,wang2015falling,yu2020optimal}. Such models are used to either learn an interpretable model from the start or as proxies to provide post-hoc explanations of pre-trained black-box models~\shortcite{gill2020responsible,lundberg2017unified,moradi2021post,ribeiro2016should,slack2020fooling}.  At this point, it is important to acknowledge that interpretability is an informal notion, and formalizing it in full generality is challenging. In our context of rule-based classifiers, we will use the \emph{sparsity of rules} (that is, fewer rules each having fewer Boolean literals), which has been considered a proxy of interpretability in various domains, specifically in the medical domain~\shortcite{gage2001validation,lakkaraju2019faithful,letham2015interpretable,malioutov2013exact,myers1962myers}.  

In this paper, we study decision functions expressible in \emph{propositional logic}. In particular, we focus on learning logical formulas from data as classifiers and define interpretability in terms of the number of Boolean literals that the formula contains. In the following, we illustrate an interpretable classifier that decides if a tumor cell is malignant or benign based on different features of tumors.


\vspace{1em}
\boxed{
\begin{array}{rl}
	& \text{A tumor is malignant if}\\
	&[(\text{compactness SE} < 0.1) \text{\textbf{ OR }}  \neg (0.1 \le \text{concave points} < 0.2)]
	\text{\textbf{ \blue{AND} }} \\
	&[\neg (0.2 \le \text{area} < 0.3) \text{\textbf{ OR }} \neg (0.1 \le \text{largest symmetry }< 0.2)]
\end{array}
}
\vspace{1em}

\begin{example}
	We learn an interpretable classification rule in Conjunctive Normal Form for predicting the classification of a tumor cell. We consider WDBC (Wisconsin Diagnostic Breast Cancer) dataset~\shortcite{agarap2018breast} to learn a CNF formula with two \emph{clauses} and four Boolean \emph{literals}. In the formula, clauses are connected by Boolean `AND', where each clause contains \emph{literals} connected by Boolean `OR'. Informally, a tumor cell is malignant if at least one literal in each clause becomes true.
\end{example}

The problem of learning rule-based classifiers is known to be computationally intractable. The earliest tractable approaches for classifiers such as decision trees and decision lists relied on heuristically chosen objective functions and greedy algorithmic techniques~\shortcite{ClarkN1989,CohenS1999,quinlan2014}. In these approaches, the size of rules is controlled by early stopping,  ad-hoc rule pruning, etc. In recent approaches, the interpretable classification problem is reduced to an optimization problem, where the accuracy and the sparsity of rules are optimized jointly~\shortcite{lakkaraju2016interpretable,narodytska2018learning}. Different optimization solvers such as linear programming~\shortcite{malioutov2013exact}, sub-modular optimizations~\shortcite{lakkaraju2016interpretable}, and Bayesian optimizations~\shortcite{letham2015interpretable} are then deployed to find the best classifier with maximum accuracy and minimum rule-size. In our study, we propose an alternate optimization approach that fits particularly well to rule-learning problems. Particularly,  we propose a \textit{maximum satisfiability} (MaxSAT)  solution for learning interpretable rule-based classifiers. 

The MaxSAT problem is an optimization analog to the satisfiability (SAT) problem, which is complete for the class $ \mathrm{FP}^{\mathrm{NP}} $. Although the MaxSAT problem is NP-hard, dramatic progress has been made in designing solvers that can handle large-scale problems arising in practice. This has encouraged researchers to reduce several optimization problems into MaxSAT such as  optimal planning~\shortcite{robinson2010partial}, automotive configuration~\shortcite{walter2013applications}, data analysis in machine learning~\shortcite{berg2019applications}, and automatic test pattern generation for cancer therapy~\shortcite{lin2012application}. 

\paragraph{Contributions.}The primary contribution of this paper is a MaxSAT-based formulation, called {\framework}, for learning interpretable classification rules expressible in propositional logic. For the simplicity of exposition, our initial focus is on learning classifiers expressible in CNF (Conjunctive Normal Form) formulas; we later discuss how the CNF learning formulation can be extended to popular interpretable classifiers such as decision lists and decision sets. In this paper, {\framework} provides precise control to jointly optimize the accuracy and the size of classification rules. {\framework} constructs and solves a MaxSAT query of $ \mathcal{O}(kn) $  size (number of clauses) to learn a $ k $-clause CNF formula on a dataset containing $ n $ samples. Naturally, the na\"ive formulation cannot scale to large values of $ n $ and $ k $. To scale {\framework} to large datasets, we propose an incremental learning technique along with the MaxSAT formulation. Our incremental learning is an integration of mini-batch learning and iterative rule-learning, which are studied separately in classical learning problems. In the presented incremental approach, {\framework} learns a $ k $-clause CNF formula using an iterative separate-and-conquer algorithm, where in each iteration, a single clause is learned by covering a part of the training data. Furthermore, to efficiently learn a single clause in a CNF, {\framework} relies on mini-batch learning, where it solves a sequence of smaller MaxSAT queries corresponding to each mini-batch. 

In our experimental evaluations, {\framework} demonstrates the best balance among prediction accuracy, interpretability, and scalability in learning classification rules. In particular, {\framework} achieves competitive prediction accuracy and interpretability w.r.t.\ state-of-the-art interpretable classifiers. Besides, {\framework} achieves impressive scalability by classifying datasets even with $ 1 $ million samples, wherein existing classifiers, including those of the non-interpretable ones, either fail to scale or achieve poor prediction accuracy. Finally, as an application, we deploy {\framework} in learning interpretable classifiers such as decision lists and decision sets.

The presented framework {\framework} extends our prior papers ~\shortcite{ghosh19imli,MM18} in the following aspects. \citeauthor{MM18}~\citeyear{MM18} presented a MaxSAT-based formulation for interpretable classification rule-learning, which jointly optimizes between the accuracy and size of rules.  \citeauthor{ghosh19imli}~\citeyear{ghosh19imli} improved the MaxSAT formulation by unifying mini-batch learning. This unification allows learning on moderate-size datasets (about $ 50 $ thousand samples). This paper takes a significant step in terms of scalability by integrating both iterative learning and mini-batch learning inside the incremental technique. The result is a practical scalable learning framework that can classify datasets with $ 1 $ million samples and can potentially scale beyond that.

We organize the rest of the paper as follows. We discuss related literature in Section~\ref{sec:related} and preliminaries in Section~\ref{sec:preliminaries}. In Section~\ref{sec:problem}, we formally define our learning problem and provide a MaxSAT-based solution in Section~\ref{sec:baseline}. We improve the MaxSAT-based formulation through an incremental learning technique in Section~\ref{sec:incremental_learning}. In Section~\ref{sec:application}, we apply {\framework} in learning different interpretable classifiers. We then discuss our experimental results in Section~\ref{sec:experiments} and conclude in Section~\ref{sec:conclusion}.

\section{Related Work}
\label{sec:related}

The progress in designing interpretable rule-based classifiers finds its root in the development of decision trees \shortcite{bessiere2009minimising,quinlan1986induction,quinlan1987simplifying}, decision lists \shortcite{rivest1987learning}, classification rules \shortcite{cohen1995fast} etc.  In early works, the focus was to improve the efficiency and scalability of the model rather than designing models that are interpretable. For example,  decision rule approaches such as  C4.5 rules \shortcite{quinlan2014}, CN2~\shortcite{ClarkN1989}, RIPPER \shortcite{cohen1995fast}, and SLIPPER  \shortcite{CohenS1999} 
rely on heuristic-based branch pruning and ad-hoc local criteria e.g., maximizing information gain, coverage, etc.

Recently, several optimization frameworks have been proposed for interpretable classification, where both accuracy and rule-size are optimized during training. For example,~\citeauthor{malioutov2013exact}~\citeyear{malioutov2013exact}  proposed exact learning of rule-based classifiers based on Boolean compressed sensing using a linear programming formulation.~\citeauthor{su2016learning}~\citeyear{su2016learning} presented two-level Boolean rules, where the trade-off between classification accuracy and interpretability is studied. In their work, the Hamming loss is used to characterize prediction accuracy, and sparsity is used to characterize the interpretability of rules.~\citeauthor{wang2015falling}~\citeyear{wang2015falling} proposed a Bayesian optimization framework for learning falling rule lists, which is an ordered list of if-then rules. Other similar approaches based on Bayesian analysis for learning classification rules are~\shortcite{letham2015interpretable,wang2017bayesian}. Building on custom discrete optimization techniques,~\citeauthor{angelino2017learning}~\citeyear{angelino2017learning} proposed an optimal learning technique for decision lists using a branch-and-bound algorithm. In a separate study,~\citeauthor{lakkaraju2016interpretable}~\citeyear{lakkaraju2016interpretable} highlighted the importance of decision sets over decision lists in terms of interpretability and considered a sub-modular optimization problem for learning a near-optimal solution for decision sets. Our proposed method for interpretable classification, however, relies on the improvement in formal methods over the decades, particularly the efficient CDCL-based solution for satisfiability (SAT) problems~\shortcite{silva2003grasp}.

Formal methods, particularly SAT and its variants, have been deployed in interpretable classification recently. In the context of learning decision trees, SAT and MaxSAT-based solutions are proposed by~\shortcite{alos2021learning,janota2020sat,narodytska2018learning,shati2021sat}. In addition, researchers have applied SAT for learning explainable decision sets~\shortcite{ignatiev2018sat,ignatiev2021scalable,schidler2021sat,yu2020computing}. In most cases, SAT/MaxSAT solutions are not sufficient in solving large-scale classification tasks because of the $ \mathrm{NP} $-hardness of the underlying problem. This observation motivates us in combining MaxSAT with more practical algorithms such as incremental learning.

Incremental learning has been studied in improving the scalability of learning problems, where data is processed in parts and results are combined to use lower computation overhead. In case of non-interpretable classifiers such as SVM, several solutions adopting incremental learning are available~\shortcite{syed1999incremental,ruping2001incremental}. For example,~\citeauthor{cauwenberghs2001incremental}~\citeyear{cauwenberghs2001incremental} proposed an online recursive algorithm for SVM that learns one support-vector at a time. Based on radial basis kernel function,~\citeauthor{ralaivola2001incremental}~\citeyear{ralaivola2001incremental} proposed a local incremental learning algorithm for SVM. In the context of deep neural networks, stochastic gradient descent is a well-known convex optimization technique\textemdash a variant of which includes computing the gradient on mini-batches~\shortcite{hinton2012neural,li2014efficient,masters2018revisiting}.  Federated learning, on the other hand, decentralizes training on multiple local nodes based on local data samples with only exchanging learned parameters to construct a global model in a central node~\shortcite{konevcny2015federated,konevcny2016federated}. Another notable technique is Lagrangian relaxation that decomposes the original problem into several sub-problems, assigns Lagrangian multipliers to make sure that sub-problems agree, and iterates by solving sub-problems and adjusting weights based on disagreements~\shortcite{fisher1981lagrangian,johnson2007lagrangian,lemarechal2001lagrangian}. To the best of our knowledge, our method is the first method that unifies incremental learning with MaxSAT based formulation to improve the scalability of learning rule-based classifiers.

Classifiers that are interpretable by design can be applied to improve the explainability of complex black-box machine learning classifiers. There is rich literature on extracting decompositional and pedagogical rules from non-linear classifiers such as support vector machines~\shortcite{barakat2004learning,barakat2005eclectic,diederich2008rule,martens2008rule,nunez2002rule} and neural networks~\shortcite{augasta2012rule,hailesilassie2016rule,setiono1995understanding,sato2001rule,zilke2016deepred,zhou2004rule}. In recent years, local model-agnostic approaches for explaining black-box classifiers  are  proposed by learning surrogate simpler classifiers such as rule-based classifiers~\shortcite{guidotti2018local,pastor2019explaining,rajapaksha2020lormika,ribeiro2018anchors}. The core idea in local approaches is to use a rule-learner that can classify synthetically generated neighboring samples with class labels provided by the black-box classifier. To this end, our framework {\framework} can be directly deployed as an efficient rule-learner and can explain the inner-working of black-box classifiers by generating interpretable rules.

\section{Preliminaries}
\label{sec:preliminaries}

In this section, we first discuss propositional satisfiability and MaxSAT, followed by rule-based classification in machine learning. In the paper, we use capital  boldface letters such as $\mathbf{X}$ to denote matrices while lower boldface letters $\mathbf{x}$ are reserved for vectors/sets. 

\subsection{Propositional Satisfiability}

Let $F$ be a Boolean formula defined over Boolean variables $\mathbf{b} = \{b_1,b_2,\dots ,b_m \}$. A literal $ v $ is a variable $b$ or its complement $\neg b$, and a clause $ C $ is a disjunction ($ \vee $) or a conjunction ($ \wedge $) of literals.\footnote{While \emph{term} is reserved to denote disjunction of literals in the literature, we use clause for both disjunction and conjunction.}  $F$ is in Conjunctive Normal Form (CNF) if $F \triangleq  \bigwedge_i C_i$ is a conjunction of clauses where each clause $C_i \triangleq  \bigvee_j v_j $  is a disjunction of literals. In contrast, $ F $ is in Disjunctive Normal Form (DNF) if $F \triangleq  \bigvee_i C_i$ is a disjunction of clauses where each clause $C_i \triangleq  \bigwedge_j v_j $ is a conjunction of literals. We use $\sigma$ to denote an assignment of variables in  $\mathbf{b}$ where $ \sigma(b_i) \in $ \{true, false\}. A \emph{satisfying assignment} $ \sigma^* $ of $F$ is an assignment  that evaluates $F$  to true and is denoted by $ \sigma^* \models F $. The propositional satisfiability (SAT) problem finds a satisfying assignment $ \sigma^* $ to a CNF formula $ F $  such that $ \forall i, \sigma^* \models C_i $, wherein $ \sigma^*\models C_i $ if and only if $ \exists v_j \in C_i, \sigma^*(v_j) = \text{true} $. Informally, $ \sigma^* $ satisfies at least one literal in each clause of a CNF formula. 

In this paper, we use true as $ 1 $ and false as $ 0 $ interchangeably. Between two vectors  $\mathbf{u}$ and $\mathbf{v}$ of the same length defined over Boolean variables or constants (such as  $ 0 $ and $ 1 $), we define $\mathbf{u} \circ \mathbf{v} $  to refer to  their inner product. Formally, $\mathbf{u} \circ \mathbf{v} \triangleq \bigvee_{i} (u_{i} \wedge v_{i})$ is a disjunction of element-wise conjunction where $u_{i}$ and $v_{i}$ denote the $ i^\text{th} $ variable/constant of $\mathbf{u}$ and $\mathbf{v}$, respectively. In this context, the conjunction ``$\wedge$'' between a variable and a constant follows the standard interpretation: $b \wedge 0 = 0$ and $b \wedge 1 = b$.  For example, if $ \mathbf{u} = [u_1, u_2, u_3] $ and $ \mathbf{v} = [0,1,1] $, then $ \mathbf{u} \circ \mathbf{v} = u_2 \vee u_3 $.

\subsection{MaxSAT}

The MaxSAT problem is an optimization analog to the SAT problem that finds an optimal assignment satisfying the maximum number of clauses in a CNF. In this work, we consider a \emph{weighted} variant of the MaxSAT problem\textemdash more specifically, a weighted-partial MaxSAT problem\textemdash that optimizes over a set of hard and soft constraints in the form of a weighted-CNF formula. In a weighted-CNF formula, a weight $ {\weight{C_i}} \in \mathbb{R}^{\ge 0} \cup \{\infty\} $ is defined over each clause $ C_i $, wherein $C_i$ is called a \emph{hard} clause if $\weight{C_i} = \infty$, and  $C_i$ is  a \emph{soft} clause, otherwise.  To avoid notational clutter, we overload {\weight{\cdot}} to denote the weight of an assignment $ \sigma $. In particular, we define $\weight{\sigma}$ as the sum of the weights of \emph{unsatisfied clauses} in a CNF, formally, $\weight{\sigma} = \Sigma_{i | \sigma \not \models C_i} \weight{C_i}$.

Given a weighted-CNF formula $F \triangleq \bigvee_i C_i$ with weight $ \weight{C_i} $, the weighted-partial MaxSAT problems finds an optimal assignment $\sigma^*$ that achieves the minimum weight. Formally,  $\sigma^* = \MaxSAT(F,\weight{\cdot})$ if $\forall \sigma \neq \sigma^*, \weight{\sigma^*} \leq \weight{\sigma}$. The optimal weight of the MaxSAT problem $  \weight{\sigma^*} $ is infinity ($ \infty $) when at least one hard clause becomes unsatisfied.  Therefore,  the weighted-partial MaxSAT problem finds $ \sigma^* $ that satisfies all hard clauses and as many soft clauses as possible such that the total weight of unsatisfied soft clauses is minimum. In this paper, we reduce the classification problem  to the solution of a weighted-partial MaxSAT problem. The knowledge of the inner workings of {MaxSAT} solvers is  not required for this paper.

\subsection{Rule-based Classification}

We consider a binary classification problem where $ \mathbf{X} \in \{0,1\}^{n\times m} $ is a binary feature matrix\footnote{Our framework allows learning on real-valued and categorical features, as discussed in Section~\ref{sec:non-binary}.} defined over Boolean features $\mathbf{x} = \{x_1, x_2, \dots, x_m\}$ and $ \mathbf{y} \in \{0,1\}^n $ is a vector of binary class labels.  The $ i^\text{th} $ row of $ \mathbf{X} $, denoted by $ \mathbf{X}_i \in \mathcal{X} $, is called a sample, which is generated from a product distribution $ \mathcal{X} $. Thus, $ \mathbf{X}_i $ is the valuation of features in $ \mathbf{x} $ and $ y_i $ is its class label.  $ \mathbf{X}_i $ is called a positive sample if $ y_i = 1 $, otherwise a negative sample.

A classifier $ \Rule: \mathbf{x} \rightarrow \hat{y} \in  \{0,1\} $ is a  function that takes a Boolean vector as input and outputs the predicted class label $\hat{y}$.  
In a rule-based classifier, we view $ \mathcal{R} $ as a propositional formula defined over Boolean features $ \mathbf{x} $ such that, if $ \Rule $ becomes true for an input (or an assignment), the prediction $ \hat{y} = 1 $ and $ \hat{y} = 0 $, otherwise.  The goal of designing $\Rule$ is to not only approximate the training data, but also generalize to unseen samples arising from the same distribution. Additionally, we prefer to learn a \emph{sparse} $ \Rule $ in order to favor interpretability. We define the structural complexity of $ \Rule $, referred to as \emph{rule-size}, in terms of the number of literals it contains. Let $\clause(\Rule,i)$ denote  the  $i^\text{th}$ clause of $\Rule$ and $ |\clause(\Rule,i)| $ be  the number of literals in $\clause(\Rule,i)$. Thus, the size of the classifier is $|\Rule| =  \Sigma_i |\clause(\Rule,i)| $, which is the sum of literals in all clauses in $ \Rule $. 

For example, let $ \Rule \triangleq (x_1 \vee x_3) \wedge (\neg x_2 \vee x_3) $ be a CNF classifier defined over three Boolean features $ \{x_1, x_2, x_3\} $. For an input $ [0, 1, 1] $, the prediction of $ \Rule  $ is $ 1 $ whereas for an input $ [1, 1, 0] $, the prediction is $ 0 $ because $ \Rule $ is true and false in these two cases, respectively. Moreover, $ |\Rule| = 2 + 2 = 4 $ is the rule-size of $ \Rule $. 

\paragraph{Decision Lists.} A decision list is a rule-based classifier consisting of ``if-then-else'' rules. Formally, a decision list~\shortcite{rivest1987learning} is an ordered list $ \Rule_L $ of pairs $ (C_1, v_1), \dots, (C_k, v_k) $ where $ C_i $ is a conjunction of literals (alternately, a single clause in a DNF formula) and $ v_i \in \{0,1\} $ is  a Boolean class label\footnote{In a more practical setting, $ v_i \in \{0, \dots, N\} $  can be multi-class for $ N \ge 1 $.}. Additionally, the last clause $ C_k \triangleq  $ true, thereby $ v_k $ is the default class. A decision list is defined as a classifier with the following interpretation: for a feature vector $ \mathbf{x} $,  $ \Rule_L(\mathbf{x}) $ is equal to $ v_i $ where $ i $ is the least index such that $ \mathbf{x} \models C_i $. Since the last clause is true, $ \Rule_L(\mathbf{x}) $ always exists. Intuitively, whichever clause (starting from the first) in $ \Rule_L $ is satisfied for an input, the associated class label is  its prediction.

\paragraph{Decision Sets.} A decision set is a set of \textit{independent} ``if-then'' rules. Formally,  a decision set $ \Rule_S $ is a set of pairs $ \{(C_1, v_1), \dots, (C_{k-1}, v_{k-1})\}  $ and a  default pair $ (C_k, v_k) $, where $ C_i $ is\textemdash similar to a decision list\textemdash a conjunction of literals and $ v_i \in \{0,1\} $ is a Boolean class label. In addition, the last clause $ C_k \triangleq  $ true and $ v_k $ is the default class. For a  decision set, if an input $ \mathbf{x} $ satisfies one clause, say $ C_i $, then the prediction is $ v_i $. If $ \mathbf{x} $ satisfies no clause, then the prediction is the default class $ v_k $. Finally, if $ \mathbf{x} $ satisfies  $ \ge 2 $ clauses, $ \mathbf{x} $ is assigned a class using a tie-breaking~\shortcite{lakkaraju2016interpretable}.

 \section{Problem Formulation}
 \label{sec:problem}
Given (i) a binary feature matrix $ \mathbf{X} \in \{0,1\}^{n \times m} $ with $ n $ samples and $ m $ features, (ii) a class-label vector $ \mathbf{y} \in \{0,1\}^n $, (iii) a positive integer $ k \ge 1 $ denoting the number of clauses, and (iv) a regularization parameter $ \lambda \in \mathbb{R}^+ $, we learn a classifier $ \Rule $ represented as a $ k $-clause CNF formula to separate samples of class $ 1 $ from class $ 0 $.

 Our goal is to learn classifiers that balance two 
 goals: of  being accurate but also interpretable.  
 Various notions of interpretability have been proposed in the context of   classification problems. A common proxy for interpretability in the context of decision rules 
 is the sparsity of rule. For instance, a rule involving fewer literals is highly interpretable.  In this work, we minimize the total number of literals in all clauses, as discussed in Section~\ref{sec:preliminaries}, which motivates us to  find $ \Rule  $ with minimum  $ |\Rule| $. Let $ \Rule $ classify all samples correctly during training. Among all the classification rules that classify all samples correctly,  we choose the sparsest (most interpretable) such $ \Rule $.

 \[
 \min\limits_{\mathcal{R}} |\mathcal{R}|\text{ such that }\forall i, y_i=\mathcal{R}(\mathbf{X}_i)
 \]

 In practical classification tasks, perfect classification is unlikely. Hence, we need to balance interpretability with classification error.  Let $ \mathcal{E}_\mathbf{X} = \{(\mathbf{X}_i,y_i) | y_i \ne \mathcal{R}(\mathbf{X}_i) \} $   be  the set of samples in $ \mathbf{X} $ that are misclassified  by $ \mathcal{R} $. Therefore, we balance between classification-accuracy and rule-sparsity and optimize the following function.\footnote{In our formulation, it is  straightforward to add class-conditional weights  (e.g., to penalize  false-alarms more than mis-detects), and to allow instance weights (per sample).}

 \begin{equation}
  \label{eq:obj}
 \min\limits_{\mathcal{R}} \;  |\mathcal{E}_\mathbf{X}| +\lambda |\mathcal{R}|
 \end{equation}
 
Higher  values of $ \lambda $ generate a rule with a smaller rule-size but of more training errors, and vice-versa. Thus, $ \lambda $ can be tuned to trade-off between accuracy and interpretability for a rule-based classifier, which we experiment extensively in Section~\ref{sec:experiments}.

\section{MaxSAT-based Interpretable Classification Rule Learning}
\label{sec:baseline}
In this section, we discuss a MaxSAT-based framework for optimal learning of an interpretable rule-based classifier, particularly a CNF classifier $ \Rule $. 
We first describe the decision variables in Section~\ref{sec:variables} and present the MaxSAT encoding  in Section~\ref{sec:encoding}. Our MaxSAT formulation assumes binary features as input. We
conclude this section by learning $ \Rule $ with non-binary features in Section~\ref{sec:non-binary} and discussing more flexible interpretability constraints of $ \Rule $ in Section~\ref{sec:complex_interpretability_objectives}.

\subsection{Description of Variables}
\label{sec:variables} 
We initially preprocess $ \mathbf{X} $  to account for the negation of  Boolean features while learning a classifier.\footnote{This preprocessing is similarly applied in  \shortcite{malioutov2013exact}.} In the preprocessing step, we negate each column in $ \mathbf{X} $ to a new column and append it to $ \mathbf{X} $. For example, if ``$ \text{age }\ge 25 $'' is a Boolean feature, we add another feature ``$ \text{age }< 25 $'' in  $ \mathbf{X} $ by negating the column ``$ \text{age }\ge 25 $''. Hence, in the rest of the paper, we refer $ m $ as the modified number of columns in $ \mathbf{X} $. We next discuss the variables in the MaxSAT problem. 

We consider two types of Boolean variables: (i) \emph{feature} variables $ b $ corresponding to input features and (ii) \emph{error} variables $ \eta $ corresponding to the classification error of samples. We define a Boolean variable $ b^i_j $ that becomes true if feature $ x_j $ appears in the $ i^\text{th} $ clause of $ \Rule $, thereby contributing to an increase in the rule-size of $ \Rule $, and $ b^i_j $ is assigned false, otherwise. Moreover, we define an error variable $ \eta_l $ to attribute to whether the sample $ \mathbf{X}_l $ is classified correctly or not. Specifically, $ \eta_l $ becomes true if $ \mathbf{X}_l $ is misclassified, and becomes false otherwise.  We next discuss the MaxSAT encoding to solve the classification problem.

\subsection{MaxSAT Encoding}
\label{sec:encoding}
We consider a weighted-partial MaxSAT formula, where we encode the objective function in Eq.~\eqref{eq:obj} as soft clauses. For each sample in $ \mathbf{X} $, we construct hard clauses specifying that the sample is either  classified correctly and $ \eta_l $ is false or it is misclassified and $ \eta_l $ is true. We next discuss the MaxSAT encoding in detail.

\begin{itemize}
	\item \textbf{Soft clauses for maximizing training accuracy:} For each training sample, {\framework} constructs a soft unit\footnote{A unit clause has a single literal.} clause $ \neg \eta_l $ to account for a penalty for misclassification. Since the penalty for misclassification of a sample is $ 1 $  in Eq.~\eqref{eq:obj},  the weight of this soft clause is also $ 1 $. 
	
	\begin{equation}
		\label{soft_clause_error}
		E_l:=\neg \eta_l;  \qquad \mathsf{wt}(E_l)= 1
	\end{equation}
	Intuitively, if a sample is misclassified, the associated error variable becomes true, thereby dissatisfying the soft clause $ E_l $.

	\item \textbf{Soft clauses for minimizing rule-sparsity:} To favor rule-sparsity, {\framework} tries to learn a classifier with as few literals as possible. Hence, for each feature variable $ b_j^i $, {\framework} constructs a unit clause as $ \neg b_j^i $. Similar to training accuracy, the weight for this clause is derived as $ \lambda $ from Eq.~\eqref{eq:obj}.
	\begin{equation}
		\label{soft_clause_feature}
		S_j^i:= \neg b_j^i; \qquad \mathsf{wt}(S_j^i)= \lambda
	\end{equation}

	\item \textbf{Hard clauses for encoding constraints:} In a MaxSAT problem, constraints that must be satisfied are encoded as hard clauses. In {\framework}, we have a learning constraint that, if the error variable is false, the associated sample must be correctly classified, and vice-versa. Let  $\mathbf{b}_i= \{b^i_{j} \mid j \in \{1,\dots,m \}\}$ be a vector of feature variables corresponding to the $ i^\text{th} $  clause in $ \mathcal{R} $. Then, we define the following hard clause.
	
	\begin{equation}
			\label{hard_clause_rule_consistency}
		H_l:= \neg \eta_l \rightarrow \Big(y_l\leftrightarrow \bigwedge_{i=1}^k {\mathbf{X}_l} \circ {\mathbf{b}_{i}}\Big); \qquad  \mathsf{wt}(H_l)=\infty
	\end{equation}
	
	In the hard clause, $ \bigwedge_{i=1}^k {\mathbf{X}_l} \circ {\mathbf{b}_{i}}\ $ is a CNF formula including variables $ b_j^i $ for which the associated  feature-value is $ 1 $ in  $ \mathbf{X}_l $. Since $ y_l \in \{0,1\} $ is a constant, the constraint to the right of the implication ``$ \rightarrow $'' is either  $ \bigwedge_{i=1}^k {\mathbf{X}_l} \circ {\mathbf{b}_{i}}\ $ or its complement. Therefore, the hard clause enforces that if the sample is correctly classified (using $ \neg \eta_l $), either  $ \bigwedge_{i=1}^k {\mathbf{X}_l} \circ {\mathbf{b}_{i}}\ $ or its complement is true depending on the class-label of the sample.  We highlight that the single-implication ``$ \rightarrow $'' in the hard clause $ H_l $ acts as a double-implication ``$ \leftrightarrow $''  due to the soft clause $ E_l $.  Because, according to the definition of ``$ \rightarrow $'', the left constraint $ \neg \eta_l $ can  be false while the right constraint of 	``$ \rightarrow $'' is true.  This, however, incurs unnecessarily dissatisfying the soft clause $ E_l $, which is a sub-optimal solution and hence this solution is not returned by the MaxSAT solver.

\end{itemize}

	  We next discuss the translation of soft and hard clauses into a CNF formula, which can be invoked by any MaxSAT solver. 

\paragraph{Translating $ E_l, S^i_j, H_l $ to a CNF Formula.} 
The soft clauses $ E_l $ and $ S_j^i $ are unit clauses and hence, no translation is required for them.  In the hard clause, when $ y_l=1 $, the simplification is $ H_l:= \neg \eta_l \rightarrow \bigwedge_{i=1}^k {\mathbf{X}_l} \circ {\mathbf{b}_{i}}  $. In this case, we apply the equivalence rule in propositional logic $ (a \rightarrow b) \equiv (\neg a \vee b) $  to encode $ H_l $ into  CNF. In contrast, when $ y_i=0 $, we simplify the hard clause as $ H_l:= \neg \eta_l \rightarrow \neg (\bigwedge_{i=1}^k {\mathbf{X}_l} \circ {\mathbf{b}_{i}}) \Rightarrow  \neg \eta_l \rightarrow \bigvee_{i=1}^k \neg({\mathbf{X}_l} \circ {\mathbf{b}_{i}})$. Since $ \mathbf{X}_l \circ \mathbf{b}_{i} $ constitutes a disjunction of literals (defined in Section~\ref{sec:preliminaries}), we apply Tseytin transformation to encode  $ \neg (\mathbf{X}_l \circ \mathbf{b}_{i}) $ into CNF. More specifically, we introduce an auxiliary variable $ z_{l,i} $ corresponding to the clause $ \neg ({\mathbf{X}_l} \circ {\mathbf{b}_{i}}) $. Formally, we replace $ H_l := \neg \eta_l \rightarrow  \bigvee_{i=1}^k \neg ({\mathbf{X}_l} \circ {\mathbf{b}_{i}}) $ with   $ \bigwedge_{i=0}^k H_{l,i} $,  where $ H_{l,0}:= (\neg \eta_l \rightarrow  \bigvee_{i=1}^k  z_{l,i}) $  and  $  H_{l,i}:= z_{l,i} \rightarrow  \neg({\mathbf{X}_l} \circ {\mathbf{b}_{i}})  $ for $ i=\{1,\dots,k\} $. Finally, we  apply the equivalence of $ (a \rightarrow b) \equiv (\neg a \vee b) $ on $ H_{l,i} $ to translate them into CNF. For either value of $ y_l $, the weight of each translated hard clause in the CNF formula is $ \infty $.



Once we translate  all soft and hard clauses into CNF, the MaxSAT query $ Q $ is the conjunction of all clauses. 
\[
Q:= \bigwedge_{l=1}^n E_l \wedge  \bigwedge_{i=1,j=1}^{i=k,j=m} S_j^i \wedge \bigwedge_{l=1}^n H_l 
\]
		
Any off-the-shelf MaxSAT solver can output an optimal assignment $ \sigma^* $ of the MaxSAT query $ (Q, \weight{\cdot}) $. We extract $ \sigma^* $ to construct the classifier $ \Rule $ and compute training errors as follows.
\begin{construction}
	\label{construction:rule}
	Let $\sigma^* = \MaxSAT(Q,\weight{\cdot})$. Then $x_j \in {\clause(\Rule,i)}$ if and only if $\sigma^*(b_{j}^{i}) = 1$. Additionally, $ \mathbf{X}_l $ is misclassified if and only if $ \sigma^*(\eta_l) = 1 $.
\end{construction}

\paragraph{Complexity of the MaxSAT Query.}

We analyze the complexity of the MaxSAT query in terms of the number of Boolean variables and clauses in the CNF formula $ Q $. 

\begin{prop}
	\label{prop:maxsat_variables}
	 To learn a $ k $-clause CNF classifier for a dataset of  $ n $ samples over $ m $ boolean features, the MaxSAT query $ Q $ defines $ km +n $ Boolean variables. Let $ n_{neg} $ be the number of negative samples in the training dataset. Then the number of auxiliary variables in $ Q $ is  $ kn_{neg}  $.
\end{prop}
\begin{prop}
	\label{prop:maxsat_clauses}
	The MaxSAT query $ Q $ has $ k m+n $ unit clauses corresponding to the constraints $ E_l $ and $ S_j^i $. For each positive sample, the hard clause $ H_l $ is translated into $ k $  clauses. For each negative sample, the CNF translation requires at most $ k m+1 $ clauses. Let $ n_{pos} $ and $ n_{neg} $ be the number of positive and negative samples in the training dataset. Therefore, the number of clauses in the MaxSAT Query $ Q $ is $ k m+n+k n_{pos}+(k m+1)n_{neg} \approx \mathcal{O}(k m  n ) $ when $ n_{pos}= n_{neg} =\frac{n}{2} $. 
\end{prop}

In the following, we illustrate the encoding for a toy example.
\subsubsection*{Example of encoding}

We illustrate the MaxSAT encoding for a toy example consisting of four samples and two binary features. Our goal is to learn a two clause CNF classifier that can approximate the training data. 
\[ 
\mathbf{X}_\mathsf{orig} = 
\begin{blockarray}{cc}
x_1 & x_2 \\
\begin{block}{[cc]}
0 & 0 \\
0 & 1 \\
1 & 0 \\
1 & 1 \\
\end{block}
\end{blockarray} \implies 
\mathbf{X} = 
\begin{blockarray}{cccc}
x_1 & \neg x_1 & x_2  & \neg x_2\\
\begin{block}{[cccc]}
0 & 1 & 0 & 1\\
0 & 1 & 1 & 0\\
1 & 0 & 0 & 1\\
1 & 0 & 1 & 0\\
\end{block}
\end{blockarray}; \quad 
\mathbf{y} = 
\begin{bmatrix} 
1 \\ 0 \\ 0 \\ 1
\end{bmatrix}
\]
In the preprocessing step, we negate the columns in $ \mathbf{X}_\mathsf{orig} $ and add complemented features $ \{\neg x_1, \neg x_2\} $ in $ \mathbf{X} $. In the MaxSAT encoding, we define $ 8 $ feature variables ($ 4 $ features $ \times $ $ 2 $ clauses in the classifier) and $ 4 $ error variables. For example, for feature $ x_2 $, introduced feature variables are $ \{b^1_3, b^2_3\} $ and for feature $ \neg x_2 $, introduced variables are $ \{b^1_4, b^2_4\} $. For four samples, error variables are $ \{\eta_1, \eta_2, \eta_3, \eta_4\} $. We next show the soft and hard clauses in the MaxSAT encoding

\[
\begin{split}
&E_1:= (\neg \eta_1); \quad
E_2:= (\neg \eta_2); \quad E_3:= (\neg \eta_3); \quad
E_4:= (\neg \eta_4)\\
& S_1^1 := (\neg b_1^1);\quad 
 S^1_2 := (\neg b^1_2);\quad 
 S^1_3 := (\neg b^1_3);\quad
 S^1_4 := (\neg b^1_4);\quad \\ 
& S_1^2 := (\neg b_1^2);\quad 
S^2_2 := (\neg b^2_2);\quad 
S^2_3 := (\neg b^2_3);\quad 
S^2_4 := (\neg b^2_4);\quad\\ 
& H_1:= (\neg \eta_1 \rightarrow ((b_2^1 \vee b_4^1)\wedge (b_2^2 \vee b_4^2)));\\
& H_2:= (\neg \eta_2 \rightarrow (\neg(b_2^1 \vee b_3^1) \vee \neg(b_2^2 \vee b_3^2)));\\
& H_3:= (\neg \eta_3 \rightarrow (\neg (b_1^1 \vee b_4^1) \vee \neg(b_1^2 \vee b_4^2)));\\
& H_4:= (\neg \eta_4 \rightarrow ((b_1^1 \vee b_3^1) \wedge (b_1^2 \vee b_3^2)));\\
\end{split}
\]

In this example, we consider regularizer $ \lambda = 0.1 $, thereby setting the weight on accuracy as $ \weight{E_i} = 1 $ and rule-sparsity weight as $ \weight{S_j^i} = 0.1 $. Intuitively, the penalty for misclassifying a sample is $ 10 $ times than the penalty for adding a feature in the classifier. For this MaxSAT query, the optimal solution classifies all samples correctly by assigning four feature variables $ \{b^1_1, b^1_4, b^2_2, b^2_3\} $ to true. Therefore, by applying Construction~\ref{construction:rule}, the classifier is $ (x_1 \vee \neg x_2)  \wedge (\neg x_1 \vee x_2) $.\footnote{The presented MaxSAT-based formulation does not learn a CNF classifier with both $ x_i $ and $ \neg x_i $ in the same clause. The reason is that a clause with both $ x_i $ and $ \neg x_i $ connected by OR $ (\vee) $ does not increase accuracy but increases rule-size and hence, this is not an optimal classifier.}

\subsection{Learning with Non-binary Features}
	\label{sec:non-binary}
	The presented MaxSAT encoding requires input samples to have binary features. Therefore, we discretize datasets with categorical and continuous features into binary features. For each continuous feature, we apply equal-width discretization that splits the feature into a fixed number of bins. For example, consider a continuous feature $ x_c \in [a,b] $. In discretization, we split the domain $ [a,b] $ into three bins with two split points $ \{a',b'\} $ such that $ a<a'<b'<b $.  Therefore, the resulting three discretized features are $ { a \le x_c < a'} $, $ {a' \le x_c < b'} $, and $ { b' \le x_c  \le b } $. An alternate to this close-interval discretization is open-interval discretization, where we consider six discretized features with each feature being compared with one threshold. In that case, the discretized features are: $ x_c \ge a, x_c \ge a', x_c \ge b', x_c < a', x_c < b', x_c \le b $. Both open-interval and close-interval discretization techniques have their use-cases where one or the other is appropriate. For simplicity, we experiment with close-interval discretization in this paper.
	
	After applying discretization on continuous features, the dataset contains categorical features only, which we convert to binary features using one-hot encoding~\shortcite{lakkaraju2019faithful,ghosh19imli}. In one-hot encoding, a Boolean vector of features is introduced with cardinality equal to the number of distinct categories. For example, consider a categorical feature having three categories `red', `green', and `yellow'. In one hot encoding, samples with category-value `red', `green', and `yellow' would be converted into binary features by taking values $ 100 $, $ 010 $, and $ 001 $, respectively.

\subsection{Flexible Interpretability Objectives}
\label{sec:complex_interpretability_objectives}
{\framework} can be extended to consider more flexible interpretability objectives than the simplified one in Eq.~\eqref{eq:obj}. In Eq.~\eqref{eq:obj}, we prioritize all features equally by providing the same weight to the clause $ S_j^i $ for all  values of $ i $ and $ j $. In practice, users may prefer rules containing certain features. In {\framework}, such an extension can be achieved by modifying the weight function and/or the definition of  $ S_j^i $. For example, to constrain the classifier to never include a feature, the weight of the clause $ S_j^i := \neg b_j^i $ can be set to $ \infty $. In contrast, to always include a feature, we can define $ S_j^i := b_j^i $ with weight $ \infty $. In both cases, we treat $ S_j^i $ as a hard clause. 

Another use case may be to learn rules where clauses have disjoint set of features. To this end, we consider a pseudo-Boolean constraint $ \sum_{i=1}^k b^i_j \le 1 $, which specifies that the feature $ x_j $ appears in at-most one of the $ k $ clauses. This constraint may be soft or hard depending on the priority in the application domain. In either case, we convert this constraint to CNF using pseudo-Boolean to CNF translation~\shortcite{philipp2015pblib}. Thus, {\framework} allows us to consider varied interpretability constraints by only modifying the MaxSAT query without requiring changes in the MaxSAT solving. Therefore, the \emph{separation between modeling and solving} turns out to be the key strength of {\framework}.

\section{Incremental Learning with MaxSAT-based Formulation}
\label{sec:incremental_learning}
The complexity of the MaxSAT query in Section~\ref{sec:encoding} increases with the number of samples in the training dataset and the number of clauses to be learned in a CNF formula. In this section, we discuss an incremental learning technique that along with MaxSAT formulation can generate classification rules efficiently. Our incremental learning is built on two concepts: (i)
mini-batch learning and (ii) iterative learning. In the following, we discuss both concepts in detail.

\subsection{Mini-batch Learning} 

Our first improvement is to implement a mini-batch learning technique tailored for rule-based classifiers. Mini-batch learning has two-fold advantages. Firstly, instead of solving a large MaxSAT query for the whole training data, this approach solves a sequence of smaller MaxSAT queries derived from mini-batches of smaller size.  Secondly, this approach extends to life-long learning~\shortcite{chen2018lifelong}, where the classifier can be updated incrementally with new samples while also generalizing to previously observed samples. In our context of rule-based classifiers, we consider the following heuristic in mini-batch learning.

\paragraph{A Heuristic for Mini-Batch Learning.} Let $ \Rule' $ be a classifier learned on the previous batch. In mini-batch learning, we aim to learn a new classifier $ \Rule $ that can generalize to both the current batch and previously seen samples. For that, we consider a soft constraint such that $ \Rule $ does not \emph{differ much} from $ \Rule' $ while training on the current batch. Thus, we hypothesize that by constraining $ \Rule $ to be syntactically similar to $ \Rule' $, it is possible to generalize well; because samples in all batches originate from the same distribution. Since our study focuses on rule-based classifiers, we consider the Hamming distance between two classifiers as a notion of their syntactic dissimilarity. In the following, we define the Hamming distance between two CNF classifiers $ \Rule $, $ \Rule' $ with the same number of clauses.
\[
	d_{H}(\Rule, \Rule') = \sum_{i=1}^k \Big(\sum_{v\in C_i} \mathds{1}(v \not\in C'_i) +  \sum_{v\in C'_i} \mathds{1}(v \not\in C_i) \Big), 
\]
 
 where $ C_i $ and $ C'_i $ are the $ i^\text{th} $ clause in $ \Rule $ and $ \Rule' $, respectively and $ \mathds{1} $ is an indicator function that returns $ 1 $ if the input is true and returns $ 0 $ otherwise.  Intuitively, $ d_{H}(\Rule, \Rule') $ calculates the total number of different literals in each (ordered) pair of clauses in $ \Rule $  and $ \Rule' $.  For example, consider $ \Rule = (x_1 \vee x_2) \wedge (\neg x_1) $ and $ \Rule' = (\neg x_1 \vee x_2) \wedge (\neg x_1) $. Then $ 	d_{H}(\Rule, \Rule') = 2 + 0 = 2 $, because in the first clause, the  literals $ \{x_1, \neg x_1\} $ are absent in either formulas, and the second clause is identical for both $ \Rule $ and $ \Rule' $. In the following, we discuss a modified objective function for mini-batch learning.

\paragraph{Objective Function.}
\label{sec:obj_incremental}
In mini-batch learning, we design an objective function to penalize both classification errors on the current batch and the Hamming distance between new and previous classifiers. Let $ (\mathbf{X}^b, \mathbf{y}^b) $ be the current mini-batch. The objective function in mini-batch learning is


\begin{align}
\label{eq:obj_incr}
\min_{\Rule} |\mathcal{E}_{\mathbf{X}_b}|+\lambda d_{H}(\Rule, \Rule').
\end{align}

In the objective function, $ \mathcal{E}_{\mathbf{X}_b} $ is the misclassified subset of samples in the current batch $ (\mathbf{X}^b, \mathbf{y}^b) $. Unlike controlling the rule-sparsity in the non-incremental approach in the earlier section, in Eq.~\eqref{eq:obj_incr}, $ \lambda $ controls the trade-off between classification errors and the syntactic differences between consecutive classifiers. Next, we discuss how to encode the modified objective function as a MaxSAT query.

\paragraph{MaxSAT Encoding of Mini-batch Learning.}
\label{sec:encoding_incremental}
In order to account for the modified objective function, we only modify the soft clause $ S_j^i $ in the MaxSAT query $ Q $. In particular, we define $ S_j^i $ to penalize for the complemented assignment  of the feature variable $ b_j^i $ in $ \Rule $ compared to $ \Rule' $.
\[
S_j^i:=
\begin{cases}
b_j^i& \text{if }  x_j \in  \clause(\Rule',i)  \\
\neg b_j^i & \text{otherwise}
\end{cases};\qquad \mathsf{wt}(S_j^i)=\lambda
\]

$ S_j^i $ is either a unit clause $ b_j^i $ or $ \neg b_j^i $ depending on whether the associated feature $ x_j $ appears in the previous classifier $ \Rule' $ or not. In this encoding, the Hamming distance of $ \Rule $ and $ \Rule' $ is minimized while attaining minimum classification errors on the current batch (using soft clause $ E_l $ in Eq.~\eqref{soft_clause_error}) To this end,  mini-batch learning starts with an empty CNF formula as $ \Rule' $ with no feature, and thus $ S_j^i := \neg b_j^i $ for the first batch. Next, we analyze the complexity of the MaxSAT encoding for mini-batch learning.

\begin{prop}
	\label{prop:maxsat_complexity_incremental}
	Let $ n' \triangleq |\mathbf{X}_b| \ll |\mathbf{X}| $ be the size of a mini-batch, $ n'_{neg} \le n' $ be the number of negative samples in the batch, and $ m $ be the number of Boolean features. According to Proposition~\ref{prop:maxsat_variables}, to learn a $ k $-clause CNF classifier in mini-batch learning, the MaxSAT encoding for each batch has $ km + n' $ Boolean variables and $ kn'_{neg} $ auxiliary variables. Let $ n'_{pos} = n' - n'_{neg} $ be the number of positive samples in the batch. Then, according to Proposition~\ref{prop:maxsat_clauses}, the number of clauses in the MaxSAT query for each batch is $ k m+n'+k n'_{pos}+(k m+1)n'_{neg} \approx \mathcal{O}(k m  n' ) $ when $ n'_{pos}= n'_{neg} =\frac{n'}{2} $.
\end{prop}

%

\paragraph{Assessing  the Performance of $ \Rule $.} Since we apply a heuristic objective in mini-batch learning,  $ \Rule $ may be optimized for the current batch while generalizing poorly on the full training data. To tackle this, after learning on each batch, we decide whether to keep $ \Rule $ or not by assessing the performance of $ \Rule $ on the training data and keep $ \Rule $ if it achieves higher performance. We measure the performance of $ \Rule $ on the full training data $ (\mathbf{X}, \mathbf{y}) $ using a weighted combination of classification errors and rule-size. In particular, we compute a combined loss function $ \mathsf{loss}(\Rule) \triangleq |\mathcal{E}_\mathbf{X}| +\lambda |\mathcal{R}| $ on the training data $ (\mathbf{X}, \mathbf{y}) $, which is indeed the value of the objective function that we minimize in the non-incremental learning in Section~\ref{sec:problem}. Additionally, we  discard the current classifier $ \Rule $ when the loss does not decrease ($ \mathsf{loss}(\Rule) > \mathsf{loss}(\Rule') $). 

We present the algorithm for mini-batch learning in Algorithm~\ref{algo:incremental}. 
\begin{algorithm}
	\caption{MaxSAT-based Mini-batch Learning}
	\label{algo:incremental}
	\begin{algorithmic}[1]
		\Function{$ \textsc{MiniBatchLearning} $}{$ \mathbf{X},\mathbf{y},\lambda, k $}
		\State $\mathcal{R} = \leftarrow \bigwedge_{i=1}^k\text{false} $ \Comment{Empty CNF formula}
		\State $ \mathsf{loss}_{\max} \leftarrow \infty$ 
		\For{$ i \leftarrow 1, \dots, N $} \Comment{$ N $ is the total batch-count}
		\State $ \mathbf{X}^b, \mathbf{y}^b \leftarrow \textsc{GetBatch}(\mathbf{X},\mathbf{y}) $
		\State $ Q, \weight{\cdot} \leftarrow \textsc{MaxSATEncoding}(\Rule, \mathbf{X}^b, \mathbf{y}^b, \lambda, k) $ \Comment{Returns a weighted-partial CNF}
		\State $ \sigma^* \leftarrow \MaxSAT(Q,\weight{\cdot}) $
		\State $ \Rule_{\mathsf{new}} \leftarrow \textsc{ConstructClassifier}(\sigma^*) $ \Comment{Construction~\ref{construction:rule}}
		\State $ \mathsf{loss} \leftarrow  |\mathcal{E}_\mathbf{X}| +\lambda |\mathcal{R}_{\mathsf{new}}| $ \Comment{Compute loss}
		\If{$ \mathsf{loss} < \mathsf{loss}_{\max} $}
		\State $ \mathsf{loss}_{\max} \leftarrow \mathsf{loss} $
		\State $ \Rule \leftarrow \Rule_{\mathsf{new}} $
		\EndIf
		\EndFor
		\State \Return $ \Rule $
		\EndFunction
	\end{algorithmic}
\end{algorithm}

\subsection{Iterative Learning}  We discuss an iterative learning algorithm for rule-based classifiers.  The major advantage of iterative learning is that we solve a smaller MaxSAT query because of learning a \textit{partial} classifier in each iteration. Our iterative approach is motivated by the set-covering algorithm\textemdash also known as separate-and-conquer algorithm\textemdash in symbolic rule learning~\shortcite{furnkranz1999separate}. In this approach, the core idea is to define the \emph{coverage of a partial classifier} (for example, a clause in a CNF classifier). For a specific definition of coverage, this  algorithm separates samples covered by the partial classifier and recursively conquers remaining samples in the training data by learning another partial classifier until no sample remains. The final classifier is an aggregation of all partial classifiers\textemdash the conjunction of clauses in a CNF formula, for example. 

Iterative learning is different from mini-batch learning in several aspects. In mini-batch learning, we learn all clauses in a CNF formula together, while in iterative learning, we learn a single clause of a CNF in each iteration. Additionally, in mini-batch learning, we improve scalability by reducing the number of samples in the training data using mini-batches, while in iterative learning, we improve scalability by reducing the number of clauses to learn at once. Therefore, an efficient integration of iterative learning and mini-batch learning would benefit scalability from both worlds. In the following, we discuss this integration by first stating iterative learning for CNF classifiers.

In iterative learning, we learn one clause of a CNF classifier in each iteration, where the clause refers to a partial classifier. The coverage of a clause in a CNF formula is the set of samples that \emph{do not satisfy} the clause. The reason is that if a sample does not satisfy at least one clause in a CNF formula, the prediction of the sample by the full formula is class $ 0 $, because CNF is a conjunction of clauses. As a result, considering covered samples in the next iteration does not change their prediction regardless of whatever clause we learn in later iterations. To this end, a single clause learning can be performed efficiently by applying mini-batch learning discussed before. In Algorithm~\ref{algo:iterative_CNF_learning}, we provide an algorithm for learning a CNF classifier iteratively by leveraging mini-batch learning. This algorithm is a double-loop algorithm, where in the outer loop we apply iterative learning and in the inner loop, we apply mini-batch learning.

\begin{algorithm}
	\caption{Iterative CNF Classifier Learning}
	\label{algo:iterative_CNF_learning}
	\begin{algorithmic}[1]
		\Procedure{IterativeCNFLearning}{$ \mathbf{X},\mathbf{y},\lambda, k $}
		\State $ \mathcal{R} \leftarrow $ true \Comment{Initial formula}
		\For{$ i \leftarrow 1, \dots, k $ and $ (\mathbf{X},\mathbf{y}) \ne \emptyset $}
		\State $ C_i \leftarrow \textsc{MiniBatchLearning}( \mathbf{X},\mathbf{y},\lambda, 1) $ \Comment{Single clause learning, $ k = 1 $}
		\label{algo:iterative_CNF_learning_incremental_learning}
		\State $ \mathbf{X}',\mathbf{y}' \leftarrow \textsc{Coverage}(\mathbf{X}, C_i) $
		\label{algo:iterative_CNF_learning_coverage}
		\If{$ (\mathbf{X}',\mathbf{y}') = \emptyset $} \Comment{Terminating conditions}
		\State break
		\EndIf
		\State $ \mathcal{R} \leftarrow \mathcal{R}  \wedge C_i $
		\State $ \mathbf{X},\mathbf{y}  \leftarrow (\mathbf{X},\mathbf{y})  \setminus (\mathbf{X}',\mathbf{y}') $ \Comment{Removing covered samples}
		\EndFor
		\State \Return $ \Rule $
		\EndProcedure
	\end{algorithmic}
\end{algorithm}

\paragraph{Terminating Conditions.} In Algorithm~\ref{algo:iterative_CNF_learning}, we terminate iterative learning based on three conditions: (i) when $ \Rule $ contains all $ k $ clauses , (ii) the training data $ (\mathbf{X},\mathbf{y}) $ is empty (that is, no sample remains uncovered), and (iii) no new sample is covered by the current partial classifier. Since the first two conditions are trivial, we elaborate on the third condition. When clause $ C_i $ cannot cover any new sample from the training dataset $ (\mathbf{X},\mathbf{y}) $, the next iteration will result in the same clause $ C_i $ because the training data remains the same. In this case, we do not include clause $ C_i $ to classifier $ \Rule $ because of zero coverage.

\section{Applying {\framework} on Learning Other Interpretable Classifiers}
\label{sec:application}
In earlier sections, we discuss the learning of CNF classifiers using  {\framework}. {\framework} can also be applied to learn other interpretable rule-based representations. In this section, we discuss how {\framework} can be applied in learning DNF classifiers, decision lists, and decision sets.

\subsection{Learning DNF Classifiers} 
\label{sec:dnf_learning}
For learning DNF classifiers, we leverage De Morgan's law where complementing a CNF formula results in a DNF formula. To learn a DNF classifier, say $ \Rule'(\mathbf{x}) $, we can trivially show that $ y = \Rule'(\mathbf{x}) \leftrightarrow \neg (y = \neg \Rule'(\mathbf{x})) $ for the feature vector $ \mathbf{x} $. Here $ \neg \Rule'(\mathbf{x}) $ is a CNF formula, by definition. Thus, we learn a DNF classifier by first complementing the class-label $ \mathbf{y} $ to $ \neg \mathbf{y}$ in the training dataset, learning a CNF classifier on $ (\mathbf{X}, \neg \mathbf{y}) $, and finally complementing the learned classifier to DNF. For example, the CNF classifier  ``(Male $ \vee $ Age $ < 50 $) $ \wedge $ (Education $ = $ Graduate $ \vee $ Income  $ \ge 1500 $)'' 	is complemented to a DNF classifier as 	 ``(not Male $ \wedge $ Age $ \ge 50 $) $ \vee $ (Education $ \ne $ Graduate $ \wedge $ Income $ \le 1500 $)''.

To learn a DNF classifier incrementally, such as through mini-batch and iterative learning, we adopt the following procedure. For learning a DNF classifier using mini-batch learning, we first learn a CNF classifier on dataset $ (\mathbf{X}, \neg \mathbf{y}) $ and complement the classifier to a DNF classifier at the end of mini-batch learning. To learn a DNF classifier in the iterative approach, we  learn a single clause of the DNF classifier in each iteration, remove covered samples, and continue till no training sample remains. In this context, the coverage of a clause in a DNF formula is the set of samples satisfying the clause.

\subsection{Learning Decision Lists}
In {\framework}, we apply an iterative learning approach for efficiently learning a  decision list. A decision list  $ \Rule_L $ is a list of pairs $ (C_1, v_1), \dots, (C_k, v_k) $, where we propose to learn one pair in each iteration.  We note that the clause $ C_i $ is a conjunction of literals\textemdash equivalently, a single clause DNF formula. Hence, our task is to deploy {\framework} to efficiently learn a single clause DNF formula $ C_i $. In particular, we opt to learn this clause for the majority class, say $ v_i $, in the training dataset by setting the majority samples as class $ 1 $ and all other samples as class $ 0 $. As a result, even if the MaxSAT-based learning, presented in this paper, is targeted for binary classification, we can learn a multi-class decision list in {\framework}.

In Algorithm~\ref{algo:iterative_decision_lists_learning}, we present the iterative algorithm for learning decision lists.  In each iteration, the algorithm learns a pair $ (C_i, v_i) $, separates the training set based on the coverage of $ (C_i, v_i) $ and conquers the remaining samples recursively. The coverage of $ (C_i, v_i) $ is the set of samples that satisfies clause $ C_i $. Finally, we add a default rule $ (C_k, v_\mathsf{default}) $ to $ \Rule_L $ where $ C_k \triangleq  $ true denoting that the clause is satisfied by all samples. We select the default class $ v_\mathsf{default} $ in the following order: (i) if any class(s) is not present in the predicted classes $ \{v_i\}_{i=1}^{k-1} $ of the decision list, 
 $ v_{\mathsf{default}} $ is the majority class among missing classes, and (ii)  $ v_{\mathsf{default}} $ is the majority class of the original training set $ (\mathbf{X}, \mathbf{y}) $, otherwise.

\begin{algorithm}
	\caption{Iterative learning of decision lists}
	\label{algo:iterative_decision_lists_learning}
	\begin{algorithmic}[1]
		\Procedure{DecisionListLearning}{$ \mathbf{X},\mathbf{y},\lambda, k $}
		\State $ \mathcal{R}_L \leftarrow \{\}$
		\For{$ i \leftarrow 1, \dots, k-1 $ and $ (\mathbf{X},\mathbf{y}) \ne \emptyset $}
		\State $ v_i \leftarrow \textsc{MajorityClass}(\mathbf{y}) $ \Comment{$ v_i $ specifies the target class}
		\State $ C_i \leftarrow \textsc{MiniBatchDNFLearning}( \mathbf{X},\mathbf{y},\lambda, 1, v_i) $ \Comment{Ref. Section~\ref{sec:dnf_learning}}
		
		\State $ \mathbf{X}',\mathbf{y}' \leftarrow \textsc{Coverage}(\mathbf{X}, C_i) $
		\If{$ (\mathbf{X}',\mathbf{y}') = \emptyset $} 
		\State break
		\EndIf
		\State $ \mathcal{R}_L \leftarrow \mathcal{R}_L \cup \{(C_i, v_i)\} $
		\State $ \mathbf{X},\mathbf{y}  \leftarrow (\mathbf{X},\mathbf{y})  \setminus (\mathbf{X}',\mathbf{y}') $ 
		\EndFor
		\State $ \mathcal{R}_L \leftarrow \mathcal{R}_L  \cup \{(\text{true}, v_{\mathsf{default}} )\}$	\Comment{Default rule}
		\State \Return $ \Rule_L $
		\EndProcedure
	\end{algorithmic}
	
\end{algorithm}

\subsection{Learning Decision Sets}
We describe an iterative procedure for learning decision sets. A decision set comprises of an individual clause-class  pair $ (C_i, v_i) $ where $ C_i $ denotes a single clause DNF formula similar to decision lists. In a decision set, a sample can satisfy multiple clauses simultaneously, which is attributed as an \textit{overlapping between clauses}~\shortcite{lakkaraju2016interpretable}. Concretely, the overlap between two clauses $ C_i $ and $ C_j $ with $ i \ne j $ is the set of samples $ \{\mathbf{X}_l | \mathbf{X}_l \models C_i \wedge \mathbf{X}_l \models C_j\} $ satisfying both clauses. One additional objective in learning a decision set is to minimize the overlap between clauses, as studied in~\shortcite{lakkaraju2016interpretable}. 
Therefore, along with optimizing accuracy and rule-sparsity, we propose an iterative procedure for decision sets that additionally minimizes the overlap between clauses.

\begin{algorithm}
	\caption{Iterative learning of decision sets}
	\label{algo:iterative_decision_sets_learning}
	\begin{algorithmic}[1]
		\Procedure{DecisionSetsLearning}{$ \mathbf{X},\mathbf{y},\lambda, k$}
		\State $ \mathcal{R}_S=\{\} $
		\State $ \mathbf{X}^\text{cc} = \{\}$  \Comment{Contains correctly covered samples}
		\For{$ i \leftarrow 1, \dots, k-1 $ and $ (\mathbf{X},\mathbf{y}) \ne \emptyset $}
		\State $ v_i \leftarrow \textsc{MajorityClass}(\mathbf{X},\mathbf{y}) $
		\State $ \mathbf{X}^\text{w} \leftarrow \mathbf{X} \cup \mathbf{X}^\text{cc} $ \Comment{Correctly covered samples are included}
		\State $ \mathbf{y}^\text{w} \leftarrow  \mathbf{y} \cup  \{\neg v_i\}^{|\mathbf{X}^\text{cc}|}$ \Comment{$ \mathbf{X}^\text{cc} $ have complemented class $ \neg v_i $ to minimize overlap}
		\State $ C_i \leftarrow \textsc{MiniBatchDNFLearning}( \mathbf{X}^\text{w},\mathbf{y}^\text{w},\lambda, 1, v_i) $ 
		\State $ \mathbf{X}',\mathbf{y}' \leftarrow \textsc{CorrectCoverage}(\mathbf{X}, \mathbf{y}, C_i, v_i) $
		\If{$ (\mathbf{X}',\mathbf{y}') = \emptyset $} 
		\State break
		\EndIf
		\State $ \mathcal{R}_S \leftarrow \mathcal{R}_S \cup \{(C_i, v_i)\} $
		\State $ \mathbf{X},\mathbf{y} \leftarrow (\mathbf{X},\mathbf{y})  \setminus (\mathbf{X}',\mathbf{y}') $
		\State $ \mathbf{X}^\text{cc} = \mathbf{X}^\text{cc} \cup \mathbf{X}'$ 
		\EndFor
		\State $ \mathcal{R}_S \leftarrow \mathcal{R}_S  \cup \{(\text{true}, v_{\mathsf{default}} )\}$
		\State \Return $ \Rule_S $
		\EndProcedure
	\end{algorithmic}
\end{algorithm}

In Algorithm~\ref{algo:iterative_decision_sets_learning}, we present an iterative algorithm for learning a decision set. The iterative algorithm is a modification of separate-and-conquer algorithm  by additionally focusing on minimizing overlaps in a decision set.  Given a training data $ (\mathbf{X}, \mathbf{y}) $, a  regularization parameter $ \lambda $, and the number of clauses $ k $, the core idea of Algorithm~\ref{algo:iterative_decision_sets_learning} is to learn a pair $ (C_i,v_i) $ in each iteration, separate covered samples from $ (\mathbf{X}, \mathbf{y}) $, and conquer remaining samples recursively. In contrast to learning decision lists, we have following modifications in Algorithm~\ref{algo:iterative_decision_sets_learning}. 

\begin{itemize}
	\item The first modification is with respect to the definition of coverage for decision sets. Unlike decision lists, we separate samples that are \textit{correctly covered} by $ (C_i, v_i) $ in each iteration. Given  $ (\mathbf{X}, \mathbf{y}) $, the correctly covered samples of $ (C_i, v_i) $ is a set $  \{(\mathbf{X}_l, \mathbf{y}_l)| \mathbf{X}_l \models C_i \wedge \mathbf{y}_l = v_i \} \subseteq (\mathbf{X}, \mathbf{y}) $ of samples that satisfy $ C_i $ and have matching class-label as $ v_i $. 
	\item The second modification is related to the  training dataset considered in each iteration. Let $ (\mathbf{X},\mathbf{y}) $ denote \emph{the remaining training dataset} in the current iteration and $ v_i $ be the majority class in  $ (\mathbf{X},\mathbf{y}) $. Hence, $ v_i $ is the target class in the current iteration. In Algorithm~\ref{algo:iterative_decision_sets_learning}, we learn clause $ C_i $ on \emph{a working dataset} $ (\mathbf{X}^\text{w},\mathbf{y}^\text{w}) $, where $ \mathbf{X}^\text{w} \triangleq \mathbf{X} \cup \mathbf{X}^\text{cc} $ comprises of both remaining samples and already covered samples in all previous iterations. However, the vector of class labels $ \mathbf{y}^\text{w} \triangleq  \mathbf{y} \cup  \{\neg v_i\}^{|\mathbf{X}^\text{cc}|} $  comprises of remaining class-label vector $ \mathbf{y} $ and complemented class label vector $ \{\neg v_i\}^{|\mathbf{X}^\text{cc}|} $ associated with feature-matrix $ \mathbf{X}^\text{cc} $. Thus, by explicitly labeling covered samples as class $ \neg v_i $, the new clause $ C_i $ learns to falsify already covered samples. This heuristic allows us to minimize the overlap of $ C_i $ compared to previously learned clauses $ \{C_j\}_{j=1}^{i-1} $.
\end{itemize}

Finally, the default clause for decision sets is learned similarly as in decision lists.

\newcommand{\maxsatquery}{\ensuremath{\mathsf{ConstructQuery}}}
\newcommand{\createclause}{\ensuremath{\mathsf{ConstructClause}}}
\newcommand{\iterativeClauseLearning}{\ensuremath{\mathsf{IterativeClauseLearning}}}
\let\oldReturn\Return
\renewcommand{\Return}{\State\oldReturn}

\section{Empirical Performance Analysis}
\label{sec:experiments}
In this section, we empirically evaluate the performance of {\framework}. We first present the experimental setup and the objective of the experiments, followed by experimental results.

\subsection{Experimental Setup}
We implement a prototype of {\framework} in Python to evaluate the performance of the proposed MaxSAT-based formulation for learning classification rules. To implement {\framework}, we deploy a state-of-the-art MaxSAT solver Open-WBO~\shortcite{martins2014open}, which returns the current best solution upon reaching a timeout. 

We compare {\framework} with state-of-the-art interpretable and non-interpretable classifiers. Among interpretable classifiers, we compare with RIPPER~\shortcite{cohen1995fast}, BRL~\shortcite{letham2015interpretable}, CORELS~\shortcite{angelino2017learning}, and BRS~\shortcite{wang2017bayesian}. Among non-interpretable classifiers, we compare with Random Forest (RF), Support Vector Machine with linear kernels (SVM), Logistic Regression classifier (LR), and k-Nearest Neighbors classifier (kNN). We deploy the Scikit-learn library in Python for implementing non-interpretable classifiers.

We experiment with real-world binary classification datasets from  UCI~\shortcite{Dua:2019}, Open-ML~\shortcite{OpenML2013}, and Kaggle repository (\url{https://www.kaggle.com/datasets}), as listed in Table~\ref{table:interpretable_classifiers}. In these datasets, the number of samples vary from about $ 200 $ to $ 1,000,000 $.  The datasets contain both real-valued and categorical features. We process them to binary features by setting the maximum number of bins as $ 10 $ during discretization. For non-interpretable classifiers such as RF, SVM, LR, and kNN, which take real-valued features as inputs, we only convert categorical features to one-hot encoded binary features.

We perform ten-fold cross-validation on each dataset and evaluate the performance of different classifiers based on the median prediction accuracy on the test data. Additionally, we compare the median size of generated rules among rule-based interpretable classifiers. We consider a comparable combination ($ 100 $) of  hyper-parameters choices for all classifiers, which we fine-tune during cross-validation. For {\framework}, we vary the number of clauses $ k \in \{1, 2, \dots, 5\} $ and the regularization parameter $ \lambda $  in a logarithmic grid by choosing $ 5 $ values between $ 10^{-4} $ and $ 10^1 $. In the mini-batch learning in {\framework}, we set the number of samples in each mini-batch, $ n' \in \{50, 100, 200, 400\} $. Thus, we consider $ \lceil n / n' \rceil $ mini-batches, where $ n $ denotes the size of training data. To construct mini-batches from a training dataset, we sequentially split the data into $ \lceil n / n' \rceil $  batches with each batch having $ n' $ samples. Furthermore, to ignore the effect of batch-ordering, we perform mini-batch learning in two rounds such that each batch participates twice in the training.

For BRL algorithm, we vary four hyper-parameters: the maximum cardinality of rules in $ \{2, 3, 4\} $, the minimum support of rules in $ \{0.05, 0.175, 0.3\} $, and the prior on the expected length and width of rules in $ \{2, 4, 6, 8\} $ and $ \{2, 5, 8\} $, respectively. For CORELS algorithm, we vary three hyper-parameters:  the maximum cardinality of rules in $ \{2, 3, 4, 5\} $, the minimum support of rules in $ \{0.01, 0.17, 0.33, 0.5  \} $, and the regularization parameter in $ \{0.005, 0.01 , 0.015, 0.02 , 0.025, 0.03\}  $. For BRS algorithm, we vary three hyper-parameters: the number of initial rules in $ \{500, 1000, 1500, 2000, 2500, 3000\} $, the maximum length of rules in $ \{1,2,3,4\} $, and the minimum support of rules in $ \{1, 4,7, 10\} $. For RF and RIPPER classifiers, we vary the cut-off on the number of samples in the leaf node using a linear grid between $ 3 $ to $ 500 $ and $ 1 $ to $ 300 $, respectively. For SVM and LR classifiers, we discretize the regularization parameter on a logarithmic grid between  $ 10^{-3} $ and $ 10^3 $. For kNN, we vary the number of neighbors in a linear grid between $ 1 $ and $ 500 $. We conduct each experiment on an Intel Xeon E$ 7-8857 $ v$ 2 $ CPU using a single core with $ 16 $ GB of RAM running on a 64bit Linux distribution based on Debian. For all classifiers, we set the training timeout to $ 1000 $ seconds.

\paragraph{Objectives of Experiments.} In the following, we present the objectives of our experimental study. 

\begin{enumerate}
	\item How are the accuracy and size of classification rules generated by  {\framework} compared to existing interpretable classifiers?
	\item How is the scalability of {\framework} in solving large-scale classification problems compared to existing interpretable classifiers?
	\item How does {\framework} perform in terms of accuracy and scalability compared to classifiers that are non-interpretable? 
	\item How does the incremental learning in {\framework} perform compared to non-incremental MaxSAT-based learning in terms of accuracy, rule-sparsity, and scalability?
	\item How do different interpretable classification rules learned using {\framework} perform in terms of accuracy and rule-size?
	\item What are the effects of different hyper-parameters in {\framework}?
\end{enumerate} 

\paragraph{Summary of Experimental Results.}
To summarize our experimental results, {\framework} achieves the best balance among prediction accuracy, interpretability, and scalability compared to existing interpretable rule-based classifiers. Particularly, compared to the most accurate classifier RIPPER, {\framework} demonstrates on average $ 1\% $ lower prediction accuracy, wherein the accuracy of {\framework} is higher than BRL, CORELS, and BRS in almost all datasets. In contrast, {\framework} generates significantly smaller rules than RIPPER, specifically in large datasets. Moreover, BRL, CORELS, and BRS report comparatively smaller rule-size than {\framework} on average, but with a significant decrease in accuracy.  In terms of scalability, {\framework} achieves the best performance compared to other interpretable and non-interpretable classifiers by classifying datasets with one million samples. While CORELS also scales to such large datasets,  its accuracy is lower than that of {\framework} by at least $ 2\% $ on average.  Therefore, {\framework} is not only scalable but also accurate in practical classification problems while also being interpretable by design.  We additionally analyze the comparative performance of different formulations presented in this paper, where the  incremental approach empirically proves its efficiency than the na\"ive MaxSAT formulation.  Furthermore, we learn and compare the performance of different interpretable representations: decision lists, decision sets, CNF, and DNF formulas using {\framework} and present the efficacy of {\framework} in learning varied interpretable classifiers. Finally, we study the effect of different hyper-parameters in {\framework}, where each hyper-parameter provides a precise control among training time,  prediction accuracy, and rule-sparsity.

\begin{table*}[!t]
	\centering
	
	\caption{Comparison of accuracy and rule-size among interpretable classifiers. Each cell from the fourth to the eighth column contains test accuracy (top) and rule-size (bottom). `\textemdash' represents a timeout. Numbers in bold represent the best performing results among different classifiers.}
	\label{table:interpretable_classifiers} 
	\small
	\begin{tabular}{lrrrrrrrrrrrr}

	\toprule
	Dataset & Size & Features & RIPPER & BRL & CORELS & BRS & IMLI \\
	
	\midrule
	\multirow{2}{*}{Parkinsons} & \multirow{2}{*}{ $ 195 $ } & \multirow{2}{*}{ $ 202 $ }  &
	$ 94.44 $  &  $ \mathbf{94.74} $  &  $ 89.74 $  &  $ 84.61 $  &  $ \mathbf{94.74} $  \\
	&&& $ 7.0 $  &  $ 11.5 $  &  $ \mathbf{2.0} $  &  $ 5.0 $  &  $ 7.5 $  \\
	\addlinespace[0.5em]
	
	\multirow{2}{*}{WDBC} & \multirow{2}{*}{ $ 569 $ } & \multirow{2}{*}{ $ 278 $ }  &
	$ \mathbf{98.08} $  &  $ 93.81 $  &  $ 92.04 $  &  $ 92.98 $  &  $ 94.74 $  \\
	&&& $ 13.0 $  &  $ 22.0 $  &  $ \mathbf{2.0} $  &  $ 7.0 $  &  $ 11.5 $  \\
	\addlinespace[0.5em]
	
	\multirow{2}{*}{Pima} & \multirow{2}{*}{ $ 768 $ } & \multirow{2}{*}{ $ 83 $ }  &
	$ 77.14 $  &  $ 68.18 $  &  $ 75.32 $  &  $ 75.32 $  &  $ \mathbf{78.43} $  \\
	&&& $ 6.0 $  &  $ 13.5 $  &  $ \mathbf{2.0} $  &  $ 3.0 $  &  $ 23.0 $  \\
	\addlinespace[0.5em]
	
	\multirow{2}{*}{Titanic} & \multirow{2}{*}{ $ 1,043 $ } & \multirow{2}{*}{ $ 38 $ }  &
	$ 78.72 $  &  $ 62.98 $  &  $ \mathbf{81.9} $  &  $ 80.86 $  &  $ 81.82 $  \\
	&&& $ 6.0 $  &  $ 15.0 $  &  $ \mathbf{4.0} $  &  $ \mathbf{4.0} $  &  $ 5.5 $  \\
	\addlinespace[0.5em]
	
	\multirow{2}{*}{MAGIC} & \multirow{2}{*}{ $ 19,020 $ } & \multirow{2}{*}{ $ 100 $ }  &
	$ \mathbf{82.68} $  &  $ 76.95 $  &  $ 78.05 $  &  $ 77.5 $  &  $ 78.26 $  \\
	&&& $ 102.0 $  &  $ 81.0 $  &  $ 4.0 $  &  $ \mathbf{3.0} $  &  $ 8.5 $  \\
	\addlinespace[0.5em]
	
	\multirow{2}{*}{Tom's HW} & \multirow{2}{*}{ $ 28,179 $ } & \multirow{2}{*}{ $ 946 $ }  &
	$ \mathbf{85.91} $  & \textemdash &  $ 83.27 $  &  $ 83.13 $  &  $ 85.24 $  \\
	&&& $ 30.0 $  & \textemdash &  $ \mathbf{4.0} $  &  $ 18.5 $  &  $ 44.5 $  \\
	\addlinespace[0.5em]
	
	\multirow{2}{*}{Credit} & \multirow{2}{*}{ $ 30,000 $ } & \multirow{2}{*}{ $ 199 $ }  &
	$ \mathbf{82.39} $  &  $ 46.12 $  &  $ 81.18 $  &  $ 80.45 $  &  $ 82.12 $  \\
	&&& $ 32.5 $  &  $ 26.5 $  &  $ \mathbf{2.0} $  &  $ 7.0 $  &  $ 17.5 $  \\
	\addlinespace[0.5em]
	
	\multirow{2}{*}{Adult} & \multirow{2}{*}{ $ 32,561 $ } & \multirow{2}{*}{ $ 94 $ }  &
	$ \mathbf{84.37} $  &  $ 72.08 $  &  $ 79.78 $  &  $ 70.75 $  &  $ 81.2 $  \\
	&&& $ 115.5 $  &  $ 46.5 $  &  $ \mathbf{4.0} $  &  $ \mathbf{4.0} $  &  $ 30.0 $  \\
	\addlinespace[0.5em]
	
	\multirow{2}{*}{Bank Marketing} & \multirow{2}{*}{ $ 45,211 $ } & \multirow{2}{*}{ $ 82 $ }  &
	$ \mathbf{90.01} $  &  $ 84.66 $  &  $ 89.62 $  &  $ 86.75 $  &  $ 89.84 $  \\
	&&& $ 36.5 $  &  $ 13.0 $  &  $ \mathbf{2.0} $  &  $ \mathbf{2.0} $  &  $ 24.5 $  \\
	\addlinespace[0.5em]
	
	\multirow{2}{*}{Connect-4} & \multirow{2}{*}{ $ 67,557 $ } & \multirow{2}{*}{ $ 126 $ }  &
	$ \mathbf{76.72} $  &  $ 65.83 $  &  $ 68.68 $  &  $ 70.49 $  &  $ 75.36 $  \\
	&&& $ 118.0 $  &  $ 18.5 $  &  $ \mathbf{4.0} $  &  $ 11.0 $  &  $ 50.5 $  \\
	\addlinespace[0.5em]
	
	\multirow{2}{*}{Weather AUS} & \multirow{2}{*}{ $ 107,696 $ } & \multirow{2}{*}{ $ 169 $ }  &
	$ \mathbf{84.22} $  &  $ 43.26 $  &  $ 83.67 $  & \textemdash &  $ 83.78 $  \\
	&&& $ 26.0 $  &  $ 22.0 $  &  $ \mathbf{2.0} $  & \textemdash &  $ 22.0 $  \\
	\addlinespace[0.5em]
	
	\multirow{2}{*}{Vote} & \multirow{2}{*}{ $ 131,072 $ } & \multirow{2}{*}{ $ 16 $ }  &
	$ \mathbf{97.12} $  &  $ 94.78 $  &  $ 95.86 $  &  $ 95.14 $  &  $ 96.69 $  \\
	&&& $ 132.0 $  &  $ 41.5 $  &  $ 3.5 $  &  $ \mathbf{1.0} $  &  $ 15.0 $  \\
	\addlinespace[0.5em]
	
	\multirow{2}{*}{Skin Seg} & \multirow{2}{*}{ $ 245,057 $ } & \multirow{2}{*}{ $ 30 $ }  &
	\textemdash &  $ 79.25 $  &  $ 91.62 $  &  $ 68.48 $  &  $ \mathbf{94.71} $  \\
	&&&\textemdash &  $ 6.0 $  &  $ 9.0 $  &  $ \mathbf{5.0} $  &  $ 30.0 $  \\
	\addlinespace[0.5em]
	
	\multirow{2}{*}{BNG(labor)} & \multirow{2}{*}{ $ 1,000,000 $ } & \multirow{2}{*}{ $ 89 $ }  &
	\textemdash & \textemdash &  $ 88.56 $  & \textemdash &  $ \mathbf{90.91} $  \\
	&&&\textemdash & \textemdash &  $ \mathbf{2.0} $  & \textemdash &  $ 24.0 $  \\
	\addlinespace[0.5em]
	
	\multirow{2}{*}{BNG(credit-g)} & \multirow{2}{*}{ $ 1,000,000 $ } & \multirow{2}{*}{ $ 97 $ }  &
	\textemdash & \textemdash &  $ 72.08 $  & \textemdash &  $ \mathbf{75.48} $  \\
	&&&\textemdash & \textemdash &  $ \mathbf{2.0} $  & \textemdash &  $ 27.5 $  \\
	\bottomrule

	\end{tabular}

\end{table*}

\subsection{Experimental Results}

In the following, we discuss our experimental results in detail. 

\subsubsection{Comparing {\framework} with Interpretable Classifiers} We compare {\framework} with existing interpretable classifiers in three aspects: test accuracy, rule-size, and scalability.

\paragraph{Test accuracy and rule-size.}  We present the experimental  results of test accuracy and rule-size among interpretable classifiers in Table~\ref{table:interpretable_classifiers}, where the first, second, and third columns represent  the name of the dataset, the number of samples, and the number of features in the dataset, respectively. In each cell from the fourth to the eighth column in the table, the top value represents the median test accuracy and the bottom value represents the median size of rules measured through ten-fold cross-validation.

In Table~\ref{table:interpretable_classifiers}, {\framework} and CORELS generate interpretable classification rules in all $ 15 $ datasets in our experiments. In contrast, within a timeout of $ 1000 $ seconds, RIPPER, BRL, and BRS fail to generate any classification rule in three datasets,  specifically in large datasets ($ \ge 200,000 $ samples).

We compare {\framework} with each interpretable classifier in detail. Compared to RIPPER, {\framework} has lower accuracy in $ 9 $ out  of $ 12 $ datasets. More specifically, the accuracy of {\framework} is $ 1\% $ lower on average than RIPPER. The improved accuracy of RIPPER, however, results in the generation of higher size classification rules than {\framework} in most datasets. In particular, {\framework} generates sparser rules than RIPPER in $ 9 $ out of $ 12 $ datasets, wherein RIPPER times out in $ 3 $ datasets. Interestingly, the difference in rule-size is more significant in larger datasets, such as in `Vote' dataset, where RIPPER learns a classifier with $ 132 $ Boolean literals compared to $ 15 $ Boolean literals by {\framework}.  Therefore, {\framework} is better than RIPPER in terms of rule-sparsity, but lags slightly in accuracy.

{\framework} performs better than BRL both in terms of accuracy and rule-sparsity. In particular, {\framework} has  higher accuracy and lower rule-size than BRL in $ 12 $ and $ 8 $ datasets, respectively, in a total of $ 12 $ datasets, wherein BRL times out in $ 3 $ datasets.  While comparing with CORELS, {\framework} achieves higher accuracy in almost all datasets ($ 14 $ out of $ 15 $ datasets). A similar trend is observed in comparison with BRS, where {\framework} achieves higher accuracy in all of $ 12 $ datasets and BRS times out in $ 3 $ datasets. CORELS and BRS, however, generates sparser rules than {\framework} in most datasets, but by costing a significant decrease in accuracy.  For example, in the largest dataset `BNG(credit-g)' with $ 1 $ Million samples, BRS times out, and CORELS generates a classifier with  $ 72.08\% $ accuracy with rule-size $ 2 $. {\framework}, in contrast, learns a classifier with $ 27.5 $ Boolean literals achieving $ 75.48\% $ accuracy, which is $ 3\% $ higher than CORELS. Therefore, {\framework}  makes a good balance between accuracy and rule-size compared to existing interpretable classifiers while also being highly scalable. In the following, we discuss the results on the scalability of all interpretable classifiers in detail.

\begin{figure}[!t]
	
	\centering
	
	\subfloat{\includegraphics[scale=0.4]{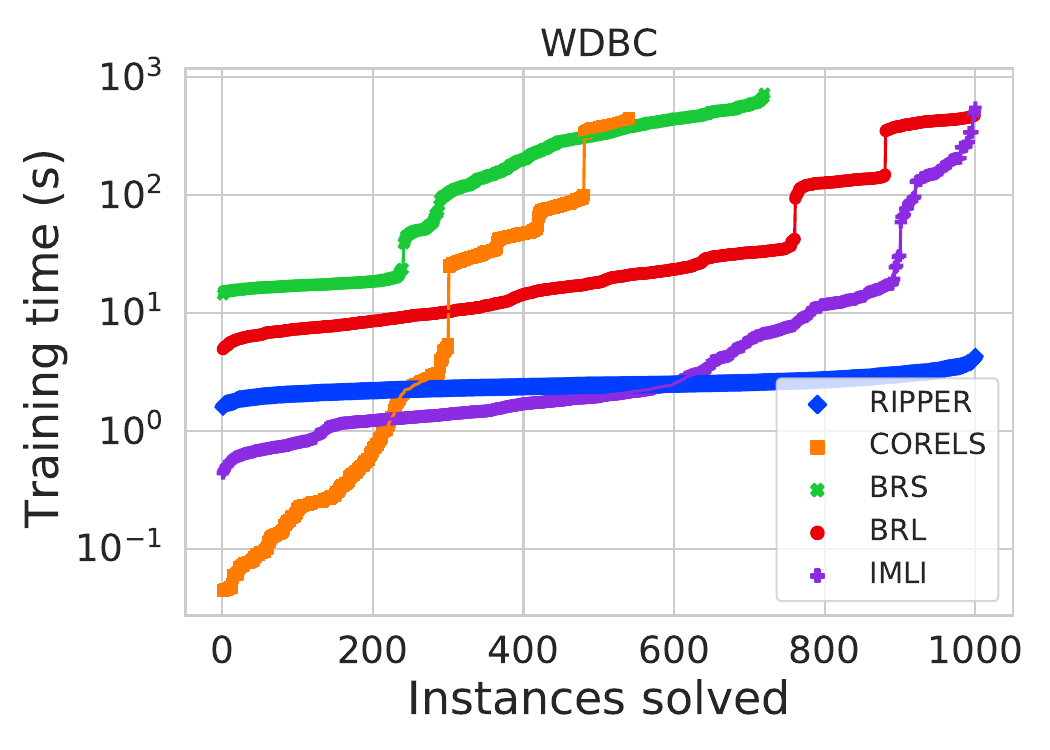}}
	\subfloat{\includegraphics[scale=0.4]{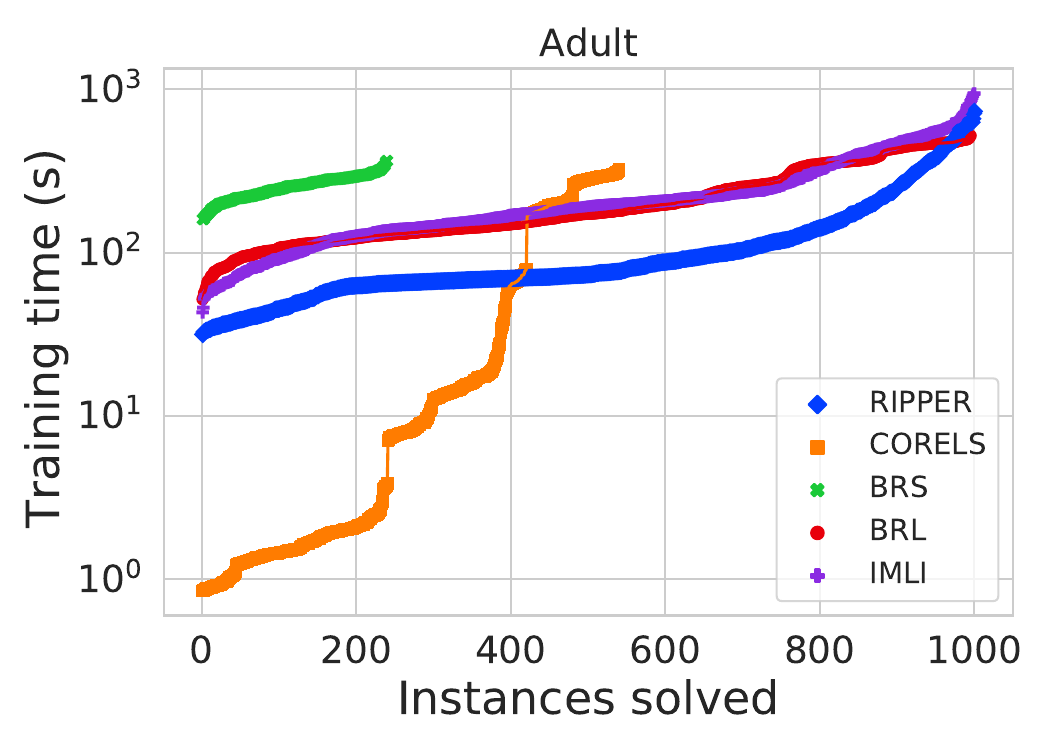}}\\
	\subfloat{\includegraphics[scale=0.4]{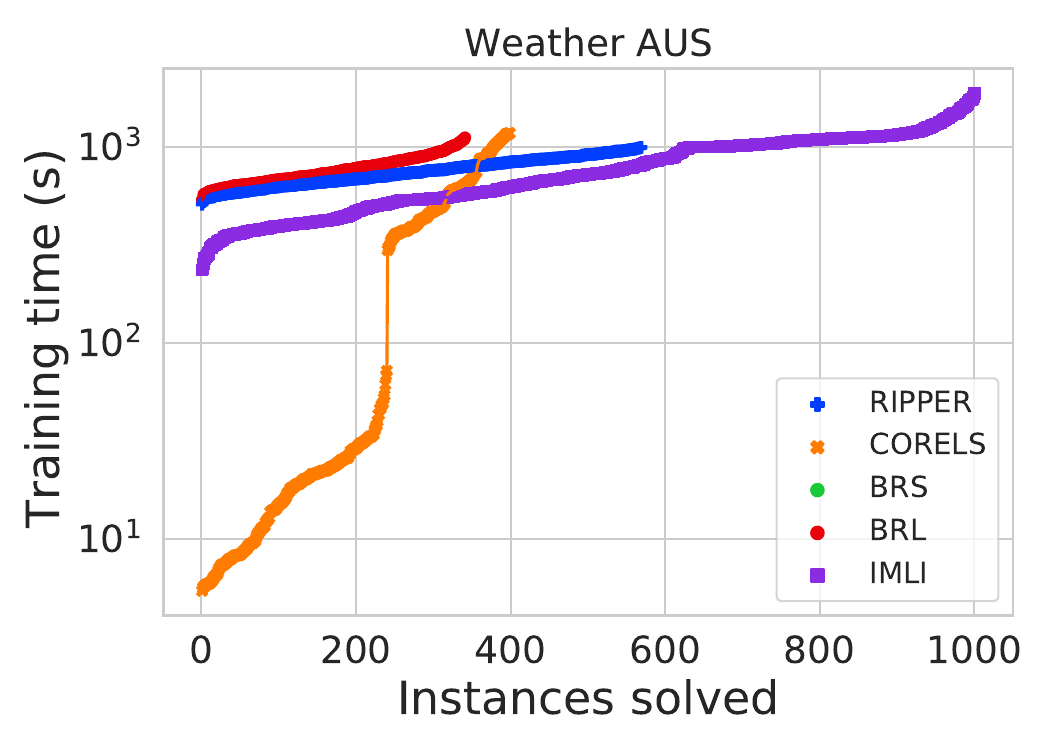}}
	\subfloat{\includegraphics[scale=0.4]{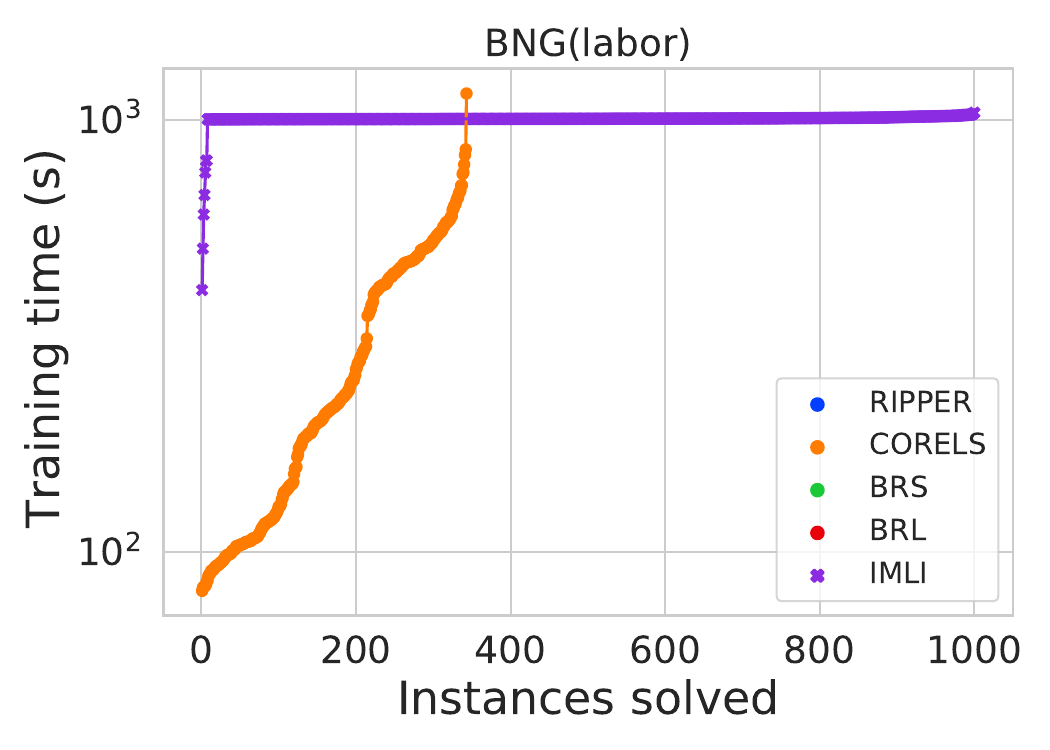}}
	
	\caption{Comparison of scalability among interpretable classifiers. The plots are arranged in increasing sizes of datasets (from left to right). In each cactus plot, {\framework} solves all $ 1000 $ instances for each dataset, while competitive classifiers often fail to scale, specially in larger datasets.}
	\label{fig:interpretable_classifiers}
\end{figure}

\paragraph{Scalability.} We analyze the scalability among interpretable classifiers by comparing  training time. In Figure~\ref{fig:interpretable_classifiers}, we use cactus plots\footnote{Cactus plots are often used in (Max)SAT community to present the scalability of different solvers/methods~\shortcite{argelich2008first,balyo2017sat}.} to represent the training time (in seconds) of all classifiers in $ 1000 $ instances ($ 10 $ folds $ \times  $ $ 100 $ choices of hyper-parameters) derived for each dataset. In the cactus plot, the number of solved instances (within $ 1000 $ seconds) is on the $ X $-axis, whereas the training time is  on the $ Y $-axis. A point $ (x,y) $ on the plot implies that a classifier yields lower than or equal to $ y $ seconds of training in $ x $ many instances.

In Figure~\ref{fig:interpretable_classifiers}, we present results in an increasing number of samples in a dataset (from left to right and top to bottom).
In WDBC and Adult datasets presented on the first row in Figure~\ref{fig:interpretable_classifiers}, CORELS solves lower than $  600 $ instances within a timeout of $ 1000 $ seconds. The scalability performance of BRS is even worse, where it solves around $ 700 $ and $ 200 $ instances in WDBC and Adult datasets, respectively. The other three classifiers: {\framework}, BRL, and RIPPER  solve all $ 1000 $ instances, where BRL takes comparatively higher training time than the other two. The performance of {\framework} and RIPPER is similar, with RIPPER being comparatively better in the two datasets. However, the efficiency of {\framework} compared to other classifiers becomes significant as the number of samples in a dataset increases. In particular, in `Weather AUS' dataset, BRS cannot solve a single instance, BRL and CORELS solves $ ~400 $ instances, and RIPPER solves around $ 600 $ instances. {\framework}, however, solves all $ 1000 $ instances in this dataset. Similarly, in `BNG(labor)' dataset, all other classifiers except {\framework} and CORELS cannot solve any instance. While {\framework} mostly takes the maximum allowable time ($ 1000 $ seconds) in solving all instances in this dataset, CORELS can solve lower than $ 400 $ instances. 
Thus, {\framework} establishes itself as the most scalable classifier compared to other state-of-the-art interpretable classifiers.

\subsubsection{Comparison with Non-interpretable Classifiers} We compare {\framework} with state-of-the-art non-interpretable classifiers such as LR, SVM, kNN, and RF in terms of their median test accuracy in Table~\ref{table:non_interpretable_classifiers}. In the majority of the datasets, {\framework} achieves comparatively lower test accuracy than the best performing non-interpretable classifier. The decrease in the test accuracy of {\framework} is attributed to two factors. Firstly, while we train {\framework} on discretized data, non-interpretable classifiers are trained on non-discretized data and thus {\framework} incurs additional classification errors due to discretization. Secondly, {\framework} learns a rule-based classifier, whereas non-interpretable classifiers can learn more flexible decision boundaries and thus fit data well. In Table~\ref{table:non_interpretable_classifiers}, we also observe that {\framework} achieves impressive scalability than competing classifiers by solving datasets with $ 1, 000,000 $  samples where most of the non-interpretable classifiers fail to learn any decision boundary on such large datasets. Thus, {\framework}, being an interpretable classifier, demonstrates lower accuracy than competing non-interpretable classifiers, but higher scalability in practice.

\begin{table*}[!t]        
	\centering
	\caption{Comparison of {\framework} with non-interpretable classifiers in terms of test accuracy. In the table, `\textemdash' represents a timeout. Numbers in bold represent the best performing results among different classifiers.}
	\label{table:non_interpretable_classifiers}
	\small
	\begin{tabular}{lrrrrrrrrrrrrrrr}

		\toprule
		Dataset & Size & LR & SVM & kNN & RF & IMLI \\
		
		\midrule
		\multirow{1}{*}{Parkinsons} & \multirow{1}{*}{ $ 195 $ }  &
		$ 89.74 $  &  $ 89.74 $  &  $ \mathbf{97.5} $  &  $ 90.0 $  &  $ 94.74 $  \\
		\multirow{1}{*}{WDBC} & \multirow{1}{*}{ $ 569 $ }  &
		$ \mathbf{98.25} $  &  $ \mathbf{98.25} $  &  $ 98.23 $  &  $ 96.49 $  &  $ 94.74 $  \\
		\multirow{1}{*}{Pima} & \multirow{1}{*}{ $ 768 $ }  &
		$ 78.43 $  &  $ 79.08 $  &  $ 74.5 $  &  $ \mathbf{79.22} $  &  $ 78.43 $  \\
		\multirow{1}{*}{Titanic} & \multirow{1}{*}{ $ 1,043 $ }  &
		$ 80.86 $  &  $ 80.38 $  &  $ 81.34 $  &  $ \mathbf{82.69} $  &  $ 81.82 $  \\
		\multirow{1}{*}{MAGIC} & \multirow{1}{*}{ $ 19,020 $ }  &
		$ 79.18 $  &  $ 79.34 $  &  $ 84.6 $  &  $ \mathbf{88.2} $  &  $ 78.26 $  \\
		\multirow{1}{*}{Tom's HW} & \multirow{1}{*}{ $ 28,179 $ }  &
		$ 96.2 $  &  $ 97.13 $  &  $ 88.15 $  &  $ \mathbf{97.78} $  &  $ 85.24 $  \\
		\multirow{1}{*}{Credit} & \multirow{1}{*}{ $ 30,000 $ }  &
		$ \mathbf{82.2} $  &  $ 81.9 $  &  $ 81.83 $  &  $ 82.15 $  &  $ 82.12 $  \\
		\multirow{1}{*}{Adult} & \multirow{1}{*}{ $ 32,561 $ }  &
		$ 85.26 $  &  $ 85.05 $  &  $ 83.8 $  &  $ \mathbf{86.69} $  &  $ 81.2 $  \\
		\multirow{1}{*}{Bank Marketing} & \multirow{1}{*}{ $ 45,211 $ }  &
		$ 90.09 $  &  $ 89.28 $  &  $ 89.43 $  &  $ \mathbf{90.27} $  &  $ 89.84 $  \\
		\multirow{1}{*}{Connect-4} & \multirow{1}{*}{ $ 67,557 $ }  &
		$ 79.39 $  & \textemdash &  $ 85.51 $  &  $ \mathbf{88.11} $  &  $ 75.36 $  \\
		\multirow{1}{*}{Weather AUS} & \multirow{1}{*}{ $ 107,696 $ }  &
		$ 85.64 $  & \textemdash &  $ 78.59 $  &  $ \mathbf{86.26} $  &  $ 83.78 $  \\
		\multirow{1}{*}{Vote} & \multirow{1}{*}{ $ 131,072 $ }  &
		$ 96.43 $  &  $ 96.37 $  &  $ 97.05 $  &  $ \mathbf{97.38} $  &  $ 96.69 $  \\
		\multirow{1}{*}{Skin Seg} & \multirow{1}{*}{ $ 245,057 $ }  &
		$ 91.86 $  & \textemdash &  $ \mathbf{99.96} $  &  $ \mathbf{99.96} $  &  $ 94.71 $  \\
		\multirow{1}{*}{BNG(labor)} & \multirow{1}{*}{ $ 1,000,000 $ }  &
		\textemdash & \textemdash & \textemdash & \textemdash &  $ \mathbf{90.91} $  \\
		\multirow{1}{*}{BNG(credit-g)} & \multirow{1}{*}{ $ 1,000,000 $ }  &
		\textemdash & \textemdash & \textemdash &  $ \mathbf{80.58} $  &  $ 75.48 $  \\
		\bottomrule

	\end{tabular}
	
\end{table*}

\begin{figure}[!t]

	\centering
	
	\subfloat{\includegraphics[scale=0.28]{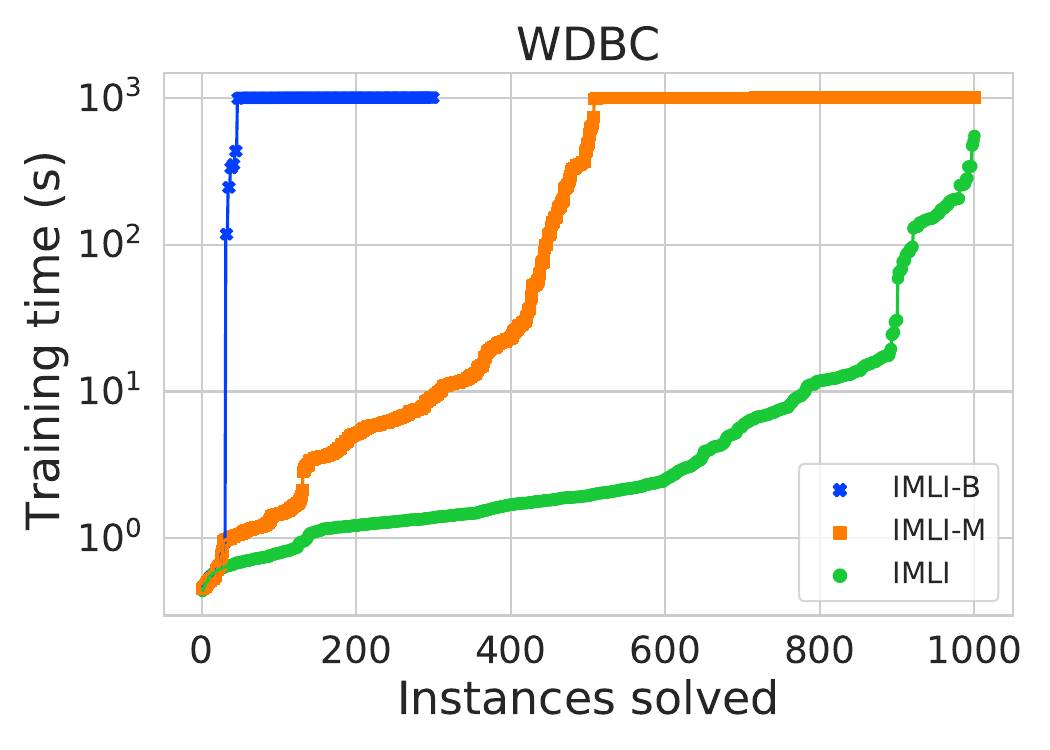}}	
	\subfloat{\includegraphics[scale=0.28]{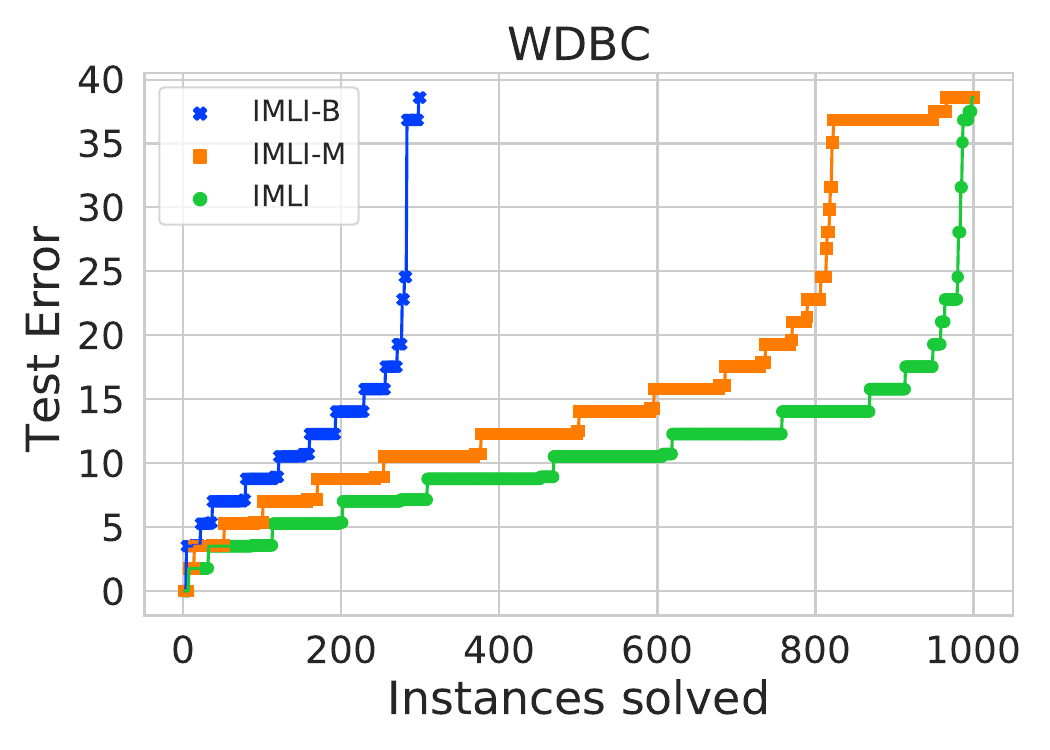}}
	\subfloat{\includegraphics[scale=0.28]{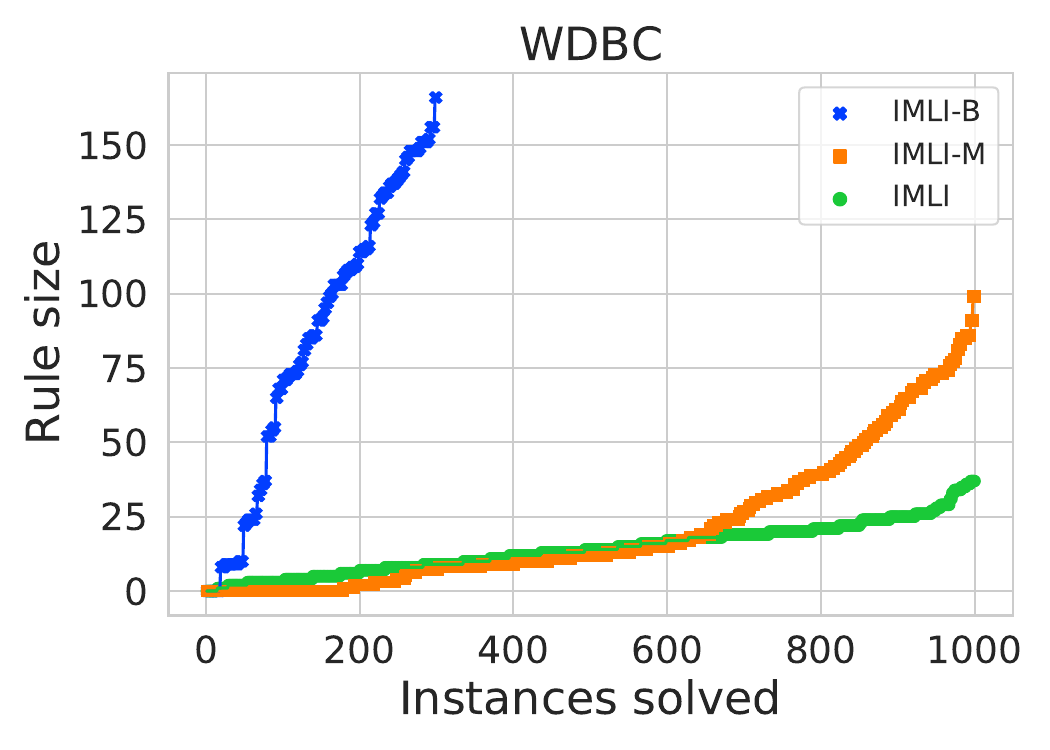}}

	\subfloat{\includegraphics[scale=0.28]{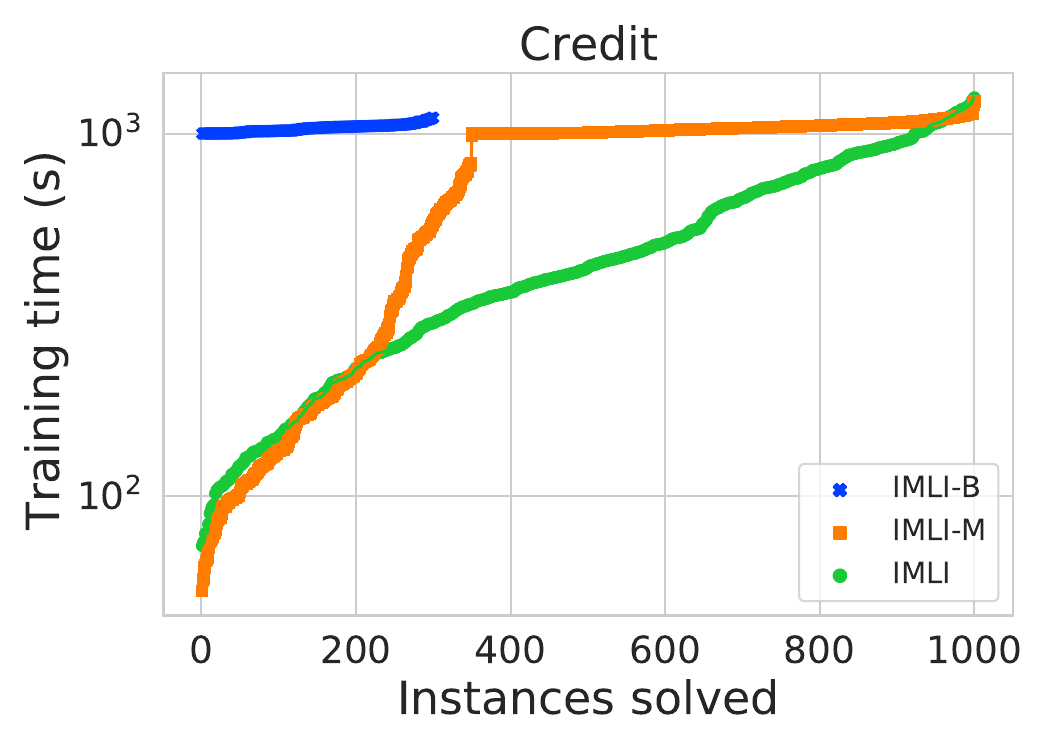}}	
	\subfloat{\includegraphics[scale=0.28]{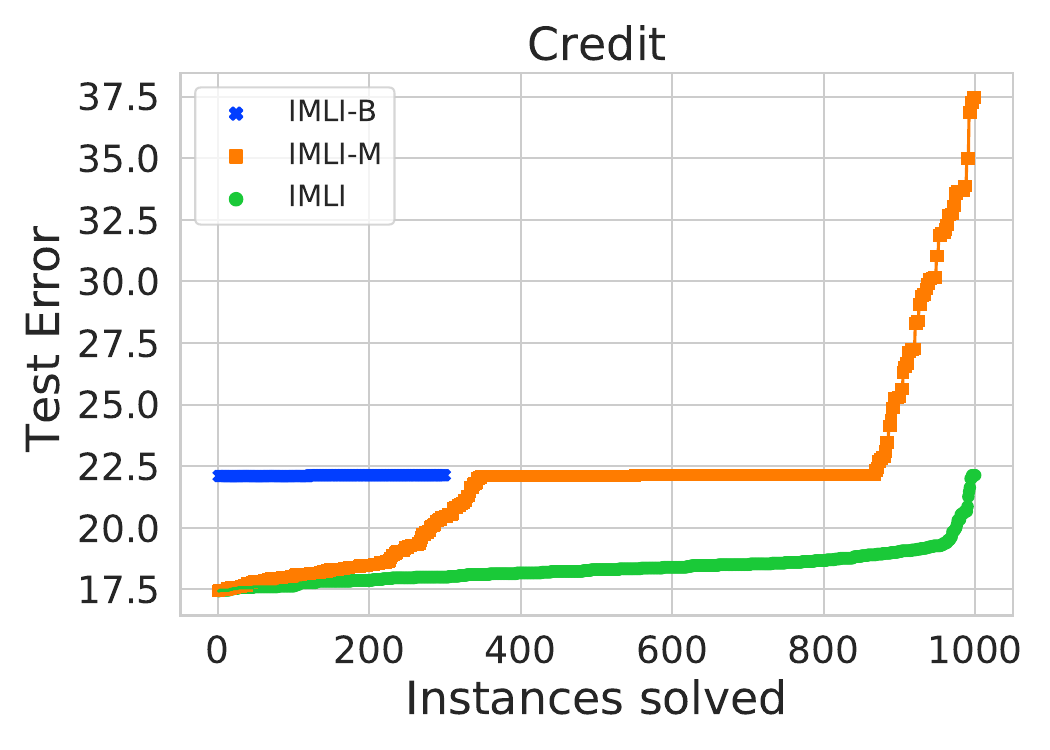}}
	\subfloat{\includegraphics[scale=0.28]{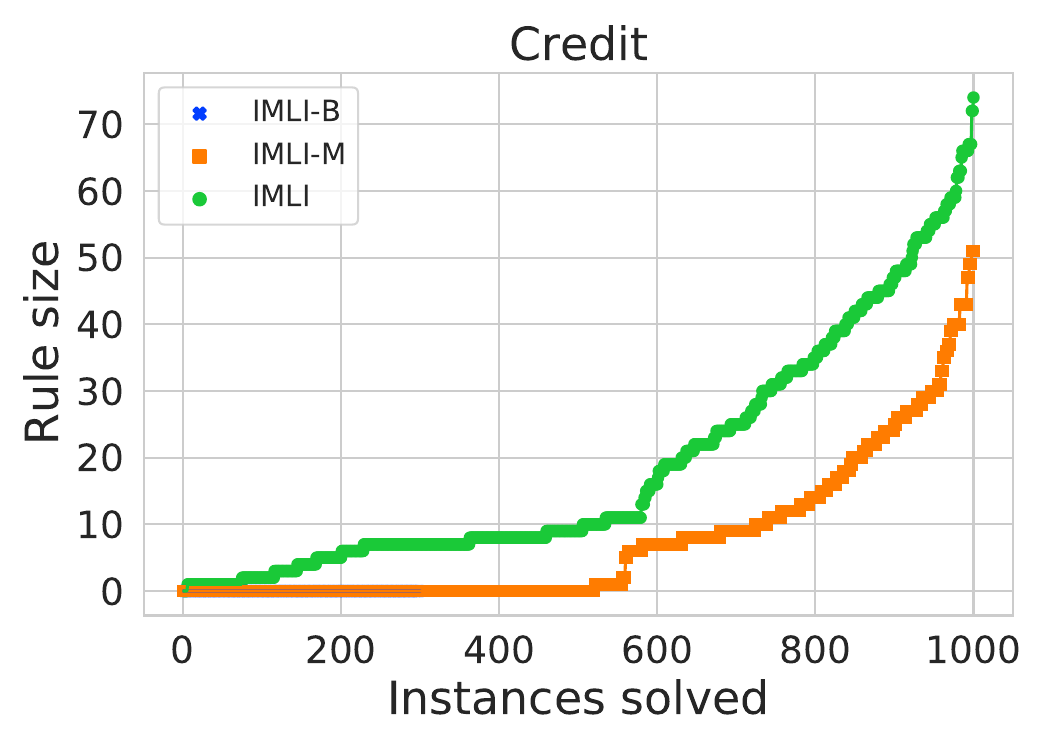}}

	\subfloat{\includegraphics[scale=0.28]{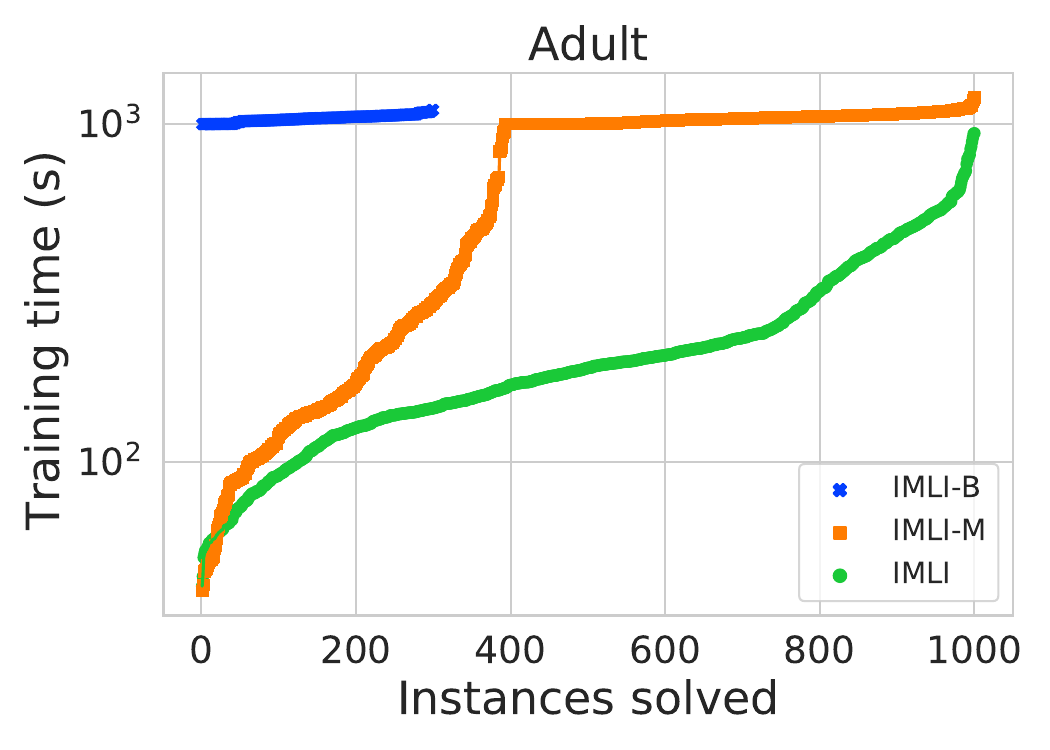}}	
	\subfloat{\includegraphics[scale=0.28]{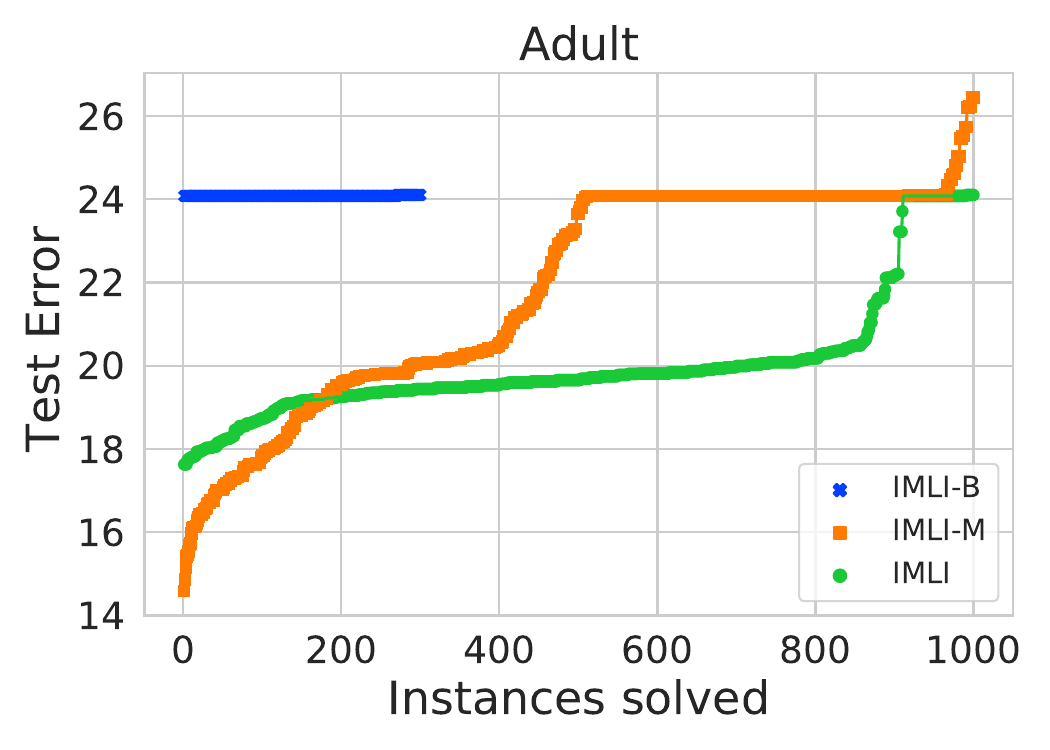}}
	\subfloat{\includegraphics[scale=0.28]{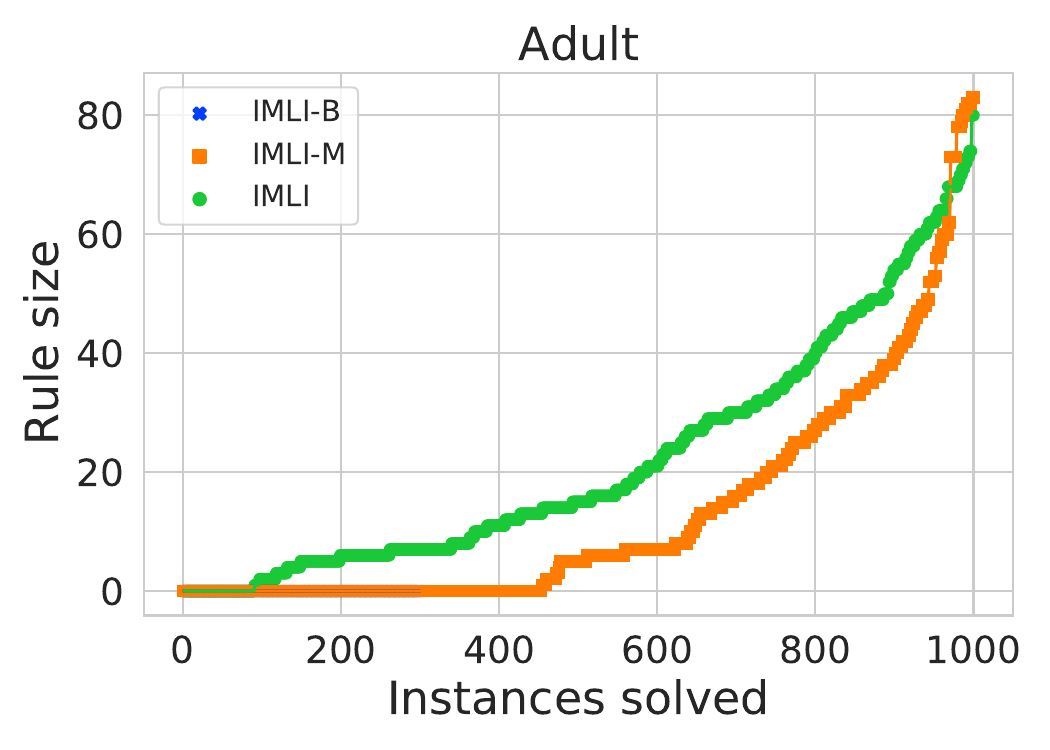}}

	\caption{Comparison of training time, test error, and rule-size among different formulations presented in the paper. In each cactus plot, the incremental formulation {\framework} with both mini-batch and iterative learning demonstrates the best performance in training time and test error than compared two formulations: non-incremental MaxSAT formulation {\framework}-$\mathsf{B}$ and incremental formulation with only mini-batch learning {\framework}-$\mathsf{M}$. In terms of rule-size, {\framework} often generates higher size rules than {\framework}-$\mathsf{M}$.}
	\label{fig:different_formulations}
\end{figure}

\subsubsection{Comparison Among Different Formulations in {\framework}}  We compare the performance of different formulations for learning classification rules as presented in this paper. In Figure~\ref{fig:different_formulations}, we show cactus plots for assessing training time (in seconds), test error (in percentage), and rule-size among different formulations. In the cactus plot, a point $ (x,y) $ denotes that the formulation yields lower than or equal to $ y $ training time (similarly, test error and rule-size) in $ x $ many instances in each dataset.

In Figure~\ref{fig:different_formulations}, we denote the baseline non-incremental MaxSAT-based formulation as {\framework}-$\mathsf{B}$\footnote{Since  {\framework}-$\mathsf{B}$ is a non-incremental formulation and does not involve any mini-batch learning, we consider two hyper-parameters for {\framework}-$\mathsf{B}$: the number of clauses in $ \{1, 2, \dots, 10\} $ and the regularization parameter $ \lambda $ as $ 10 $ values chosen from a logarithmic grid between $ 10^{-4} $ and $ 10^1 $.}, incremental MaxSAT-based formulation with only mini-batch learning as {\framework}-$\mathsf{M}$, and incremental MaxSAT-based formulation with both mini-batch and iterative learning as {\framework}.  We first observe the training time of different formulations in the left-most column in Figure~\ref{fig:different_formulations}, where {\framework}-$\mathsf{B}$ soon times out and solves lower than $ 300 $ instances out of $ 1000 $ instances in each dataset. This result suggests that the non-incremental formulation cannot scale in practical classification task. Comparing between {\framework} and {\framework}-$\mathsf{M}$, both formulations solve all $ 1000 $ instances in each dataset with {\framework}-$\mathsf{M}$ undertaking significantly higher training time than $ {\framework} $.  Therefore, {\framework} achieves better scalability than {\framework}-$\mathsf{M}$ indicating that an integration of mini-batch and iterative learning achieves a significant progress in terms of scalability than mini-batch learning alone. 

We next focus on the test error of different formulations in the middle column in Figure~\ref{fig:different_formulations}. Firstly, {\framework}-$\mathsf{B}$ has a higher test error than the other two formulations since {\framework}-$\mathsf{B}$ times out in most instances and learns a sub-optimal classification rule with reduced prediction accuracy. In contrast, {\framework} has the lowest test error compared to two formulations in all datasets. This result indicates the effectiveness of integrating both iterative and mini-batch learning with MaxSAT-based formulation in generating more accurate classification rules.

Moving focus on the rule-size in the rightmost column in Figure~\ref{fig:different_formulations}, {\framework}-$\mathsf{B}$ achieves the highest rule-size in WDBC dataset. In contrast, the rule-size of {\framework}-$\mathsf{B}$  is lowest (zero) in Credit and Adult datasets. In the last two datasets, {\framework}-$\mathsf{B}$ times out during training and returns the default rule ``true'' by predicting all samples as class $ 1 $. The other two formulations {\framework} and {\framework}-$\mathsf{M}$  demonstrate a similar trend in rule-size in all datasets with {\framework}-$\mathsf{M}$ generating comparatively smaller size rules in Credit and Adult datasets. In this context,  the improvement of rule-sparsity of {\framework}-$\mathsf{M}$ is due to a comparatively higher test error (or lower accuracy) than {\framework} as observed in all three datasets. Therefore, {\framework} appears to be the best performing formulation w.r.t.\ training time, test error, and rule-size by balancing between accuracy and rule-size while being more scalable.

\begin{table*}[!t]        
	\centering
	\caption{Comparison of test accuracy (top value) and rule-size (bottom value) among different rule-based representations learned using {\framework}. Numbers in bold denote the best performing results among different representations.}
	\label{table:different_representations}
	\small
	\begin{tabular}{lrrrrrrrrrrrrrrr}

\toprule
Dataset & CNF & DNF & Decision Sets & Decision Lists \\

\midrule
\multirow{2}{*}{Parkinsons}  &
$ 94.74 $  &  $ 89.47 $  &  $ \mathbf{94.87} $  &  $ 89.74 $  \\
& $ 7.5 $  &  $ \mathbf{6.0} $  &  $ 15.0 $  &  $ 6.5 $  \\
\addlinespace[0.5em]

\multirow{2}{*}{WDBC}  &
$ 94.74 $  &  $ \mathbf{96.49} $  &  $ 95.61 $  &  $ 95.61 $  \\
& $ 11.5 $  &  $ 15.0 $  &  $ 15.5 $  &  $ \mathbf{10.0} $  \\
\addlinespace[0.5em]

\multirow{2}{*}{Pima}  &
$ \mathbf{78.43} $  &  $ 77.13 $  &  $ 76.97 $  &  $ 76.97 $  \\
& $ 23.0 $  &  $ \mathbf{9.0} $  &  $ 15.0 $  &  $ 13.5 $  \\
\addlinespace[0.5em]

\multirow{2}{*}{Titanic}  &
$ 81.82 $  &  $ 82.29 $  &  $ 81.82 $  &  $ \mathbf{82.3} $  \\
& $ \mathbf{5.5} $  &  $ 10.5 $  &  $ 8.5 $  &  $ 8.0 $  \\
\addlinespace[0.5em]

\multirow{2}{*}{MAGIC}  &
$ \mathbf{78.26} $  &  $ 77.44 $  &  $ 75.87 $  &  $ 77.79 $  \\
& $ \mathbf{8.5} $  &  $ 41.5 $  &  $ 10.0 $  &  $ 14.0 $  \\
\addlinespace[0.5em]

\multirow{2}{*}{Tom's HW}  &
$ 85.24 $  &  $ 85.15 $  &  $ 85.72 $  &  $ \mathbf{85.95} $  \\
& $ 44.5 $  &  $ \mathbf{26.5} $  &  $ 45.0 $  &  $ 59.5 $  \\
\addlinespace[0.5em]

\multirow{2}{*}{Credit}  &
$ 82.12 $  &  $ 82.15 $  &  $ 82.03 $  &  $ \mathbf{82.22} $  \\
& $ 17.5 $  &  $ 14.0 $  &  $ \mathbf{9.5} $  &  $ 21.5 $  \\
\addlinespace[0.5em]

\multirow{2}{*}{Adult}  &
$ 81.2 $  &  $ \mathbf{84.28} $  &  $ 80.07 $  &  $ 80.96 $  \\
& $ 30.0 $  &  $ 34.5 $  &  $ \mathbf{7.0} $  &  $ 24.5 $  \\
\addlinespace[0.5em]

\multirow{2}{*}{Bank Marketing}  &
$ \mathbf{89.84} $  &  $ 89.77 $  &  $ 89.67 $  &  $ 89.79 $  \\
& $ 24.5 $  &  $ 7.5 $  &  $ \mathbf{6.0} $  &  $ 10.5 $  \\
\addlinespace[0.5em]

\multirow{2}{*}{Connect-4}  &
$ \mathbf{75.36} $  &  $ 70.63 $  &  $ 68.09 $  &  $ 69.83 $  \\
& $ 50.5 $  &  $ 42.0 $  &  $ \mathbf{4.5} $  &  $ 24.0 $  \\
\addlinespace[0.5em]

\multirow{2}{*}{Weather AUS}  &
$ 83.78 $  &  $ \mathbf{84.23} $  &  $ 83.69 $  &  $ 83.85 $  \\
& $ 22.0 $  &  $ 14.0 $  &  $ \mathbf{4.0} $  &  $ 26.0 $  \\
\addlinespace[0.5em]

\multirow{2}{*}{Skin Seg}  &
$ \mathbf{94.71} $  &  $ 93.68 $  &  $ 87.92 $  &  $ 91.17 $  \\
& $ 30.0 $  &  $ 15.0 $  &  $ \mathbf{3.0} $  &  $ 7.0 $  \\
\bottomrule

	\end{tabular}
	
\end{table*}

\subsubsection{Performance Evaluation of Different Interpretable Representations in {\framework}}

We deploy {\framework} to learn different interpretable rule-based representations: CNF and DNF classifiers, decision lists, and decision sets, and present their comparative performance w.r.t.\ test accuracy and rule-size in Table~\ref{table:different_representations}. In each cell in this table, the top value represents the test accuracy and the bottom value represents the size of generated rules.

We learn all four interpretable representations on twelve datasets, where the CNF classifier appears to be the most accurate representation by achieving the highest accuracy in five datasets. In contrast, both DNF and decision lists achieve the highest accuracy in three datasets each; decision sets demonstrate the least performance in test accuracy by being more accurate in one dataset. To this end, the poor accuracy of decision sets is traded off by its rule-size as decision sets generate the sparsest rules compared to other representations. More precisely, decision sets have the smallest rule-size in six datasets, while CNF, DNF, and decision lists have the smallest rule-size in two, three, and one dataset, respectively. These results suggest that CNF classifiers are more favored in applications where higher accuracy is preferred, while decision sets are preferred in applications where higher interpretability is desired. In both cases, one could deploy {\framework} for learning varied representations of classification rules.

\begin{figure}
	
	\centering
	
	\subfloat{\includegraphics[scale=0.3]{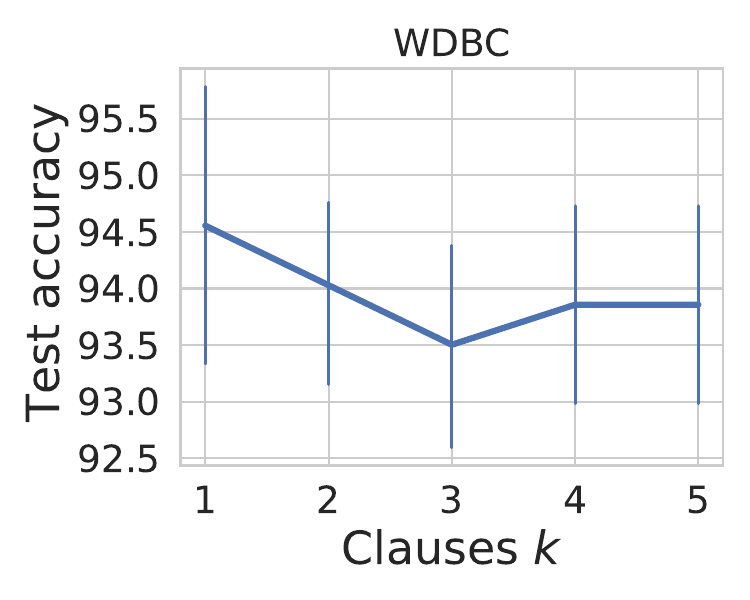}}
	\subfloat{\includegraphics[scale=0.3]{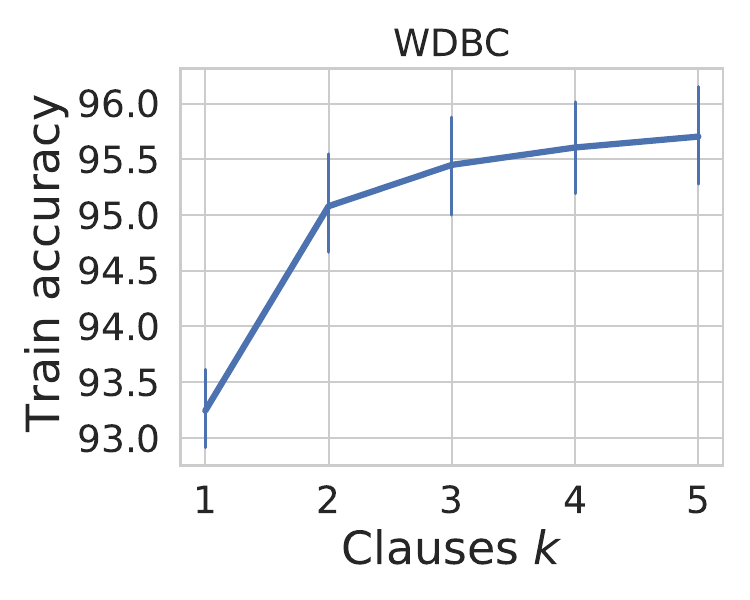}}
	\subfloat{\includegraphics[scale=0.3]{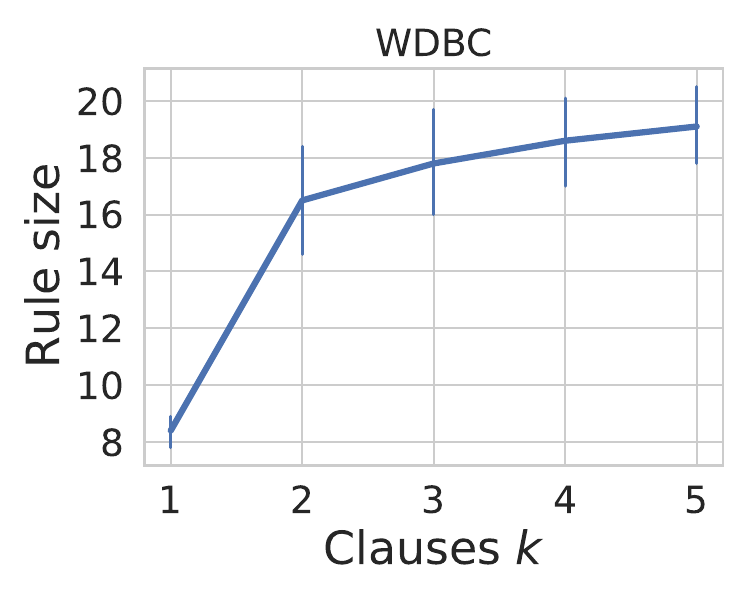}}
	\subfloat{\includegraphics[scale=0.3]{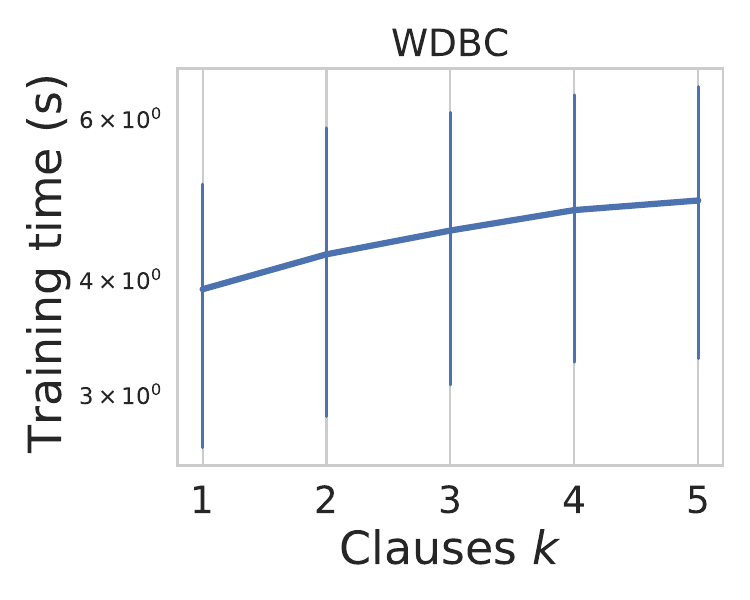}}

	\subfloat{\includegraphics[scale=0.3]{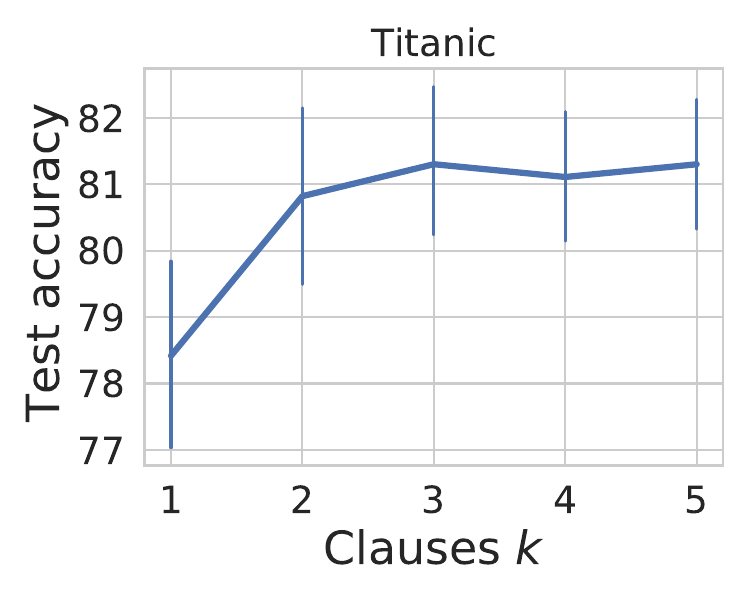}}
	\subfloat{\includegraphics[scale=0.3]{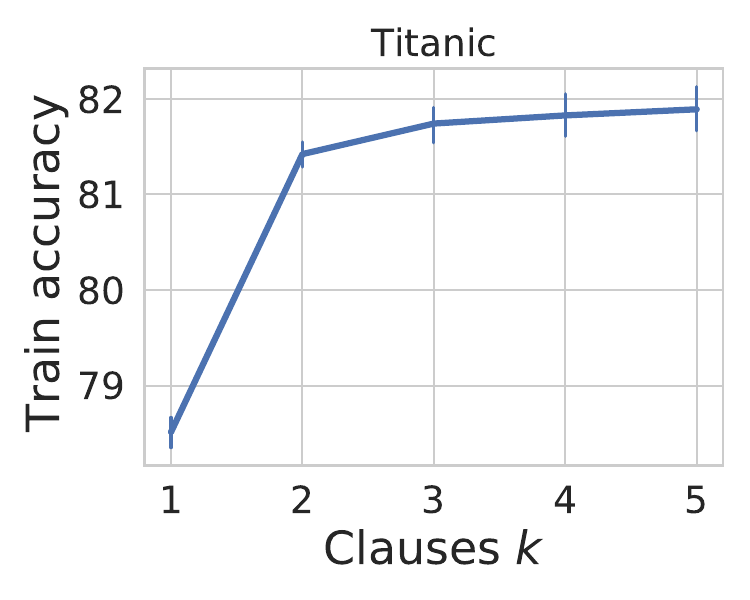}}
	\subfloat{\includegraphics[scale=0.3]{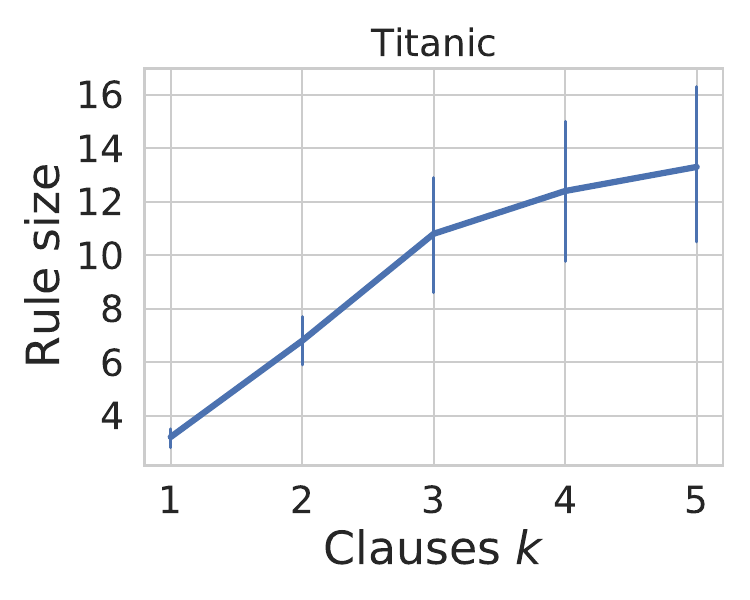}}
	\subfloat{\includegraphics[scale=0.3]{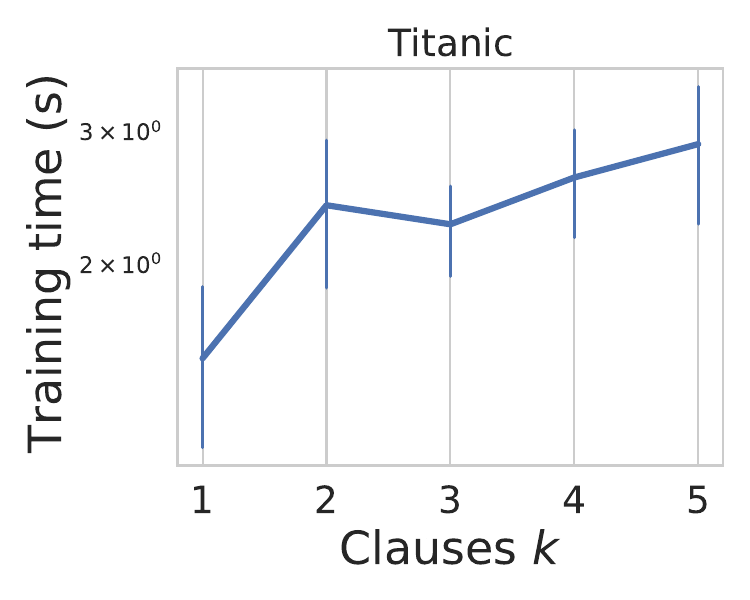}}

	\subfloat{\includegraphics[scale=0.3]{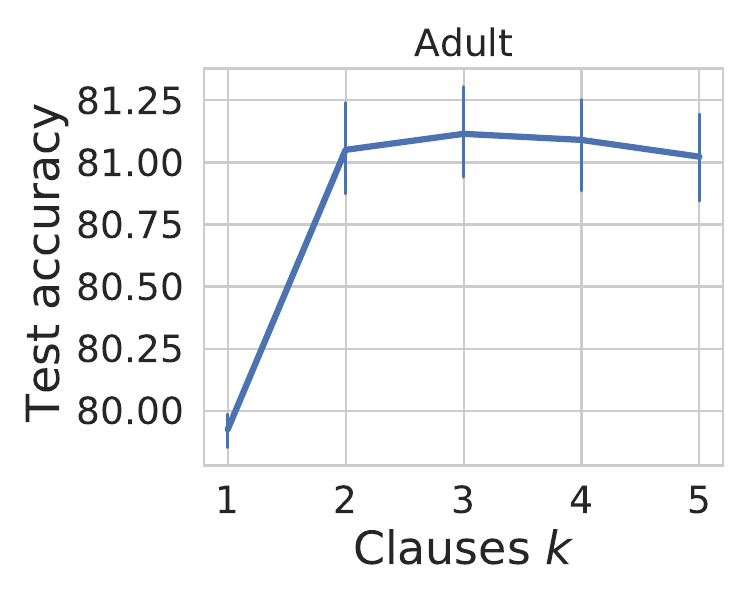}}
	\subfloat{\includegraphics[scale=0.3]{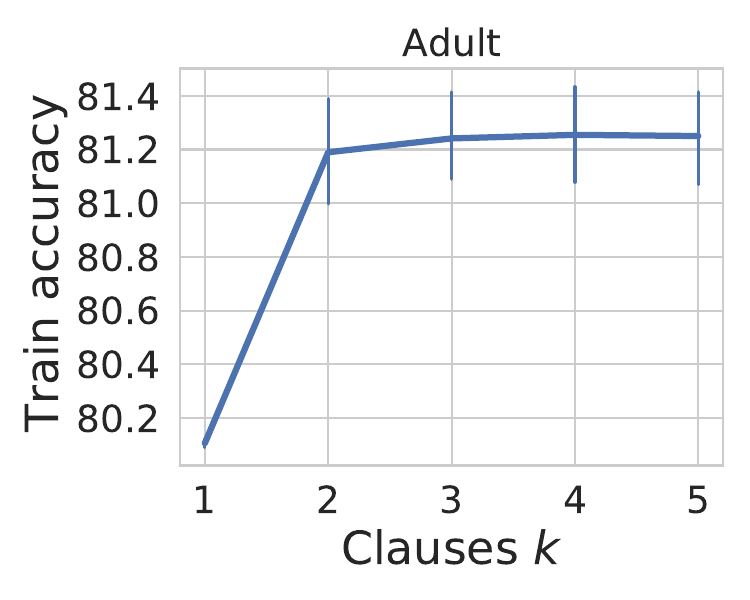}}
	\subfloat{\includegraphics[scale=0.3]{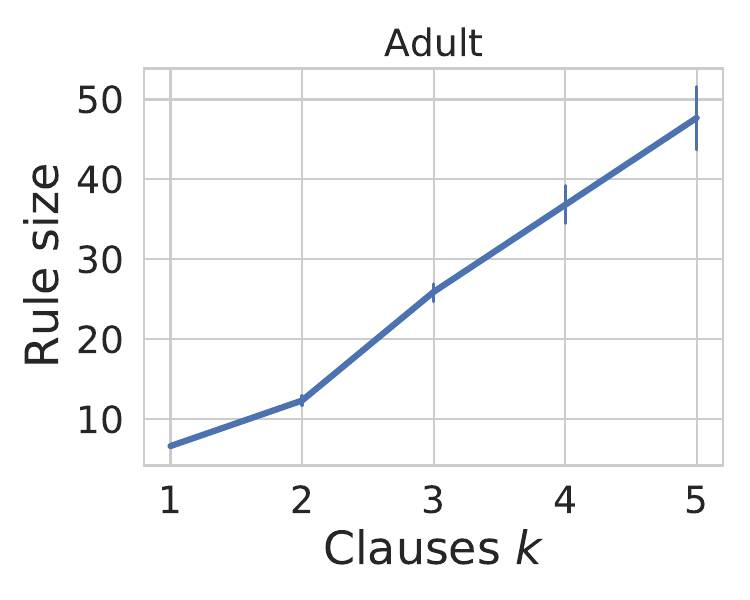}}
	\subfloat{\includegraphics[scale=0.3]{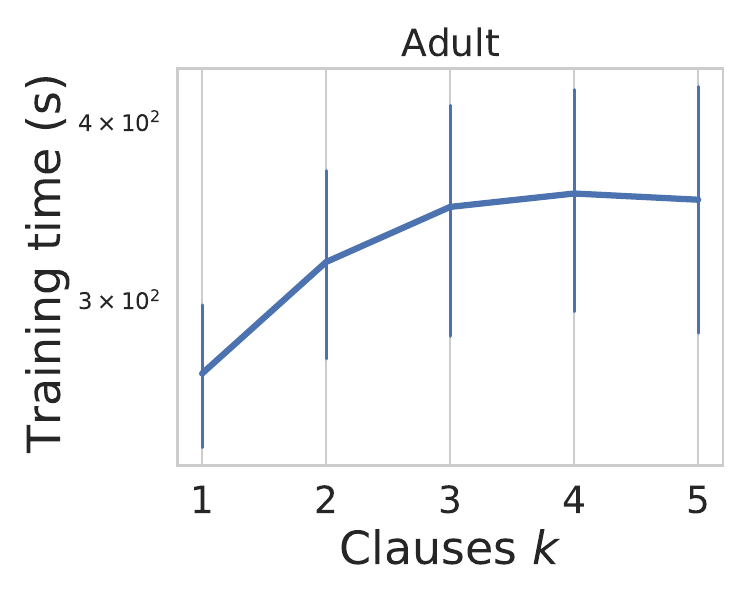}}

	\caption{Effect of the number of clauses $ k $ on accuracy (test and train), rule-size, and training time. As $ k $ increases, both train and test accuracy of {\framework} increase while generating rules with higher size by incurring higher training time. }
	\label{fig:effect_of_number_of_clauses}
\end{figure}

\subsubsection{Ablation Study} We experiment the effect of different hyper-parameters in {\framework} on prediction accuracy, rule-size, and training time in different datasets. In the following, we discuss the impact of the number of clauses, regularization parameter, and size of mini-batches in {\framework}.

\paragraph{Effect of the number of clauses $ k $.} In Figure~\ref{fig:effect_of_number_of_clauses}, we vary  $ k $ while learning CNF classifiers in {\framework}. As $ k $ increases, both training and test accuracy generally increase in different datasets (plots in the first and second columns). Similarly, the size of rules increases with $ k $ by incurring higher training time (plots in the third and fourth columns). The reason is that a higher value of $ k $ allows more flexibility in fitting the data well by incurring more training time and generating higher size classification rules. Therefore, the number of clauses in {\framework} provides control on training-time vs accuracy and also on accuracy vs rule-sparsity. 

\begin{figure}
	
	\centering

	\subfloat{\includegraphics[scale=0.3]{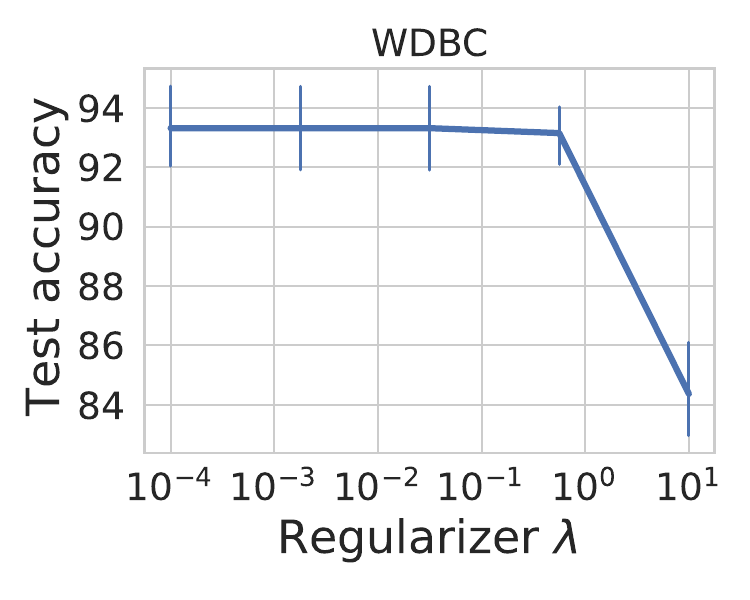}}
	\subfloat{\includegraphics[scale=0.3]{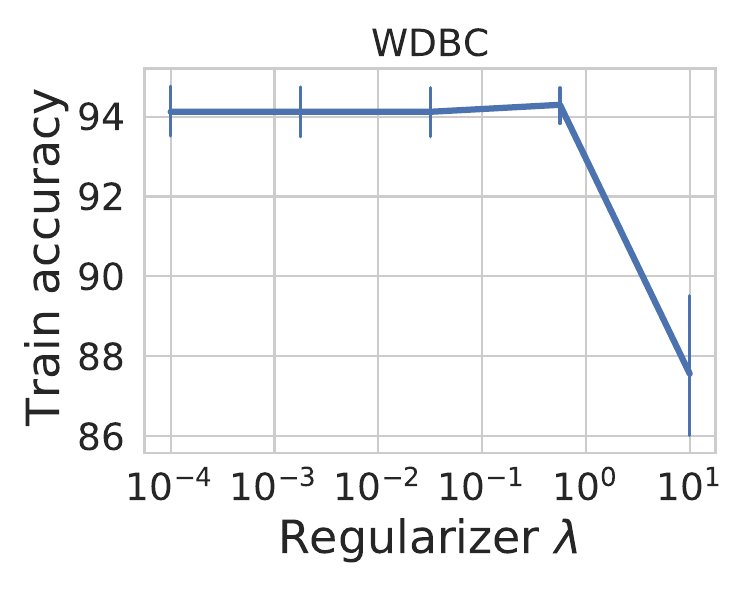}}
	\subfloat{\includegraphics[scale=0.3]{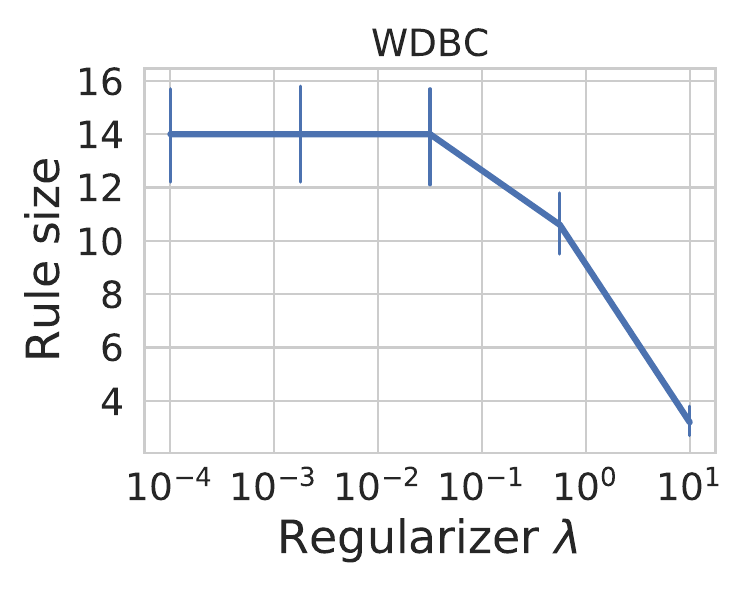}}
	\subfloat{\includegraphics[scale=0.3]{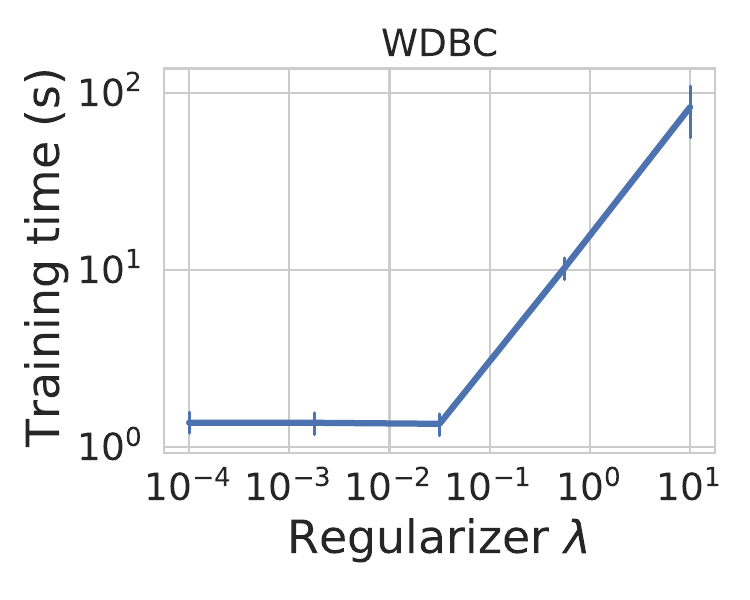}}

	\subfloat{\includegraphics[scale=0.3]{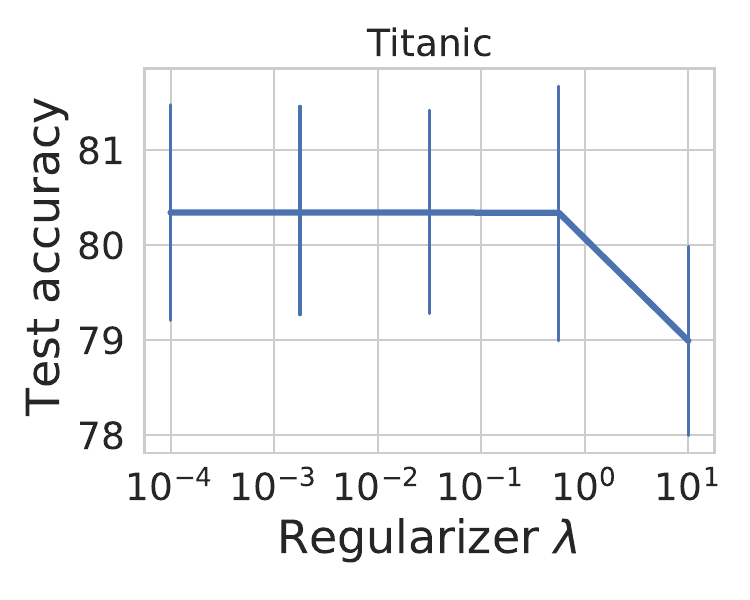}}
	\subfloat{\includegraphics[scale=0.3]{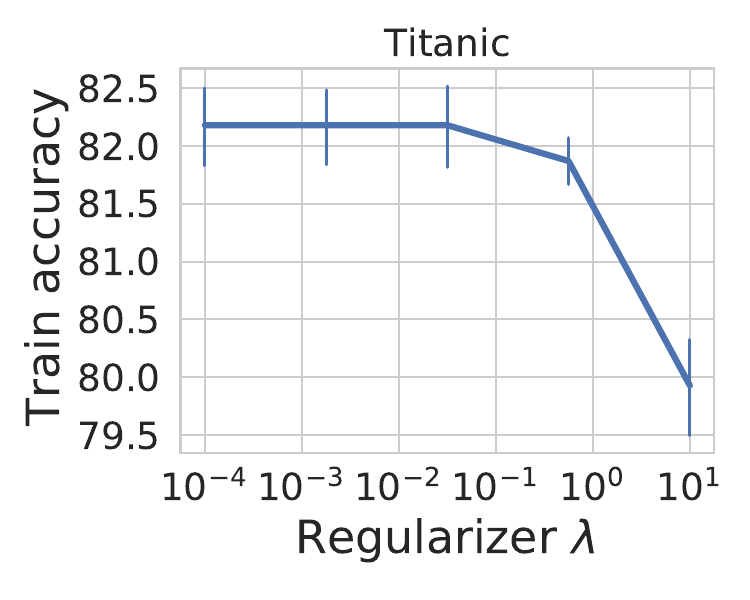}}
	\subfloat{\includegraphics[scale=0.3]{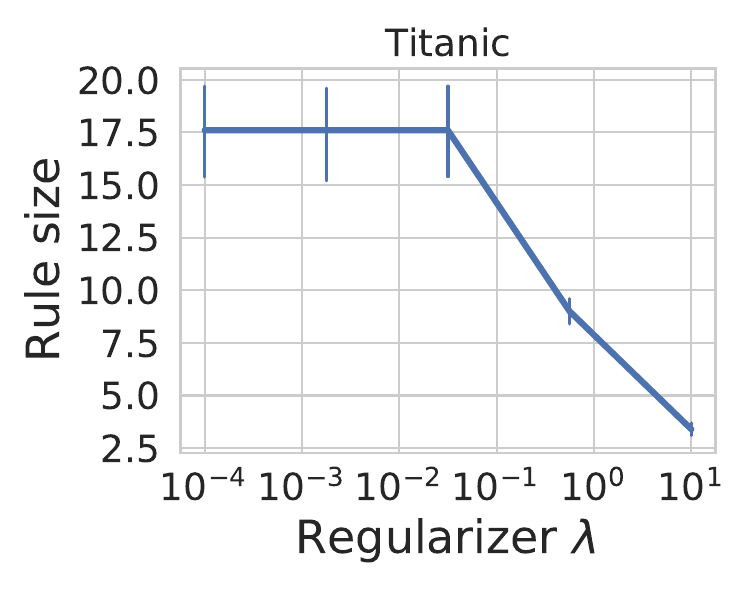}}
	\subfloat{\includegraphics[scale=0.3]{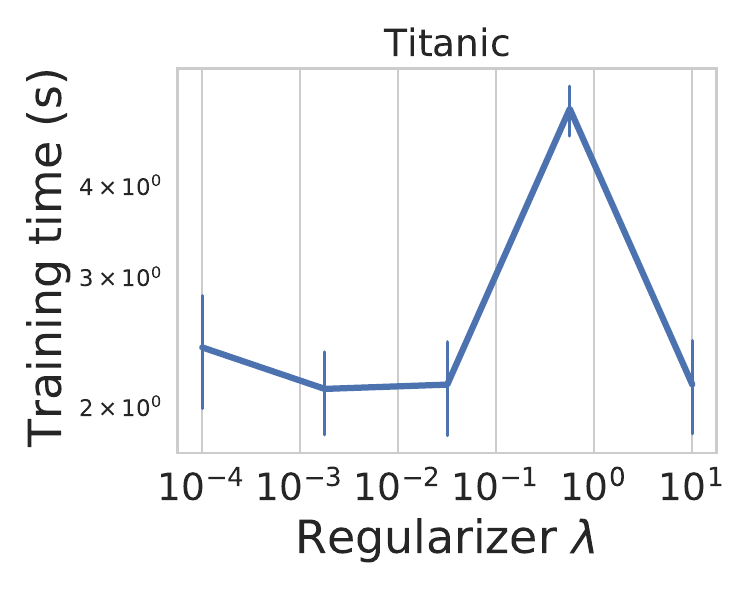}}	
	
	\subfloat{\includegraphics[scale=0.3]{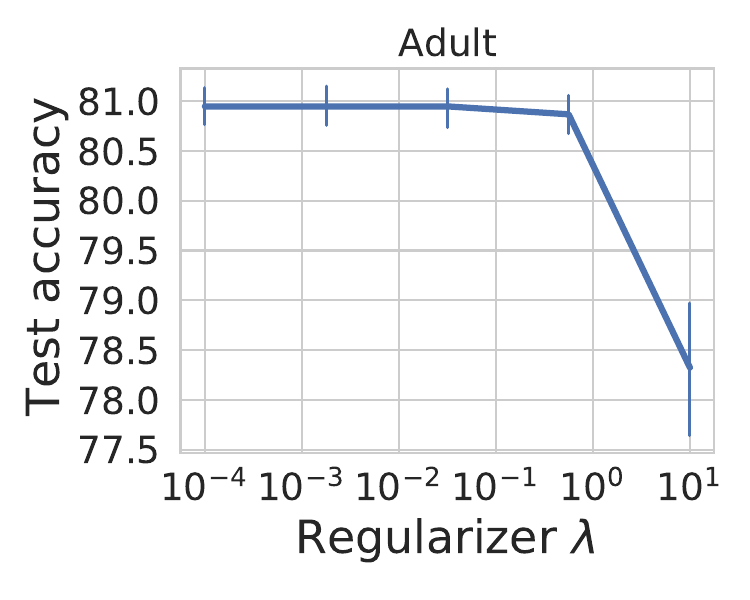}}
	\subfloat{\includegraphics[scale=0.3]{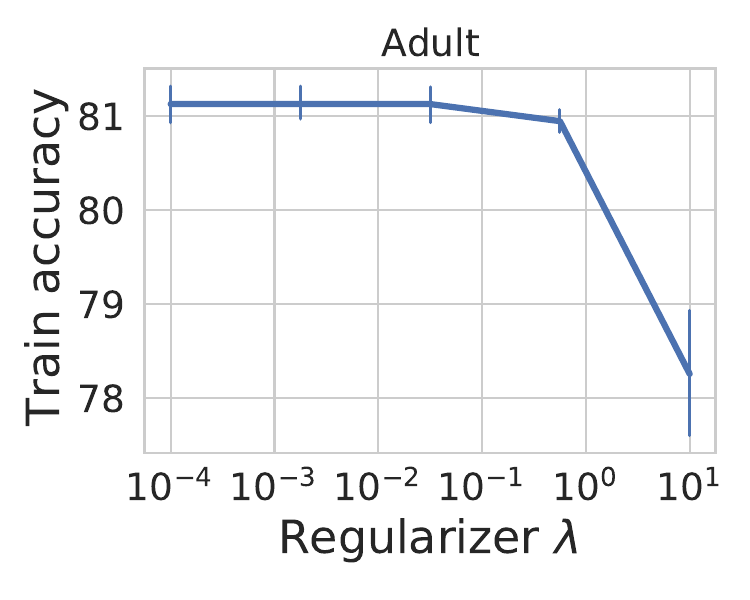}}
	\subfloat{\includegraphics[scale=0.3]{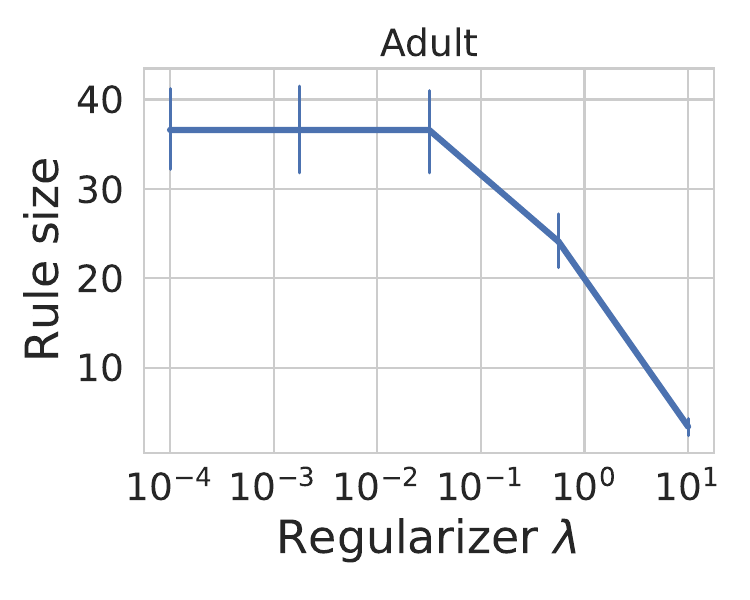}}
	\subfloat{\includegraphics[scale=0.3]{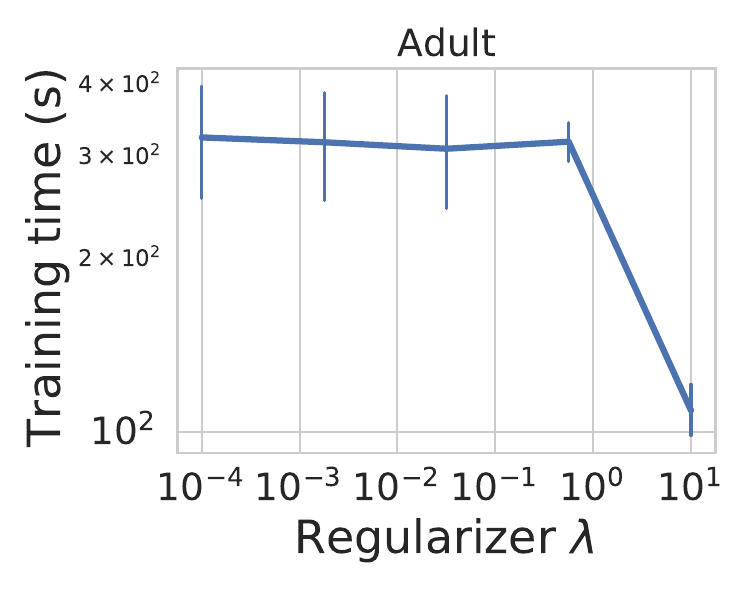}}

	\caption{Effect of regularization $ \lambda $ on accuracy (test and train), rule-size, and training time. As $ \lambda $ increases, lower priority is given to accuracy. As a result, both training and test accuracy decrease with $ \lambda $ by generating smaller rules. }
	\label{fig:effect_of_number_of_regularization}
\end{figure}

\paragraph{Effect of regularizer $ \lambda $.} In Figure~\ref{fig:effect_of_number_of_regularization}, we vary $ \lambda $ in a logarithmic grid between $ 10^{-4} $ and $ 10^1 $. As stated in Eq.~\eqref{eq:obj_incr}, a higher value of $ \lambda $ puts more priority on the minimal changes in rules between consecutive mini-batches in incremental learning while allowing higher mini-batch training errors. Thus, in the first and second columns in Figure~\ref{fig:effect_of_number_of_regularization}, as $ \lambda $ increases, both training and test accuracy gradually decrease. In addition, the size of rules (plots in the third column) also decreases. Finally, we observe that the training time generally decreases with $ \lambda $. This observation indicates that higher $ \lambda $ puts lower computational load to the MaxSAT solver as a fraction of training examples is allowed to be misclassified. Thus, similar to the number of clauses, regularization parameter $ \lambda $ in {\framework} allows to trade-off between accuracy and rule-size in a precise manner.

\begin{figure}
	
	\centering
	
	\subfloat{\includegraphics[scale=0.3]{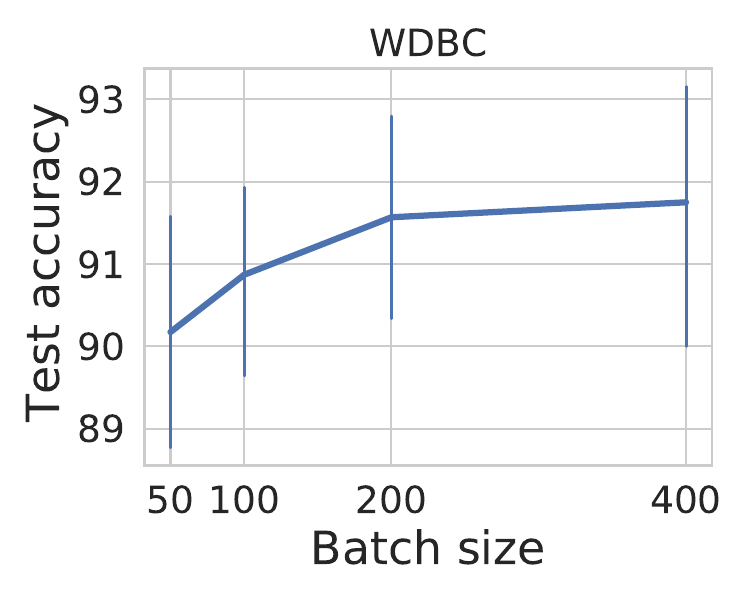}}
	\subfloat{\includegraphics[scale=0.3]{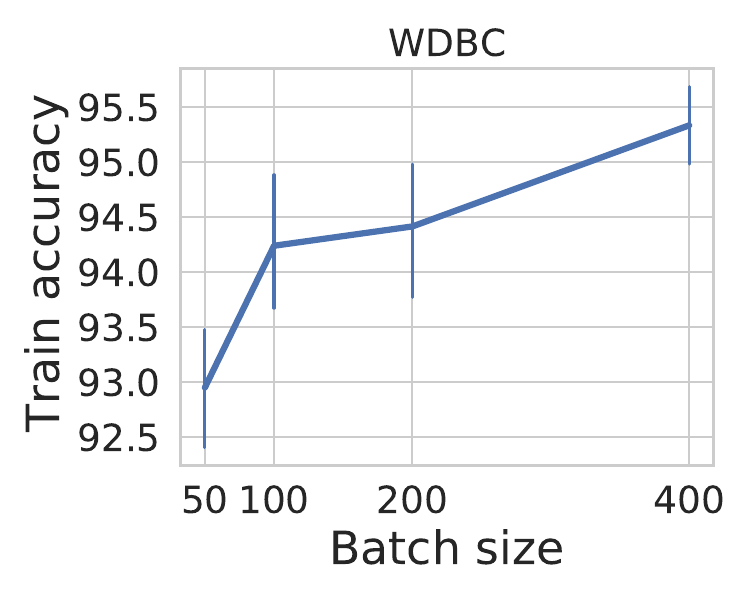}}
	\subfloat{\includegraphics[scale=0.3]{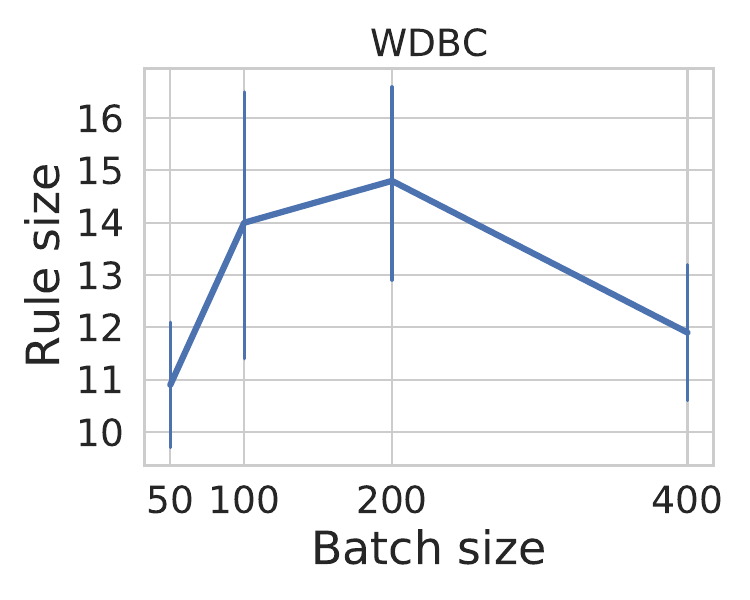}}
	\subfloat{\includegraphics[scale=0.3]{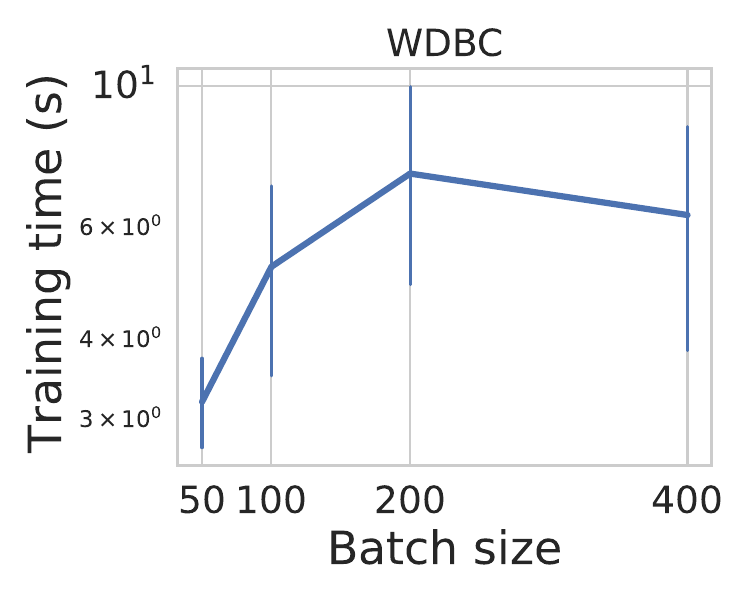}}

	\subfloat{\includegraphics[scale=0.3]{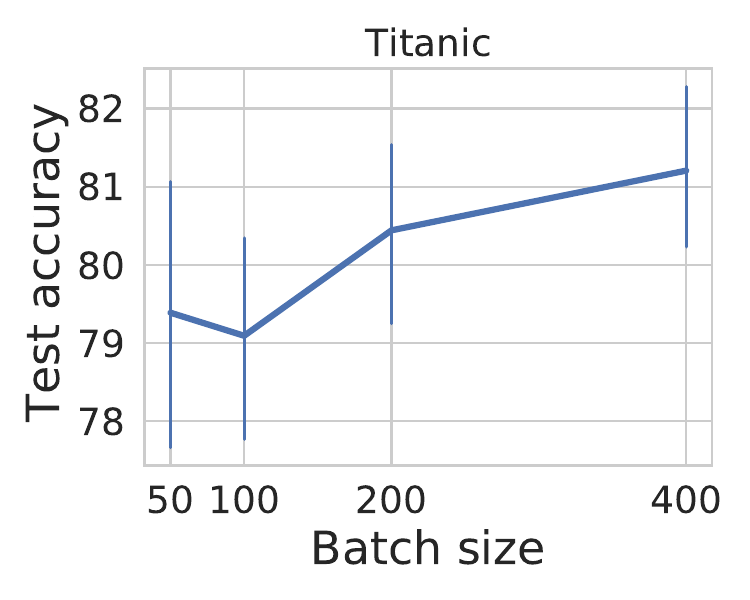}}
	\subfloat{\includegraphics[scale=0.3]{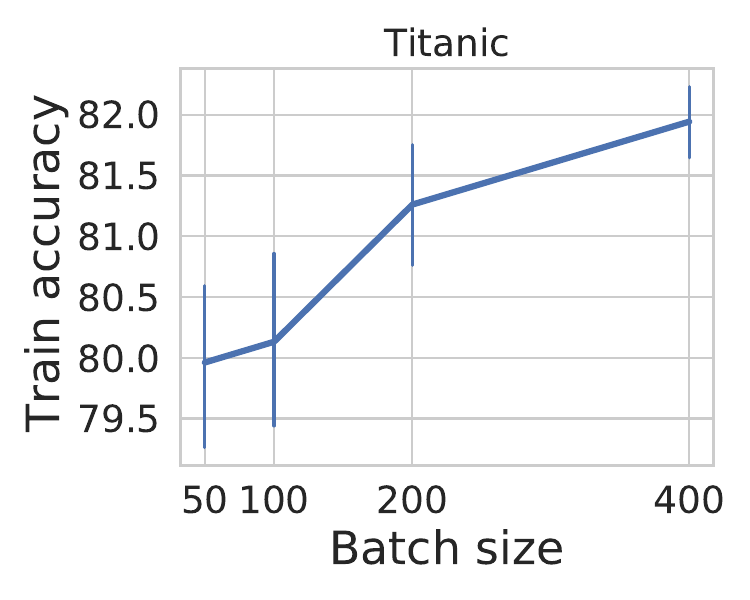}}
	\subfloat{\includegraphics[scale=0.3]{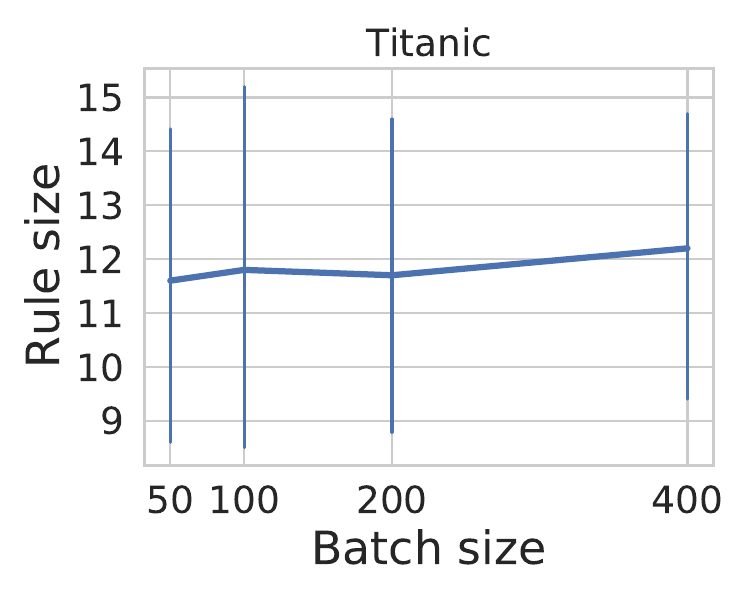}}
	\subfloat{\includegraphics[scale=0.3]{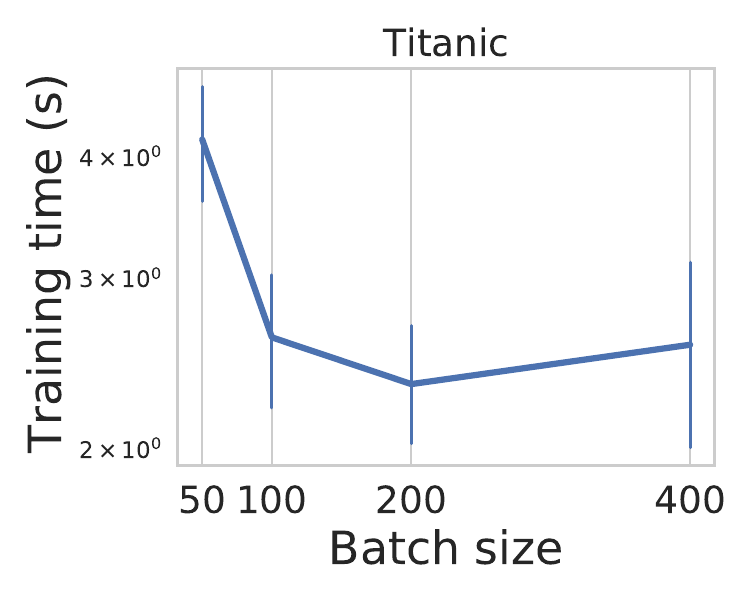}}	
	
	\subfloat{\includegraphics[scale=0.3]{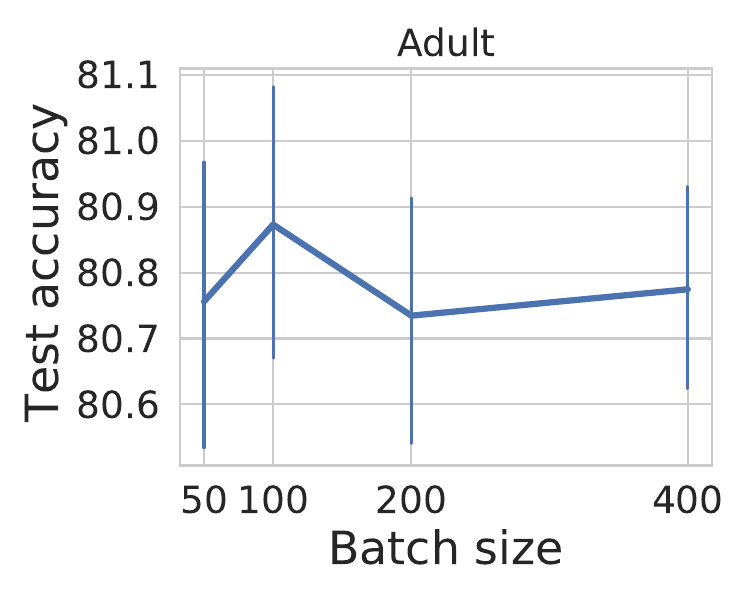}}
	\subfloat{\includegraphics[scale=0.3]{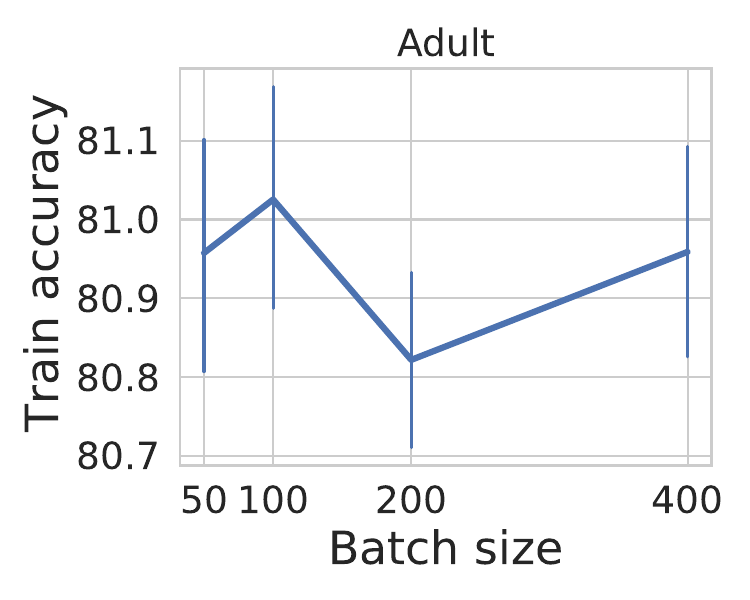}}
	\subfloat{\includegraphics[scale=0.3]{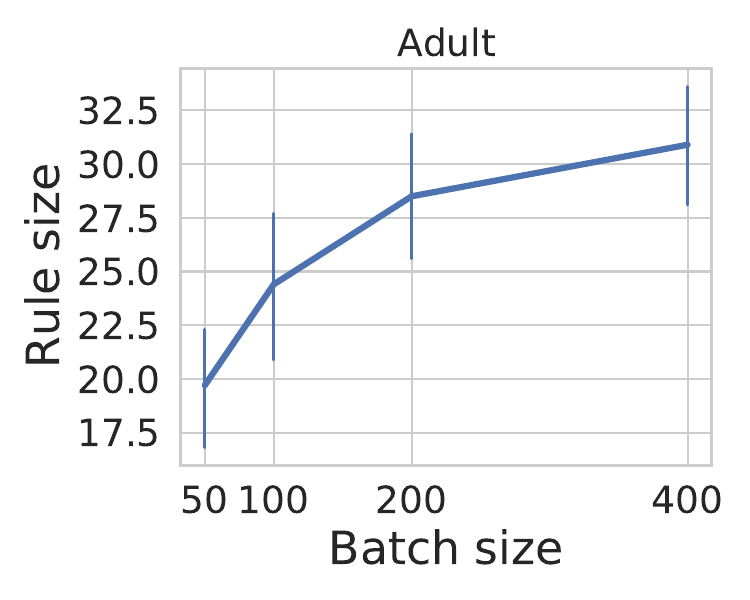}}
	\subfloat{\includegraphics[scale=0.3]{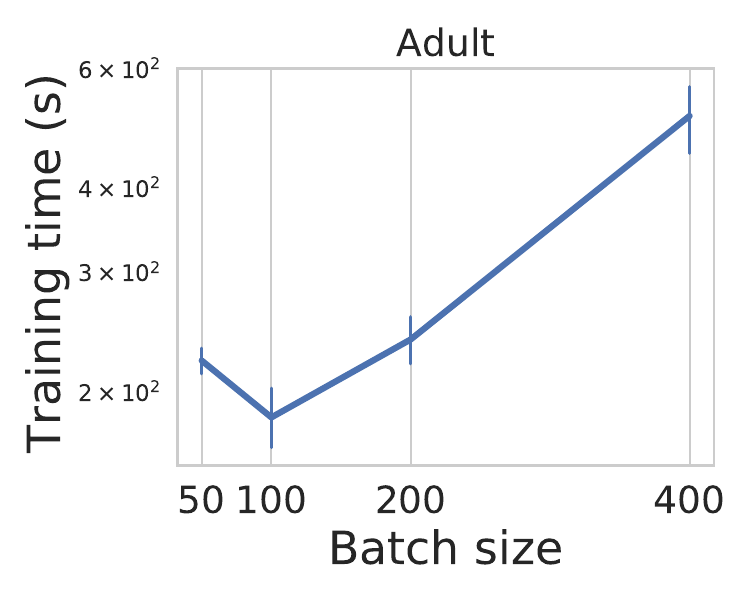}}

	\caption{Effect of bath size on accuracy (test and train), rule-size, and training time. As we consider more samples in a mini-batch, {\framework} generates more accurate and larger size classification rules.}
	\label{fig:effect_of_batch_size}
\end{figure}

\paragraph{Effect of the size of mini-batch.} In Figure~\ref{fig:effect_of_batch_size}, we present the effect of mini-batch size in {\framework}. As we consider more samples in a batch, both test and training accuracy increase in general as presented in the first and second columns in Figure~\ref{fig:effect_of_batch_size}. Similarly, the size of generated rules also increases with the number of samples. Due to solving higher size MaxSAT queries, the training time also increases in general with an increase in mini-batch size. Therefore, by varying the size of mini-batches, {\framework} allows controlling on training time vs the prediction accuracy (and rule-size) of generated rules.

%

\section{Conclusion}
\label{sec:conclusion}
Interpretable machine learning is gaining more focus with applications in many safety-critical domains. Considering the growing demand for interpretable models, it is challenging to design learning frameworks that satisfy all aspects: being accurate, interpretable, and scalable in practical classification tasks.  In this paper, we have proposed a MaxSAT-based framework {\framework} for learning interpretable rule-based classifiers expressible in CNF formulas. {\framework} is built on an efficient integration of incremental learning, specifically mini-batch and iterative learning, with MaxSAT-based formulation.  In our empirical evaluation, {\framework} achieves the best balance among prediction accuracy, interpretability, and scalability. In particular, {\framework} demonstrates competitive prediction accuracy and rule-size compared to existing interpretable rule-based classifiers. In addition, {\framework} achieves impressive scalability than both interpretable and non-interpretable classifiers by learning interpretable rules on million-size datasets with higher accuracy.  Finally, {\framework}  generates other popular interpretable classifiers such as decision lists and decision sets using the same framework.

\section*{Acknowledgments}
Dmitry Malioutov was at the T.J. Watson IBM Research Center while performing this work. This work was supported in part by National Research Foundation Singapore under its NRF Fellowship Programme [NRF-NRFFAI1-2019-0004], Ministry of Education Singapore Tier 2 grant MOE-T2EP20121-0011, and Ministry of Education Singapore Tier 1 Grant [R-252-000-B59-114].  The computational work for this article was performed on resources of Max Planck Institute for Software Systems, Germany and the National Supercomputing Centre, Singapore \url{https://www.nscc.sg}.

\appendix

\section{Illustration of Interpretable CNF Classification Rules}

In the following, we present representative CNF classifiers learned in different datasets. In each dataset, if an input satisfies the CNF formula, it is predicted class $ 1 $ and vice-versa.

\subsubsection*{ Parkinsons}

\noindent 0.3 $\le$ Average vocal fundamental frequency $<$ 0.4 \textbf{OR} 0.5 $\le$ Average vocal fundamental frequency $<$ 0.6 \textbf{OR} 0.1 $\le$ Maximum vocal fundamental frequency $<$ 0.2 \textbf{OR} 0.5 $\le$ Minimum vocal fundamental frequency $<$ 0.6 \textbf{OR} 0.1 $\le$ Shimmer:APQ5 $<$ 0.2 \textbf{OR} 0.5 $\le$ DFA $<$ 0.6 \textbf{OR} 0.3 $\le$ spread1 $<$ 0.4 \textbf{OR} 0.8 $\le$ spread2 $<$ 0.9 \textbf{OR} 0.3 $\le$ PPE $<$ 0.4 \textbf{OR}  \textbf{NOT} -$\infty$ $\le$ MDVP:APQ $<$ 0.1 \blue{\textbf{AND}}\\\\ \textbf{NOT} 0.8 $\le$ Average vocal fundamental frequency $<$ 0.9 \\

\subsubsection*{ WDBC}

\noindent 0.4 $\le$ perimeter $<$ 0.5 \textbf{OR} 0.8 $\le$ symmetry $<$ 0.9 \textbf{OR} 0.5 $\le$ largest concave points $<$ 0.6 \textbf{OR} 0.6 $\le$ largest concave points $<$ 0.7 \textbf{OR} 0.7 $\le$ largest concave points $<$ 0.8 \textbf{OR} 0.6 $\le$ largest symmetry $<$ 0.7 \textbf{OR}  \textbf{NOT} -$\infty$ $\le$ area SE $<$ 0.1 \blue{\textbf{AND}}\\\\0.4 $\le$ texture $<$ 0.5 \textbf{OR} 0.5 $\le$ texture $<$ 0.6 \textbf{OR} 0.5 $\le$ perimeter $<$ 0.6 \textbf{OR} 0.5 $\le$ smoothness $<$ 0.6 \textbf{OR} 0.4 $\le$ concave points $<$ 0.5 \textbf{OR} 0.5 $\le$ concave points $<$ 0.6 \textbf{OR} 0.2 $\le$ largest concavity $<$ 0.3 \textbf{OR} 0.6 $\le$ largest concave points $<$ 0.7 \textbf{OR} 0.7 $\le$ largest concave points $<$ 0.8 \textbf{OR} 0.4 $\le$ largest symmetry $<$ 0.5 \textbf{OR} 0.6 $\le$ largest symmetry $<$ 0.7 \\

\subsubsection*{ Pima}

\noindent x1 = 11 \textbf{OR} x1 = 14 \textbf{OR} x1 = 15 \textbf{OR} 0.7 $\le$ x2 $<$ 0.8 \textbf{OR} 0.8 $\le$ x2 $<$ 0.9 \textbf{OR} 0.9 $\le$ x2 $<$ $\infty$ \textbf{OR} 0.9 $\le$ x3 $<$ $\infty$ \textbf{OR} 0.4 $\le$ x5 $<$ 0.5 \textbf{OR} 0.7 $\le$ x5 $<$ 0.8 \textbf{OR} 0.7 $\le$ x6 $<$ 0.8 \textbf{OR} 0.4 $\le$ x7 $<$ 0.5 \textbf{OR} 0.5 $\le$ x7 $<$ 0.6 \textbf{OR} 0.9 $\le$ x7 $<$ $\infty$ \blue{\textbf{AND}}\\\\x1 = 7 \textbf{OR} x1 = 8 \textbf{OR} 0.6 $\le$ x2 $<$ 0.7 \textbf{OR} 0.8 $\le$ x2 $<$ 0.9 \textbf{OR} 0.9 $\le$ x2 $<$ $\infty$ \textbf{OR} -$\infty$ $\le$ x3 $<$ 0.1 \textbf{OR} 0.7 $\le$ x3 $<$ 0.8 \textbf{OR} 0.9 $\le$ x3 $<$ $\infty$ \textbf{OR} 0.2 $\le$ x5 $<$ 0.3 \textbf{OR} 0.7 $\le$ x6 $<$ 0.8 \textbf{OR} 0.1 $\le$ x7 $<$ 0.2 \textbf{OR} 0.3 $\le$ x7 $<$ 0.4 \textbf{OR} 0.4 $\le$ x8 $<$ 0.5 \textbf{OR} 0.8 $\le$ x8 $<$ 0.9 \blue{\textbf{AND}}\\\\x1 = 2 \textbf{OR} x1 = 6 \textbf{OR} x1 = 7 \textbf{OR} x1 = 9 \textbf{OR} -$\infty$ $\le$ x3 $<$ 0.1 \textbf{OR} 0.8 $\le$ x3 $<$ 0.9 \textbf{OR} 0.1 $\le$ x5 $<$ 0.2 \textbf{OR} 0.2 $\le$ x5 $<$ 0.3 \textbf{OR} 0.7 $\le$ x5 $<$ 0.8 \textbf{OR} 0.5 $\le$ x6 $<$ 0.6 \textbf{OR} 0.6 $\le$ x6 $<$ 0.7 \textbf{OR} 0.4 $\le$ x7 $<$ 0.5 \textbf{OR} 0.9 $\le$ x7 $<$ $\infty$ \textbf{OR} 0.1 $\le$ x8 $<$ 0.2 \textbf{OR} 0.3 $\le$ x8 $<$ 0.4 \textbf{OR} 0.6 $\le$ x8 $<$ 0.7 \textbf{OR} 0.8 $\le$ x8 $<$ 0.9 \blue{\textbf{AND}}\\\\x1 = 8 \textbf{OR} 0.5 $\le$ x2 $<$ 0.6 \textbf{OR} 0.8 $\le$ x2 $<$ 0.9 \textbf{OR} 0.9 $\le$ x2 $<$ $\infty$ \textbf{OR} 0.7 $\le$ x3 $<$ 0.8 \textbf{OR} -$\infty$ $\le$ x4 $<$ 0.1 \textbf{OR} 0.1 $\le$ x5 $<$ 0.2 \textbf{OR} 0.3 $\le$ x5 $<$ 0.4 \textbf{OR} 0.5 $\le$ x7 $<$ 0.6 \textbf{OR} 0.1 $\le$ x8 $<$ 0.2 \textbf{OR} 0.4 $\le$ x8 $<$ 0.5 \textbf{OR} 0.8 $\le$ x8 $<$ 0.9 \\

\subsubsection*{ Titanic}

\noindent -$\infty$ $\le$ age $<$ 0.1 \textbf{OR} 0.9 $\le$ age $<$ $\infty$ \textbf{OR} 0.9 $\le$ fare $<$ $\infty$ \textbf{OR}  \textbf{NOT} sex \blue{\textbf{AND}}\\\\passenger-class = 2 \textbf{OR} 0.4 $\le$ age $<$ 0.5 \textbf{OR} 0.6 $\le$ age $<$ 0.7 \textbf{OR} siblings-or-spouces-aboard = 0 \textbf{OR} 0.1 $\le$ fare $<$ 0.2 \textbf{OR} 0.5 $\le$ fare $<$ 0.6 \textbf{OR} embarked = C \blue{\textbf{AND}}\\\\-$\infty$ $\le$ age $<$ 0.1 \textbf{OR} 0.1 $\le$ age $<$ 0.2 \textbf{OR} 0.7 $\le$ age $<$ 0.8 \textbf{OR} siblings-or-spouces-aboard = 1 \textbf{OR} parents-or-childred-aboard = 1 \textbf{OR} embarked = C \textbf{OR}  \textbf{NOT} passenger-class = 3 \blue{\textbf{AND}}\\\\-$\infty$ $\le$ age $<$ 0.1 \textbf{OR} 0.2 $\le$ age $<$ 0.3 \textbf{OR} 0.3 $\le$ age $<$ 0.4 \textbf{OR} parents-or-childred-aboard = 0 \textbf{OR}  \textbf{NOT} passenger-class = 3 \textbf{OR}  \textbf{NOT} siblings-or-spouces-aboard = 0 \blue{\textbf{AND}}\\\\ \textbf{NOT} passenger-class = 2 \textbf{OR}  \textbf{NOT} 0.7 $\le$ age $<$ 0.8 \\

\subsubsection*{ MAGIC}

\noindent -$\infty$ $\le$ length $<$ 0.1 \textbf{OR} -$\infty$ $\le$ size $<$ 0.1 \textbf{OR} 0.8 $\le$ conc $<$ 0.9 \textbf{OR} -$\infty$ $\le$ alpha $<$ 0.1 \textbf{OR} 0.1 $\le$ alpha $<$ 0.2 \textbf{OR} 0.2 $\le$ dist $<$ 0.3 \\

\subsubsection*{ Tom's HW}

\noindent 0.4 $\le$ x10 $<$ 0.5 \textbf{OR} 0.1 $\le$ x11 $<$ 0.2 \textbf{OR} 0.4 $\le$ x11 $<$ 0.5 \textbf{OR} 0.4 $\le$ x12 $<$ 0.5 \textbf{OR} 0.3 $\le$ x13 $<$ 0.4 \textbf{OR} 0.6 $\le$ x13 $<$ 0.7 \textbf{OR} 0.3 $\le$ x15 $<$ 0.4 \textbf{OR} 0.6 $\le$ x15 $<$ 0.7 \textbf{OR} 0.4 $\le$ x16 $<$ 0.5 \textbf{OR} 0.1 $\le$ x57 $<$ 0.2 \textbf{OR} 0.5 $\le$ x59 $<$ 0.6 \textbf{OR} 0.3 $\le$ x60 $<$ 0.4 \textbf{OR} 0.2 $\le$ x62 $<$ 0.3 \textbf{OR} 0.2 $\le$ x73 $<$ 0.3 \textbf{OR} 0.1 $\le$ x74 $<$ 0.2 \textbf{OR}  \textbf{NOT} -$\infty$ $\le$ x74 $<$ 0.1 \textbf{OR}  \textbf{NOT} -$\infty$ $\le$ x96 $<$ 0.1 \blue{\textbf{AND}}\\\\0.4 $\le$ x9 $<$ 0.5 \textbf{OR} 0.8 $\le$ x9 $<$ 0.9 \textbf{OR} 0.2 $\le$ x12 $<$ 0.3 \textbf{OR} 0.3 $\le$ x12 $<$ 0.4 \textbf{OR} 0.3 $\le$ x15 $<$ 0.4 \textbf{OR} 0.6 $\le$ x15 $<$ 0.7 \textbf{OR} 0.1 $\le$ x60 $<$ 0.2 \textbf{OR} 0.1 $\le$ x61 $<$ 0.2 \textbf{OR} 0.6 $\le$ x78 $<$ 0.7 \textbf{OR}  \textbf{NOT} -$\infty$ $\le$ x11 $<$ 0.1 \textbf{OR}  \textbf{NOT} -$\infty$ $\le$ x13 $<$ 0.1 \textbf{OR}  \textbf{NOT} -$\infty$ $\le$ x46 $<$ 0.1 \textbf{OR}  \textbf{NOT} -$\infty$ $\le$ x64 $<$ 0.1 \textbf{OR}  \textbf{NOT} -$\infty$ $\le$ x72 $<$ 0.1 \textbf{OR}  \textbf{NOT} -$\infty$ $\le$ x77 $<$ 0.1 \blue{\textbf{AND}}\\\\-$\infty$ $\le$ x50 $<$ 0.1 \textbf{OR} 0.1 $\le$ x57 $<$ 0.2 \blue{\textbf{AND}}\\\\-$\infty$ $\le$ x49 $<$ 0.1 \textbf{OR}  \textbf{NOT} -$\infty$ $\le$ x17 $<$ 0.1 \blue{\textbf{AND}}\\\\0.3 $\le$ x14 $<$ 0.4 \textbf{OR} 0.6 $\le$ x74 $<$ 0.7 \textbf{OR} 0.9 $\le$ x79 $<$ $\infty$ \textbf{OR}  \textbf{NOT} -$\infty$ $\le$ x9 $<$ 0.1 \textbf{OR}  \textbf{NOT} -$\infty$ $\le$ x15 $<$ 0.1 \textbf{OR}  \textbf{NOT} -$\infty$ $\le$ x73 $<$ 0.1 \textbf{OR}  \textbf{NOT} -$\infty$ $\le$ x80 $<$ 0.1 \\

\subsubsection*{ Credit}

\noindent Repayment-status-in-September = 2 \textbf{OR} Repayment-status-in-September = 3 \textbf{OR} Repayment-status-in-August = 3 \textbf{OR} Repayment-status-in-May = 3 \textbf{OR} Repayment-status-in-May = 5 \textbf{OR} Repayment-status-in-April = 3 \blue{\textbf{AND}}\\\\0.4 $\le$ Age $<$ 0.5 \textbf{OR} Repayment-status-in-September = 1 \textbf{OR} Repayment-status-in-August = 0 \textbf{OR} Repayment-status-in-July = 2 \textbf{OR} Repayment-status-in-May = 2 \textbf{OR} Repayment-status-in-April = 7 \textbf{OR} 0.3 $\le$ Amount-of-bill-statement-in-April $<$ 0.4 \textbf{OR}  \textbf{NOT} Repayment-status-in-May = -1 \textbf{OR}  \textbf{NOT} -$\infty$ $\le$ Amount-of-bill-statement-in-May $<$ 0.1 \blue{\textbf{AND}}\\\\Gender \textbf{OR} -$\infty$ $\le$ Age $<$ 0.1 \textbf{OR} 0.3 $\le$ Age $<$ 0.4 \textbf{OR} Repayment-status-in-September = 1 \textbf{OR} 0.1 $\le$ Amount-of-bill-statement-in-August $<$ 0.2 \textbf{OR}  \textbf{NOT} Education = 3 \textbf{OR}  \textbf{NOT} 0.2 $\le$ Amount-of-bill-statement-in-September $<$ 0.3 \\

\subsubsection*{ Adult}

\noindent education =  Doctorate \textbf{OR} education =  Prof-school \textbf{OR} 0.1 $\le$ capital-gain $<$ 0.2 \textbf{OR} 0.2 $\le$ capital-gain $<$ 0.3 \textbf{OR} 0.9 $\le$ capital-gain $<$ $\infty$ \textbf{OR} 0.4 $\le$ capital-loss $<$ 0.5 \textbf{OR} 0.5 $\le$ capital-loss $<$ 0.6 \blue{\textbf{AND}}\\\\relationship =  Husband \textbf{OR} relationship =  Wife \textbf{OR} 0.5 $\le$ capital-loss $<$ 0.6 \textbf{OR} 0.6 $\le$ capital-loss $<$ 0.7 \textbf{OR} 0.8 $\le$ capital-loss $<$ 0.9 \textbf{OR}  \textbf{NOT} -$\infty$ $\le$ capital-gain $<$ 0.1 \blue{\textbf{AND}}\\\\0.1 $\le$ age $<$ 0.2 \textbf{OR} 0.7 $\le$ age $<$ 0.8 \textbf{OR} workclass =  State-gov \textbf{OR} education =  Bachelors \textbf{OR} marital-status =  Separated \textbf{OR} occupation =  Exec-managerial \textbf{OR} occupation =  Farming-fishing \textbf{OR} occupation =  Prof-specialty \textbf{OR} occupation =  Protective-serv \textbf{OR} occupation =  Tech-support \textbf{OR} relationship =  Unmarried \textbf{OR} 0.4 $\le$ capital-loss $<$ 0.5 \textbf{OR} 0.6 $\le$ capital-loss $<$ 0.7 \textbf{OR} 0.2 $\le$ hours-per-week $<$ 0.3 \textbf{OR} 0.4 $\le$ hours-per-week $<$ 0.5 \textbf{OR} 0.7 $\le$ hours-per-week $<$ 0.8 \textbf{OR} 0.9 $\le$ hours-per-week $<$ $\infty$ \textbf{OR}  \textbf{NOT} -$\infty$ $\le$ capital-gain $<$ 0.1 \\

\subsubsection*{ Bank Marketing}

\noindent 0.7 $\le$ age $<$ 0.8 \textbf{OR} 0.2 $\le$ duration $<$ 0.3 \textbf{OR} 0.3 $\le$ duration $<$ 0.4 \textbf{OR} 0.4 $\le$ duration $<$ 0.5 \textbf{OR} 0.5 $\le$ duration $<$ 0.6 \textbf{OR} poutcome = success \blue{\textbf{AND}}\\\\housing \textbf{OR} 0.8 $\le$ age $<$ 0.9 \textbf{OR} job = admin. \textbf{OR} job = management \textbf{OR} job = self-employed \textbf{OR} job = services \textbf{OR} job = unemployed \textbf{OR} 0.2 $\le$ duration $<$ 0.3 \textbf{OR} 0.1 $\le$ campaign $<$ 0.2 \textbf{OR} poutcome = success \blue{\textbf{AND}}\\\\job = services \textbf{OR} 0.2 $\le$ balance $<$ 0.3 \textbf{OR} contact = cellular \textbf{OR} 0.1 $\le$ duration $<$ 0.2 \textbf{OR} 0.2 $\le$ duration $<$ 0.3 \textbf{OR} 0.2 $\le$ pdays $<$ 0.3 \textbf{OR} poutcome = unknown \textbf{OR}  \textbf{NOT} -$\infty$ $\le$ balance $<$ 0.1 \textbf{OR}  \textbf{NOT} -$\infty$ $\le$ previous $<$ 0.1 \blue{\textbf{AND}}\\\\0.2 $\le$ age $<$ 0.3 \textbf{OR} 0.3 $\le$ age $<$ 0.4 \textbf{OR} 0.8 $\le$ age $<$ 0.9 \textbf{OR} job = management \textbf{OR} job = student \textbf{OR} job = unemployed \textbf{OR} education = secondary \textbf{OR} 0.1 $\le$ balance $<$ 0.2 \textbf{OR} 0.1 $\le$ duration $<$ 0.2 \textbf{OR} -$\infty$ $\le$ pdays $<$ 0.1 \textbf{OR} 0.2 $\le$ pdays $<$ 0.3 \textbf{OR} poutcome = unknown \\

\subsubsection*{ Connect-4}

\noindent b2 = 1 \textbf{OR} c2 = 1 \textbf{OR} d2 = 1 \textbf{OR} d4 = 1 \textbf{OR} d5 = 0 \textbf{OR} e2 = 1 \textbf{OR} f3 = 0 \textbf{OR}  \textbf{NOT} d1 = 0 \blue{\textbf{AND}}\\\\b2 = 1 \textbf{OR} b3 = 1 \textbf{OR} b4 = 1 \textbf{OR} d4 = 1 \textbf{OR} f2 = 1 \textbf{OR}  \textbf{NOT} d3 = 0 \textbf{OR}  \textbf{NOT} f3 = 2 \blue{\textbf{AND}}\\\\c1 = 2 \textbf{OR} c2 = 2 \textbf{OR} c3 = 1 \textbf{OR} c3 = 2 \textbf{OR} c4 = 1 \textbf{OR} c6 = 0 \textbf{OR} d2 = 1 \textbf{OR} d4 = 1 \textbf{OR} e2 = 1 \textbf{OR} e3 = 1 \textbf{OR} f4 = 1 \textbf{OR}  \textbf{NOT} a3 = 2 \textbf{OR}  \textbf{NOT} b6 = 2 \blue{\textbf{AND}}\\\\a1 = 0 \textbf{OR} a2 = 0 \textbf{OR} a6 = 0 \textbf{OR} b2 = 1 \textbf{OR} b4 = 1 \textbf{OR} b5 = 0 \textbf{OR} c2 = 1 \textbf{OR} c4 = 1 \textbf{OR} c5 = 0 \textbf{OR} d1 = 1 \textbf{OR} d2 = 1 \textbf{OR} e2 = 1 \textbf{OR} g1 = 0 \textbf{OR} g2 = 0 \textbf{OR}  \textbf{NOT} d5 = 2 \textbf{OR}  \textbf{NOT} d6 = 2 \blue{\textbf{AND}}\\\\b2 = 1 \textbf{OR} b4 = 1 \textbf{OR} c3 = 1 \textbf{OR} d3 = 1 \textbf{OR} e2 = 1 \textbf{OR} f2 = 1 \textbf{OR} f3 = 1 \textbf{OR} g3 = 0 \textbf{OR} g5 = 0 \textbf{OR}  \textbf{NOT} c2 = 0 \textbf{OR}  \textbf{NOT} c4 = 2 \textbf{OR}  \textbf{NOT} d2 = 2 \\

\subsubsection*{ Weather AUS}

\noindent 0.7 $\le$ Rainfall $<$ 0.8 \textbf{OR} 0.8 $\le$ Humidity3pm $<$ 0.9 \textbf{OR} 0.9 $\le$ Humidity3pm $<$ $\infty$ \blue{\textbf{AND}}\\\\RainToday \textbf{OR} 0.5 $\le$ WindGustSpeed $<$ 0.6 \textbf{OR} 0.7 $\le$ Humidity9am $<$ 0.8 \textbf{OR} 0.9 $\le$ Humidity3pm $<$ $\infty$ \textbf{OR} 0.4 $\le$ Pressure3pm $<$ 0.5 \textbf{OR}  \textbf{NOT} 0.9 $\le$ Humidity9am $<$ $\infty$ \blue{\textbf{AND}}\\\\0.1 $\le$ MinTemp $<$ 0.2 \textbf{OR} 0.8 $\le$ MinTemp $<$ 0.9 \textbf{OR} 0.9 $\le$ WindSpeed9am $<$ $\infty$ \textbf{OR} 0.9 $\le$ Humidity9am $<$ $\infty$ \textbf{OR} 0.8 $\le$ Humidity3pm $<$ 0.9 \textbf{OR} 0.9 $\le$ Humidity3pm $<$ $\infty$ \textbf{OR} 0.5 $\le$ Temp3pm $<$ 0.6 \textbf{OR}  \textbf{NOT} 0.7 $\le$ Rainfall $<$ 0.8 \blue{\textbf{AND}}\\\\WindGustDir = NNW \textbf{OR} WindDir3pm = W \textbf{OR}  \textbf{NOT} 0.8 $\le$ Temp3pm $<$ 0.9 \blue{\textbf{AND}}\\\\WindGustDir = NNW \textbf{OR} WindGustDir = NW \textbf{OR} 0.1 $\le$ Pressure9am $<$ 0.2 \textbf{OR}  \textbf{NOT} 0.7 $\le$ Temp3pm $<$ 0.8 \\

\subsubsection*{ Vote}

\noindent  \textbf{NOT} physician-fee-freeze \blue{\textbf{AND}}\\\\ \textbf{NOT} adoption-of-the-budget-resolution \textbf{OR}  \textbf{NOT} anti-satellite-test-ban \textbf{OR}  \textbf{NOT} synfuels-corporation-cutback \blue{\textbf{AND}}\\\\adoption-of-the-budget-resolution \textbf{OR} el-salvador-aid \textbf{OR}  \textbf{NOT} duty-free-exports \blue{\textbf{AND}}\\\\mx-missile \textbf{OR}  \textbf{NOT} adoption-of-the-budget-resolution \textbf{OR}  \textbf{NOT} el-salvador-aid \textbf{OR}  \textbf{NOT} anti-satellite-test-ban \blue{\textbf{AND}}\\\\adoption-of-the-budget-resolution \textbf{OR} mx-missile \textbf{OR}  \textbf{NOT} synfuels-corporation-cutback \textbf{OR}  \textbf{NOT} education-spending \\

\subsubsection*{ Skin Seg}

\noindent 0.2 $\le$ Red $<$ 0.3 \textbf{OR} 0.3 $\le$ Red $<$ 0.4 \textbf{OR} 0.4 $\le$ Red $<$ 0.5 \textbf{OR} 0.9 $\le$ Red $<$ $\infty$ \textbf{OR} 0.9 $\le$ Green $<$ $\infty$ \textbf{OR} 0.4 $\le$ Blue $<$ 0.5 \textbf{OR}  \textbf{NOT} 0.9 $\le$ Blue $<$ $\infty$ \blue{\textbf{AND}}\\\\0.2 $\le$ Red $<$ 0.3 \textbf{OR} 0.7 $\le$ Red $<$ 0.8 \textbf{OR} 0.8 $\le$ Red $<$ 0.9 \textbf{OR} 0.9 $\le$ Red $<$ $\infty$ \textbf{OR}  \textbf{NOT} 0.8 $\le$ Blue $<$ 0.9 \blue{\textbf{AND}}\\\\0.7 $\le$ Red $<$ 0.8 \textbf{OR} 0.8 $\le$ Red $<$ 0.9 \textbf{OR} 0.9 $\le$ Red $<$ $\infty$ \textbf{OR} -$\infty$ $\le$ Green $<$ 0.1 \textbf{OR}  \textbf{NOT} 0.7 $\le$ Blue $<$ 0.8 \\

\subsubsection*{ BNG(labor)}

\noindent 0.5 $\le$ wage-increase-first-year $<$ 0.6 \textbf{OR} pension = empl-contr \textbf{OR} contribution-to-dental-plan = full \blue{\textbf{AND}}\\\\0.6 $\le$ wage-increase-first-year $<$ 0.7 \textbf{OR} 0.7 $\le$ wage-increase-second-year $<$ 0.8 \textbf{OR} 0.5 $\le$ shift-differential $<$ 0.6 \textbf{OR}  \textbf{NOT} longterm-disability-assistance \blue{\textbf{AND}}\\\\0.2 $\le$ wage-increase-first-year $<$ 0.3 \textbf{OR} 0.3 $\le$ wage-increase-first-year $<$ 0.4 \textbf{OR} 0.4 $\le$ wage-increase-first-year $<$ 0.5 \textbf{OR} 0.5 $\le$ wage-increase-first-year $<$ 0.6 \textbf{OR} 0.6 $\le$ wage-increase-first-year $<$ 0.7 \textbf{OR} 0.7 $\le$ wage-increase-first-year $<$ 0.8 \textbf{OR} 0.6 $\le$ wage-increase-second-year $<$ 0.7 \textbf{OR} cost-of-living-adjustment = tcf \textbf{OR} 0.3 $\le$ working-hours $<$ 0.4 \textbf{OR} 0.4 $\le$ working-hours $<$ 0.5 \textbf{OR} 0.5 $\le$ working-hours $<$ 0.6 \textbf{OR} pension = ret-allw \textbf{OR} 0.7 $\le$ standby-pay $<$ 0.8 \textbf{OR} contribution-to-dental-plan = full \textbf{OR} contribution-to-health-plan = half \textbf{OR}  \textbf{NOT} education-allowance \blue{\textbf{AND}}\\\\0.5 $\le$ shift-differential $<$ 0.6 \textbf{OR} contribution-to-dental-plan = full \textbf{OR} contribution-to-dental-plan = half \textbf{OR} contribution-to-health-plan = half \textbf{OR} contribution-to-health-plan = none \blue{\textbf{AND}}\\\\0.3 $\le$ wage-increase-first-year $<$ 0.4 \textbf{OR} 0.6 $\le$ wage-increase-first-year $<$ 0.7 \textbf{OR} 0.7 $\le$ wage-increase-second-year $<$ 0.8 \textbf{OR} 0.4 $\le$ working-hours $<$ 0.5 \textbf{OR} pension = empl-contr \textbf{OR} pension = ret-allw \textbf{OR} 0.9 $\le$ statutory-holidays $<$ inf \\

\subsubsection*{ BNG(credit-g)}

\noindent foreign-worker \textbf{OR} checking-status = 'no checking' \textbf{OR} checking-status = >=200 \textbf{OR} -$\infty$ $\le$ duration $<$ 0.1 \textbf{OR} 0.1 $\le$ duration $<$ 0.2 \textbf{OR} credit-history = 'critical/other existing credit' \textbf{OR} purpose = 'used car' \textbf{OR} savings-status = 'no known savings' \textbf{OR} savings-status = >=1000 \textbf{OR} other-parties = guarantor \blue{\textbf{AND}}\\\\foreign-worker \textbf{OR} checking-status = 'no checking' \textbf{OR} checking-status = >=200 \textbf{OR} credit-history = 'delayed previously' \textbf{OR} purpose = 'used car' \textbf{OR} purpose = radio/tv \textbf{OR} other-parties = guarantor \textbf{OR} 0.4 $\le$ age $<$ 0.5 \textbf{OR}  \textbf{NOT} other-payment-plans = bank \blue{\textbf{AND}}\\\\checking-status = 'no checking' \textbf{OR} credit-history = 'critical/other existing credit' \textbf{OR} purpose = 'used car' \textbf{OR} purpose = radio/tv \textbf{OR} purpose = retraining \textbf{OR} employment = 4$\le$X$<$7 \textbf{OR} property-magnitude = 'real estate' \textbf{OR}  \textbf{NOT} checking-status = $<$0 \\

\bibliographystyle{theapa}
\bibliography{main}

\end{document}